\def\preprint{1}

\documentclass[11pt]{article}

\usepackage{ifthen}
\ifthenelse{\preprint = 1}{
  \usepackage{amsthm,fullpage}
  \newtheorem{theorem}{Theorem}[section]
  \newtheorem{lemma}[theorem]{Lemma}
  \newtheorem{corollary}[theorem]{Corollary}
}{
}
\usepackage[normalem]{ulem}
\usepackage{amsmath,amsfonts,amssymb,hyperref,multirow}
\usepackage{algorithm}
\usepackage{algpseudocode}
\usepackage{xcolor}
\hypersetup{colorlinks}

\newcommand{\Acal}{{\cal A}}

\newcommand{\Ecal}{{\cal E}}
\newcommand{\Fcal}{{\cal F}}

\newcommand{\Hcal}{{\cal H}}

\newcommand{\Ocal}{{\cal O}}
\newcommand{\Pcal}{{\cal P}}

\newcommand{\Scal}{{\cal S}}
\newcommand{\Tcal}{{\cal T}}

\newcommand{\Wcal}{{\cal W}}
\newcommand{\Xcal}{{\cal X}}
\newcommand{\Ycal}{{\cal Y}}

\newcommand{\oline}[1]{\mkern 1.5mu\overline{\mkern-1.5mu#1}}

\renewcommand{\hbar}{\oline{h}}

\newcommand{\kbar}{\oline{k}}

\newcommand{\wbar}{\oline{w}}

\newcommand{\Ftilde}{\tilde{F}}

\newcommand{\gtilde}{\tilde{g}}

\newcommand{\wtilde}{\tilde{w}}

\newcommand{\chat}{\hat{c}}

\newcommand{\Fhat}{\hat{F}}
\newcommand{\fhat}{\hat{f}}

\newcommand{\ghat}{\hat{g}}

\newcommand{\khat}{\hat{k}}
\newcommand{\Lhat}{\widehat{L}}

\newcommand{\shat}{\hat{s}}

\newcommand{\what}{\hat{w}}

\newcommand{\xhat}{\hat{x}}

\DeclareMathOperator{\st}       {s.\!t.}

\DeclareMathOperator{\trace}    {trace}
\DeclareMathOperator{\argmin}   {arg\,min}

\newcommand{\half} {\frac12}
\newcommand{\thalf}{\tfrac12}
\newcommand{\var}{{\mathbb{V}}}
\newcommand{\sgn}{\mbox{{sgn}}}
\renewcommand{\(}{\left(}
\renewcommand{\)}{\right)}

\newcommand{\E}{\mathbb{E}}

\newcommand{\N}[1]{\mathbb{N}^{#1}}
\renewcommand{\P}{\mathbb{P}}

\newcommand{\R}[1]{\mathbb{R}^{#1}}

\newcommand{\baligned}     {\begin{aligned}}
\newcommand{\ealigned}     {\end{aligned}}
\newcommand{\barray}       {\begin{array}}
\newcommand{\earray}       {\end{array}}
\newcommand{\bbmatrix}     {\begin{bmatrix}}
\newcommand{\ebmatrix}     {\end{bmatrix}}
\newcommand{\bcases}       {\begin{cases}}
\newcommand{\ecases}       {\end{cases}}
\newcommand{\bcenter}      {\begin{center}}
\newcommand{\ecenter}      {\end{center}}
\newcommand{\bcolumn}      {\begin{column}}
\newcommand{\ecolumn}      {\end{column}}
\newcommand{\bcolumns}     {\begin{columns}}
\newcommand{\ecolumns}     {\end{columns}}
\newcommand{\benumerate}   {\begin{enumerate}}
\newcommand{\eenumerate}   {\end{enumerate}}
\newcommand{\bequation}    {\begin{equation}}
\newcommand{\eequation}    {\end{equation}}
\newcommand{\bequationn}   {\begin{equation*}}
\newcommand{\eequationn}   {\end{equation*}}
\newcommand{\bfigure}      {\begin{figure}}
\newcommand{\efigure}      {\end{figure}}
\newcommand{\bflushright}  {\begin{flushright}}
\newcommand{\eflushright}  {\end{flushright}}
\newcommand{\bitemize}     {\begin{itemize}}
\newcommand{\eitemize}     {\end{itemize}}
\newcommand{\bpmatrix}     {\begin{pmatrix}}
\newcommand{\epmatrix}     {\end{pmatrix}}
\newcommand{\bsubequations}{\begin{subequations}}
\newcommand{\esubequations}{\end{subequations}}
\newcommand{\btable}       {\begin{table}}
\newcommand{\etable}       {\end{table}}
\newcommand{\btabular}     {\begin{tabular}}
\newcommand{\etabular}     {\end{tabular}}
\newcommand{\bvmatrix}     {\begin{vmatrix}}
\newcommand{\evmatrix}     {\end{vmatrix}}
\newcommand{\bequalin}     {\bequationn\baligned}
\newcommand{\eequalin}     {\ealigned\eequationn}
\newcommand{\bequali}      {\bsubequations\begin{align}}
\newcommand{\eequali}      {\end{align}\esubequations}

\newtheorem{assumption}[theorem]{Assumption}
\newtheorem{example}[theorem]{Example}
\newtheorem{remark}[theorem]{Remark}

\newcommand{\balgorithm}  {\begin{algorithm}}
\newcommand{\ealgorithm}  {\end{algorithm}}
\newcommand{\balgorithmic}{\begin{algorithmic}}
\newcommand{\ealgorithmic}{\end{algorithmic}}
\newcommand{\bassumption} {\begin{assumption}}
\newcommand{\eassumption} {\end{assumption}}
\newcommand{\bcorollary}  {\begin{corollary}}
\newcommand{\ecorollary}  {\end{corollary}}
\newcommand{\bdefinition} {\begin{definition}}
\newcommand{\edefinition} {\end{definition}}
\newcommand{\bexample}    {\begin{example}}
\newcommand{\eexample}    {\end{example}}
\newcommand{\blemma}      {\begin{lemma}}
\newcommand{\elemma}      {\end{lemma}}
\newcommand{\bproblem}    {\begin{problem}}
\newcommand{\eproblem}    {\end{problem}}
\newcommand{\bproof}      {\begin{proof}}
\newcommand{\eproof}      {\end{proof}}
\newcommand{\bremark}     {\begin{remark}}
\newcommand{\eremark}     {\end{remark}}
\newcommand{\btheorem}    {\begin{theorem}}
\newcommand{\etheorem}    {\end{theorem}}
\usepackage{placeins,afterpage}
\usepackage[customcolors]{hf-tikz}
\usepackage{adjustbox}
\hypersetup{colorlinks,urlcolor=blue}

\renewcommand{\figurename}{Fig.}
\graphicspath{{figures/}{}}
\def\indicator#1{{\mathrm{1\mkern-4.6mu l\!}\left[{#1}\right]}}
\newcommand{\VCdim}{{d_{\Hcal}}}
\newcounter{figtemp}
\def\nmb{n_{\rm mb}}

\title{Optimization Methods for Large-Scale Machine Learning}

\author{L\'eon Bottou\thanks{
        Facebook AI Research, New York, NY, USA.
        E-mail: \texttt{\href{mailto:leon@bottou.org}{leon@bottou.org}}}
   \and Frank E.~Curtis\thanks{
        Department of Industrial and Systems Engineering, Lehigh University, Bethlehem, PA, USA.
        Supported by U.S. Department of Energy grant DE--SC0010615 
        and U.S. National Science Foundation grant DMS--1016291.
        E-mail: \texttt{\href{mailto:frank.e.curtis@gmail.com}{frank.e.curtis@gmail.com}}}
   \and Jorge Nocedal\thanks{
        Department of Industrial Engineering and Management Sciences, 
        Northwestern University, Evanston, IL, USA.
        Supported by Office of Naval Research grant N00014-14-1-0313 P00003 and Department of Energy grant 
        DE-FG02-87ER25047.
        E-mail: \texttt{\href{mailto:j-nocedal@northwestern.edu}{j-nocedal@northwestern.edu}}}}

\begin{document}

\maketitle

\begin{abstract}
  This paper provides a review and commentary on the past, present, and future of numerical optimization algorithms in the context of machine learning applications.  Through case studies on text classification and the training of deep neural networks, we discuss how optimization problems arise in machine learning and what makes them challenging.  A major theme of our study is that large-scale machine learning represents a distinctive setting in which the stochastic gradient~(SG) method has traditionally played a central role while conventional gradient-based nonlinear optimization techniques typically falter.  Based on this viewpoint, we present a comprehensive theory of a straightforward, yet versatile SG algorithm, discuss its practical behavior, and highlight  opportunities for designing algorithms with improved performance.  This leads to a discussion about the next generation of optimization methods for large-scale machine learning, including an investigation of two main streams of research on techniques that diminish noise in the stochastic directions and methods that make use of second-order derivative approximations.
\end{abstract}

\ifthenelse{\preprint = 0}{



}{}

\ifthenelse{\preprint = 1}{
  \tableofcontents
  \newpage
}{}

\section{Introduction}\label{sec.introduction}
\setcounter{equation}{0}
\setcounter{theorem}{0}
\setcounter{algorithm}{0}
\setcounter{figure}{0}
\setcounter{table}{0}

The promise of artificial intelligence has been a topic of both public and private interest for decades.  Starting in the 1950s, there has been great hope that classical artificial intelligence techniques based on logic, knowledge representation, reasoning, and planning would result in revolutionary software that could, amongst other things, understand language, control robots, and provide expert advice.  Although advances based on such techniques may be in store in the future, many researchers have started to doubt these classical approaches, choosing instead to focus their efforts on the design of systems based on statistical techniques, such as in the rapidly evolving and expanding field of \emph{machine learning}.

Machine learning and the intelligent systems that have been borne out of it---such as search engines, recommendation platforms, and speech and image recognition software---have become an indispensable part of modern society.  Rooted in statistics and relying heavily on the efficiency of numerical algorithms, machine learning techniques capitalize on the world's increasingly powerful computing platforms and the availability of datasets of immense size.  In addition, as the fruits of its efforts have become so easily accessible to the public through various modalities---such as \emph{the cloud}---interest in machine learning is bound to continue its dramatic rise, yielding further  societal, economic, and scientific impacts.

One of the pillars of machine learning is \emph{mathematical optimization}, which, in this context, involves the numerical computation of parameters for a system designed to make decisions based on yet unseen data.  That is, based on currently available data, these parameters are chosen to be optimal with respect to a given learning problem.  The success of certain optimization methods for machine learning has inspired great numbers in various research communities to tackle even more challenging machine learning problems, and to design new methods that are more widely applicable.

The purpose of this paper is to provide a review and commentary on the past, present, and future of the use of numerical optimization algorithms in the context of machine learning applications.  A major theme of this work is that large-scale machine learning represents a distinctive setting in which traditional nonlinear optimization techniques typically falter, and so should be considered secondary to alternative classes of approaches that respect the statistical nature of the underlying problem of interest.

Overall, this paper attempts to provide answers for the following questions.

\benumerate
  \item How do optimization problems arise in machine learning applications and what makes them challenging?
  \item What have been the most successful optimization methods for large-scale machine learning and why?
  \item What recent advances have been made in the design of algorithms and what are open questions in this research area?
\eenumerate

We answer the first question with the aid of two case studies.  The first, a study of text classification, represents an application in which the success of machine learning has been widely recognized and celebrated.  The second, a study of perceptual tasks such as speech or image recognition, represents an application in which machine learning still has had great success, but in a much more enigmatic manner that leaves many questions unanswered.  These case studies also illustrate the variety of optimization problems that arise in machine learning: the first involves \emph{convex} optimization problems---derived from the use of logistic regression or support vector machines---while the second typically involves \emph{highly nonlinear and nonconvex} problems---derived from the use of deep neural networks.

With these case studies in hand, we turn our attention to the latter two questions on optimization algorithms, the discussions around which represent the bulk of the paper.  Whereas traditional gradient-based methods may be effective for solving small-scale learning problems in which a \emph{batch} approach may be used, in the context of large-scale machine learning it has been a \emph{stochastic} algorithm---namely, the \emph{stochastic gradient} method (SG) proposed by Robbins and Monro \cite{RobbMonr51}---that has been the core strategy of interest.  Due to this central role played by SG, we discuss its fundamental theoretical and practical properties within a few contexts of interest.  We also discuss recent trends in the design of optimization methods for machine learning, organizing them according to their relationship to SG.  We discuss: $(i)$ noise reduction methods that attempt to borrow from the strengths of batch methods, such as their fast convergence rates and ability to exploit parallelism; $(ii)$ methods that incorporate approximate second-order derivative information with the goal of dealing with nonlinearity and ill-conditioning; and $(iii)$ methods for solving regularized problems designed to avoid overfitting and allow for the use of high-dimensional models.  Rather than contrast SG and other methods based on the results of numerical experiments---which might bias our review toward a limited test set and implementation details---we focus our attention on fundamental computational trade-offs and theoretical properties of optimization methods.

We close the paper with a look back at what has been learned, as well as additional thoughts about the future of optimization methods for machine learning.

\section{Machine Learning Case Studies}\label{sec.case_studies}
\setcounter{equation}{0}
\setcounter{theorem}{0}
\setcounter{algorithm}{0}
\setcounter{figure}{0}
\setcounter{table}{0}

Optimization problems arise throughout machine learning.  We provide two case studies that illustrate their role in the selection of prediction functions in state-of-the-art machine learning systems.  We focus on cases that involve very large datasets and for which the number of model parameters to be optimized is also large.  By remarking on the structure and scale of such problems, we provide a glimpse into the challenges that make them difficult to solve.

\subsection{Text Classification via Convex Optimization}\label{sec.text_classification}

The assignment of natural language text to predefined classes based on their contents is one of the fundamental tasks of information management \cite{DumaPlatHechSaha98}.  Consider, for example, the task of determining whether a text document is one that discusses politics.  Educated humans can make a determination of this type unambiguously, say by observing that the document contains the names of well-known politicians.  Early text classification systems attempted to consolidate knowledge from human experts by building on such observations to formally characterize the word sequences that signify a discussion about a topic of interest (e.g., politics).  Unfortunately, however, concise characterizations of this type are difficult to formulate.  Rules about which word sequences do or do not signify a topic need to be refined when new documents arise that cannot be classified accurately based on previously established rules.  The need to coordinate such a growing collection of possibly contradictory rules limits the applicability of such systems to relatively simple tasks.

By contrast, the statistical machine learning approach begins with the collection of a sizeable set of examples $\{(x_1,y_1),\dots,(x_n,y_n)\}$, where for each $i \in \{1,\dots,n\}$ the vector $x_i$ represents the \emph{features} of a text document (e.g., the words it includes) and the scalar $y_i$ is a \emph{label} indicating whether the document belongs ($y_i=1$) or not ($y_i=-1$) to a particular class (i.e., topic of interest).  With such a set of examples, one can construct a classification program, defined by a \emph{prediction function}~$h$, and measure its performance by counting how often the program prediction~$h(x_i)$ differs from the correct prediction~$y_i$.  In this manner, it seems judicious to search for a prediction function that minimizes the frequency of observed misclassifications, otherwise known as the \emph{empirical risk} of misclassification:
\bequation\label{eqn.sec2.empiricalerror}
  R_n(h) = \frac{1}{n} \sum_{i=1}^n \indicator{h(x_i) \neq y_i},\ \ \text{where}\ \ \indicator{A} = \bcases 1 & \text{if $A$ is true,} \\ 0 & \text{otherwise.} \ecases
\eequation
The idea of minimizing such a function gives rise to interesting conceptual issues.  Consider, for example, a function that simply memorizes the examples, such as
\bequation\label{eqn.sec2.byrote}
  h^{\rm rote}(x) = \bcases y_i & \text{if $x=x_i$ for some $i \in \{1,\dots,n\}$}, \\ \pm1 & \text{(arbitrarily) otherwise.} \ecases
\eequation
This prediction function clearly minimizes \eqref{eqn.sec2.empiricalerror}, but it offers no performance guarantees on documents that do not appear in the examples.  To avoid such rote memorization, one should aim to find a prediction function that \emph{generalizes} the concepts that may be learned from the examples.

One way to achieve good generalized performance is to choose amongst a carefully selected class of prediction functions, perhaps satisfying certain smoothness conditions and enjoying a convenient parametric representation.  How such a class of functions may be selected is not straightforward; we discuss this issue in more detail in~\S\ref{sec.SRM}. For now, we only mention that common practice for choosing between prediction functions belonging to a given class is to compare them using \emph{cross-validation} procedures that involve splitting the examples into three disjoint subsets: a \emph{training set}, a \emph{validation set}, and a \emph{testing set}.  The process of optimizing the choice of $h$ by minimizing~$R_n$ in~\eqref{eqn.sec2.empiricalerror} is carried out on the training set over the set of candidate prediction functions, the goal of which is to pinpoint a small subset of viable candidates.  The generalized performance of each of these remaining candidates is then estimated using the validation set, the best performing of which is chosen as the selected function.  The testing set is only used to estimate the generalized performance of this selected function.

Such experimental investigations have, for instance, shown that the \emph{bag of words} approach works very well for text classification \cite{DumaPlatHechSaha98,Joac97}.  In such an approach, a text document is represented by a feature vector~$x\in\R{d}$ whose components are associated with a prescribed set of vocabulary words; i.e., each nonzero component indicates that the associated word appears in the document.  (The capabilities of such a representation can also be increased by augmenting the set of words with entries representing short sequences of words.)  This encoding scheme leads to very sparse vectors.  For instance, the canonical encoding for the standard RCV1 dataset \cite{LewiYangRoseLi04} uses a vocabulary of $d=47,\!152$ words to represent news stories that typically contain fewer than $1,\!000$ words. Scaling techniques can be used to give more weight to distinctive words, while scaling to ensure that each document has $\|x\|=1$ can be performed to compensate for differences in document lengths \cite{LewiYangRoseLi04}.

Thanks to such a high-dimensional sparse representation of documents, it has been deemed empirically sufficient to consider prediction functions of the form $h(x;w,\tau) = w^Tx - \tau$.  Here, $w^Tx$ is a linear discriminant parameterized by $w\in\R{d}$ and $\tau \in \R{}$ is a bias that provides a way to compromise between precision and recall.\footnote{Precision and recall, defined as the probabilities $\P[y=1|h(x)=1]$ and $\P[h(x)=1|y=1]$, respectively, are convenient measures of classifier performance when the class of interest represents only a small fraction of all documents.}  The accuracy of the predictions could be determined by counting the number of times that $\mathrm{sign}(h(x;w,\tau))$ matches the correct label, i.e., $1$ or $-1$.  However, while such a prediction function may be appropriate for classifying new documents, formulating an optimization problem around it to choose the parameters $(w,\tau)$ is impractical in large-scale settings due to the combinatorial structure introduced by the sign function, which is discontinuous.  Instead, one typically employs a continuous approximation through a \emph{loss} function that measures a cost for predicting $h$ when the true label is $y$; e.g., one may choose a \emph{log-loss} function of the form $\ell(h,y) = \log(1+\exp(-hy))$.  Moreover, one obtains a class of  prediction functions via the addition of a regularization term parameterized by a scalar $\lambda > 0$, leading to a convex optimization problem:
\bequation\label{eqn.sec2.textsvm}
  \min_{(w,\tau) \in \R{d}\times\R{}}\ \frac{1}{n}\sum_{i=1}^{n} \ell(h(x_i;w,\tau),y_i) + \frac{\lambda}{2} \|w\|_2^2.
\eequation
This problem may be solved multiple times for a given training set with various values of $\lambda>0$, with the ultimate solution $(w_*,\tau_*)$ being the one that yields the best performance on a validation set.

Many variants of problem~\eqref{eqn.sec2.textsvm} have also appeared in the literature.  For example, the well-known support vector machine problem~\cite{CortVapn95} amounts to using the \emph{hinge loss} $\ell(h,y)=\max(0,1-hy)$.  Another popular feature is to use an $\ell_1$-norm regularizer, namely $\lambda \|w\|_1$, which favors sparse solutions and therefore restricts attention to a subset of vocabulary words.  Other more sophisticated losses can also target a specific precision/recall tradeoff or deal with hierarchies of classes; e.g., see \cite{GassPapp10,DengDing14}. The choice of loss function is usually guided by experimentation.

In summary, both theoretical arguments~\cite{CortVapn95} and experimental evidence~\cite{Joac97} indicate that a carefully selected family of prediction functions and such high-dimensional representations of documents leads to good performance while avoiding \emph{overfitting}---recall \eqref{eqn.sec2.byrote}---of the training set.  The use of a simple surrogate, such as \eqref{eqn.sec2.textsvm}, facilitates experimentation and has been very successful for a variety of problems even beyond text classification. Simple surrogate models are, however, not the most effective in all applications.  For example, for certain perceptual tasks, the use of deep neural networks---which lead to large-scale, highly nonlinear, and \emph{nonconvex} optimization problems---has produced great advances that are unmatched by approaches involving simpler, convex models.  We discuss such problems next.

\subsection{Perceptual Tasks via Deep Neural Networks}\label{sec.deep_neural_nets}

\begin{figure}[ht]
\center
\includegraphics[width=0.9\linewidth]{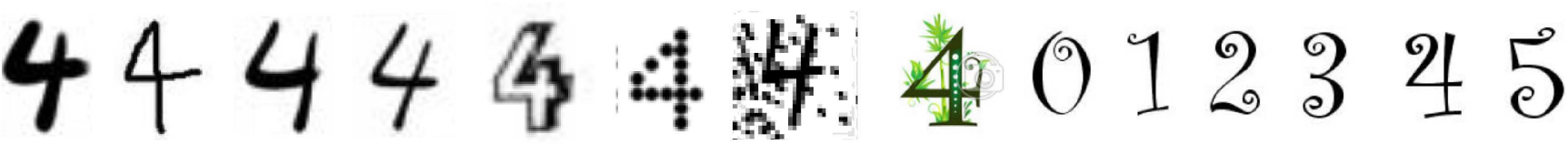}
\caption{\label{fig.sec2.fours}\relax
  No known prescribed rules express all pixel combinations that represent \emph{four}.}
\end{figure}

Just like text classification tasks, perceptual tasks such as speech and image recognition are not well performed in an automated manner using computer programs based on sets of prescribed rules.  For instance, the infinite diversity of writing styles easily defeats attempts to concisely specify which pixel combinations represent the digit \emph{four}; see Figure~\ref{fig.sec2.fours}.
One may attempt to design heuristic techniques to handle such eclectic instantiations of the same object, but as previous attempts to design such techniques have often failed, computer vision researchers are increasingly embracing machine learning techniques.

During the past five years, spectacular applicative successes on perceptual problems have been achieved by machine learning techniques through the use of \emph{deep neural networks} (DNNs).  Although there are many kinds of DNNs, most recent advances have been made using essentially the same types that were popular in the 1990s and almost forgotten in the 2000s \cite{NipsAll}.  What have made recent successes possible are the availability of much larger datasets and greater computational resources.

Because DNNs were initially inspired by simplified models of biological neurons~\cite{RumeHintWill86a,RumeHintWill86b}, they are often described with jargon borrowed from neuroscience.  For the most part, this jargon turns into a rather effective language for describing a prediction function~$h$ whose value is computed by applying successive transformations to a given input vector~$x_i \in \R{d_0}$. These transformations are made in \emph{layers}.  For example, a canonical fully connected layer performs the computation
\bequation\label{eqn.sec2.flayer}
  x_i^{(j)} = s(W_j\,x_i^{(j-1)} + b_j) \in \R{d_j},
\eequation
where $x_i^{(0)} = x_i$, the matrix $W_j \in \R{d_j\times d_{j-1}}$ and vector $b_j \in \R{d_j}$ contain the $j$th layer parameters, and $s$ is a component-wise nonlinear \emph{activation} function.  Popular choices for the activation function include the sigmoid function $s(x)=1/(1+\exp(-x))$ and the hinge function $s(x)=\max\{0,x\}$ (often called a rectified linear unit (ReLU) in this context).  In this manner, the ultimate output vector $x_i^{\smash{(J)}}$ leads to the prediction function value $h(x_i;w)$, where the parameter vector $w$ collects all the parameters $\{(W_1,b_1),\dots,(W_J,b_J)\}$ of the successive layers.

Similar to~\eqref{eqn.sec2.textsvm}, an optimization problem in this setting involves the collection of a training set $\{(x_1,y_1)\dots(x_n,y_n)\}$ and the choice of a loss function $\ell$ leading to
\bequation\label{oxf}
  \min_{w\in\R{d}}\ \frac{1}{n}\sum_{i=1}^n \ell(h(x_i;w),y_i).
\end{equation}
However, in contrast to~\eqref{eqn.sec2.textsvm}, this optimization problem is highly nonlinear and nonconvex, making it intractable to solve to global optimality.  That being said, machine learning experts have made great strides in the use of DNNs by computing approximate solutions by gradient-based methods.  This has been made possible by the conceptually straightforward, yet crucial observation that the gradient of the objective in \eqref{oxf} with respect to the parameter vector $w$ can be computed by the chain rule using algorithmic differentiation \cite{Grie14}.  This differentiation technique is known in the machine learning community as \emph{back propagation} \cite{RumeHintWill86a,RumeHintWill86b}.

The number of layers in a DNN and the size of each layer are usually determined by performing comparative experiments and evaluating the system performance on a validation set, as in the procedure in \S\ref{sec.text_classification}.  A contemporary fully connected neural network for speech recognition typically has five to seven layers.  This amounts to tens of millions of parameters to be optimized, the training of which may require up to thousands of hours of speech data (representing hundreds of millions of training examples) and weeks of computation on a supercomputer.  Figure~\ref{fig.sec2.speechresults} illustrates the word error rate gains achieved by using DNNs for acoustic modeling in three state-of-the-art speech recognition systems. These gains in accuracy are so significant that DNNs are now used in all the main commercial speech recognition products.

\bfigure[ht]
  \center
  \includegraphics[width=.6\linewidth]{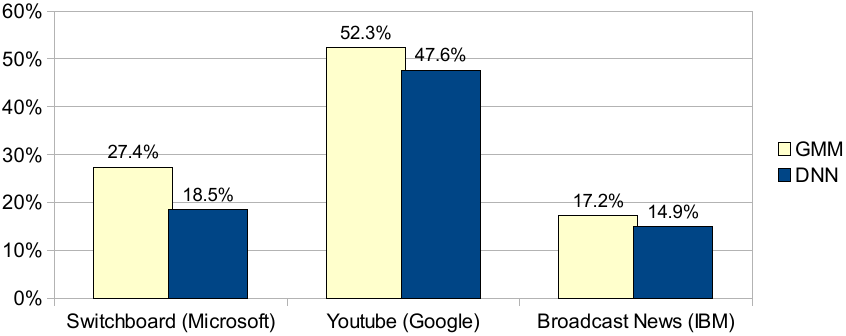}
  \caption{Word error rates reported by three different research groups on three standard speech recognition benchmarks.  For all three groups, deep neural networks (DNNs) significantly outperform the traditional Gaussian mixture models (GMMs)~\cite{DengHintKing13}.  These experiments were performed between 2010 and 2012 and were instrumental in the recent revival of DNNs.}
  \label{fig.sec2.speechresults}
\efigure

At the same time, convolutional neural networks (CNNs) have proved to be very effective for computer vision and signal processing tasks~\cite{LeCuBoseDenkHendHoweHubbJack89,BottFogeBlanLien89,LeCuBottBengHaff98,KrizSutsHint12}.  Such a network is composed of convolutional layers, wherein the parameter matrix $W_j$ is a circulant matrix and the input $x_i^{(j-1)}$ is intepreted as a multichannel image.  The product $W_j x_i^{(j-1)}$ then computes the convolution of the image by a trainable filter while the activation function---which are piecewise linear functions as opposed to sigmoids---can perform more complex operations that may be interpreted as image rectification, contrast normalization, or subsampling.  Figure~\ref{fig.sec2.cnn} represents the architecture of the winner of the landmark 2012 ImageNet Large Scale Visual Recognition Competition (ILSVRC) \cite{ImageNet}.  The figure illustrates a CNN with five convolutional layers and three fully connected layers~\cite{KrizSutsHint12}.  The input vector represents the pixel values of a $224\times224$ image while the output scores represent the odds that the image belongs to each of $1,\!000$ categories.  This network contains about 60 million parameters, the training of which on a few million labeled images takes a few days on a dual GPU workstation.

\bfigure[ht]
  \center
  \includegraphics[width=.75\linewidth]{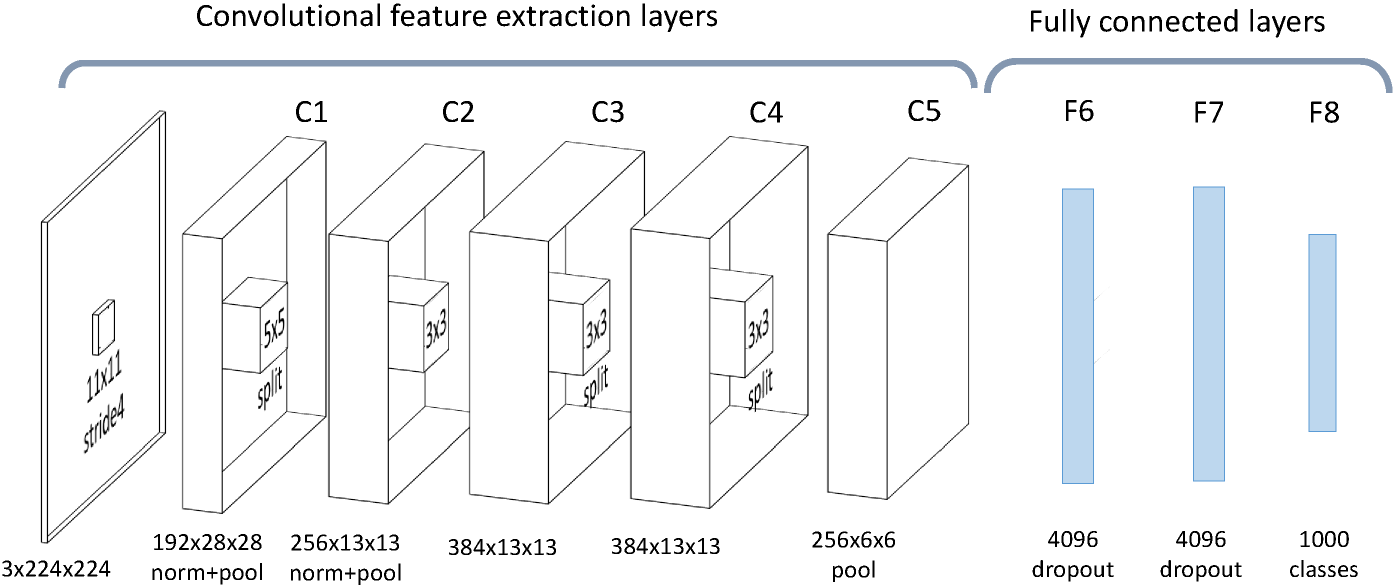}
  \caption{Architecture for image recognition. The 2012 ILSVRC winner consists of eight layers~\cite{KrizSutsHint12}. Each layer performs a linear transformation (specifically, convolutions in layers C1--C5 and matrix multiplication in layers F6--F8) followed by nonlinear transformations (rectification in all layers, contrast normalization in C1--C2, and pooling in C1--C2 and C5).  Regularization with dropout noise is used in layers F6--F7.}
  \label{fig.sec2.cnn}
\efigure

Figure~\ref{fig.sec2.imagenetresults} illustrates the historical error rates of the winner of the 2012 ILSVRC.  In this competition, a classification is deemed successful if the correct category appeared among the top five categories returned by the system.  The large performance gain achieved in 2012 was confirmed in the following years, and today CNNs are considered the tool of choice for visual object recognition~\cite{RazaAzizSullCarl14}.  They are currently deployed by numerous Internet companies for image search and face recognition.

\bfigure[ht]
  \center
  \includegraphics[width=.75\linewidth]{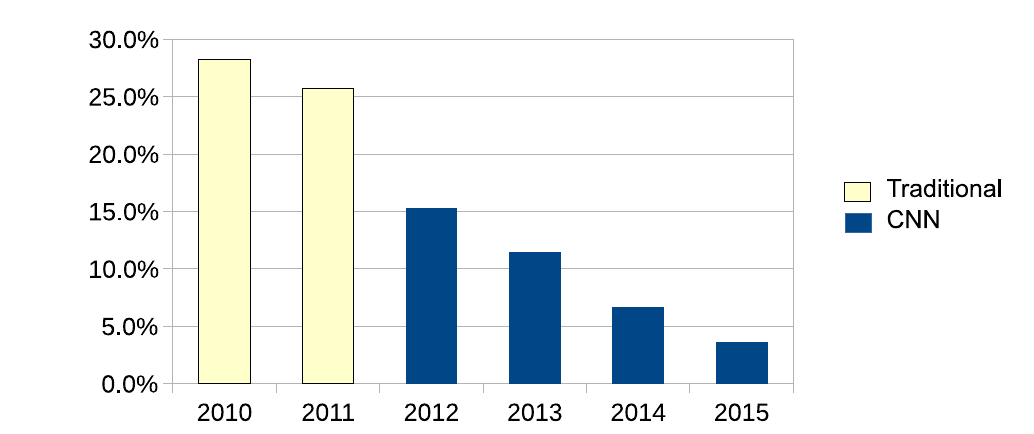}
  \caption{Historical \emph{top5} error rate of the annual winner of the ImageNet image classification challenge.  A convolutional neural network (CNN) achieved a significant performance improvement over all traditional methods in 2012.  The following years have cemented CNNs as the current state-of-the-art in visual object recognition \cite{KrizSutsHint12,RazaAzizSullCarl14}.}
  \label{fig.sec2.imagenetresults}
\end{figure}

The successes of DNNs in modern machine learning applications are undeniable.  Although the training process requires extreme skill and care---e.g., it is crucial to initialize the optimization process with a good starting point and to monitor its progress while correcting conditioning issues as they appear \cite{LeCuBottOrrMuel98}---the mere fact that one can do anything useful with such large, highly nonlinear and nonconvex models is remarkable.

\subsection{Formal Machine Learning Procedure}\label{sec.SRM}

Through our case studies, we have illustrated how a process of machine learning leads to the selection of a prediction function $h$ through solving an optimization problem.  Moving forward, it is necessary to formalize our presentation by discussing in greater detail the principles behind the selection process, stressing the theoretical importance of \emph{uniform laws of large numbers} as well as the practical importance of  \emph{structural risk minimization}.

For simplicity, we continue to focus on the problems that arise in the context of \emph{supervised classification}; i.e., we focus on the optimization of prediction functions for labeling unseen data based on information contained in a set of labeled training data.  Such a focus is reasonable as many unsupervised and other learning techniques reduce to optimization problems of comparable form; see, e.g., \cite{Vapn83}.

\paragraph{\textbf{Fundamentals}}

Our goal is to determine a prediction function $h : \Xcal \to \Ycal$ from an input space~$\Xcal$ to an output space~$\Ycal$ such that, given $x \in \Xcal$, the value $h(x)$ offers an accurate prediction about the true output $y$.  That is, our goal is to choose a prediction function that avoids rote memorization and instead generalizes the concepts that can be learned from a given set of examples. To do this, one should choose the prediction function~$h$ by attempting to minimize a risk measure over an adequately selected family of prediction functions \cite{VapnCher68}, call it $\Hcal$.

To formalize this idea, suppose that the examples are sampled from a joint probability distribution function $P(x,y)$ that simultaneously represents the distribution~$P(x)$ of inputs as well as the conditional probability $P(y|x)$ of the label $y$ being appropriate for an input~$x$.  (With this view, one often refers to the examples as \emph{samples}; we use both terms throughout the rest of the paper.)  Rather than one that merely minimizes the empirical risk \eqref{eqn.sec2.empiricalerror}, one should seek to find $h$ that yields a small \emph{expected risk} of misclassification \emph{over all possible inputs}, i.e., an~$h$ that minimizes
\begin{equation}\label{eqn.sec2.expectederror}
  R(h) = \P[h(x) \neq y] = \E[\indicator{h(x) \neq y}],
\end{equation}
where $\P[A]$ and $\E[A]$ respectively denote the probability and expected value of $A$.  Such a framework is \emph{variational} since we are optimizing over a set of functions, and is \emph{stochastic} since the objective function involves an expectation.  

While one may desire to minimize the expected risk~\eqref{eqn.sec2.expectederror}, in practice one must attempt to do so without explicit knowledge of $P$.  Instead, the only tractable option is to construct a surrogate problem that relies solely on the examples $\{(x_i,y_i)\}_{i=1}^n$.  Overall, there are two main issues that must be addressed: $(i)$ how to choose the parameterized family of prediction functions~$\Hcal$ and $(ii)$ how to determine (and find) the particular prediction function $h \in \Hcal$ that is optimal.

\paragraph{\textbf{Choice of Prediction Function Family}}

The family of functions $\Hcal$ should be determined with three \emph{potentially competing} goals in mind.  First, $\Hcal$ should contain prediction functions that are able to achieve a low empirical risk over the training set, so as to avoid bias or underfitting the data.  This can be achieved by selecting a rich family of functions or by using \emph{a priori} knowledge to select a well-targeted family.  Second, the gap between expected risk and empirical risk, namely, $R(h)-R_n(h)$, should be small over all $h \in \Hcal$.  Generally, this gap decreases when one uses more training examples, but, due to potential overfitting, it increases when one uses richer families of functions (see below).  This latter fact puts the second goal at odds with the first.  Third, $\Hcal$ should be selected so that one can efficiently solve the corresponding optimization problem, the difficulty of which may increase when one employs a richer family of functions and/or a larger training set.

Our observation about the gap between expected and empirical risk can be understood by recalling certain \emph{laws of large numbers}.  For instance, when the expected risk represents a misclassification probability as in \eqref{eqn.sec2.expectederror}, the Hoeffding inequality \cite{Hoef63} guarantees that, with probability at least $1-\eta$, one has
\bequationn
  |R(h)-R_n(h)| \leq \sqrt{\frac{1}{2n}\log\(\frac{2}{\eta}\)}\ \ \text{for a given}\ \ h \in \Hcal.
\eequationn
This bound offers the intuitive explanation that the gap decreases as one uses more training examples.  However, this view is insufficient for our purposes since, in the context of machine learning, $h$ is not a fixed function!  Rather, $h$ is the variable over which one is optimizing.

For this reason, one often turns to \emph{uniform laws of large numbers} and the concept of the Vapnik-Chervonenkis (VC) dimension of $\Hcal$, a measure of the \emph{capacity} of such a family of functions \cite{VapnCher68}.  For the intuition behind this concept, consider, e.g., a binary classification scheme in $\R{2}$ where one assigns a label of $1$ for points above a polynomial and $-1$ for points below.  The set of linear polynomials has a low capacity in the sense that it is only capable of accurately classifying training points that can be separated by a line; e.g., in two variables, a linear classifier has a VC dimension of three.  A set of high-degree polynomials, on the other hand, has a high capacity since it can accurately separate training points that are interspersed; the VC dimension of a polynomial of degree $D$ in $d$ variables is
$
\left(\begin{array}{c} d+D \\ d \end{array} \right).
$
That being said, the gap between empirical and expected risk can be larger for a set of high-degree polynomials since the high capacity allows them to overfit a given set of training data.

Mathematically, with the VC dimension measuring capacity, one can establish one of the most important results in learning theory: with $\VCdim$ defined as the VC dimension of $\Hcal$, one has with probability at least $1-\eta$ that
\begin{equation}\label{vcdim}
  \sup_{h\in\Hcal} |R(h)-R_n(h)| \leq \Ocal\(\sqrt{\frac{1}{2n}\log\left(\frac{2}{\eta}\right) + \frac{\VCdim}{n}\log\left(\frac{n}{\VCdim}\right)}\).
\end{equation}
This bound gives a more accurate picture of the dependence of the gap on the choice of $\Hcal$.  For example, it shows that for a fixed $\VCdim$,  uniform convergence is obtained by increasing the number of training points~$n$.   However, it also shows that, for a fixed~$n$, the gap can widen for larger~$\VCdim$.  Indeed, to maintain the same gap, one must increase~$n$ at the same rate if $\VCdim$ is increased.  The uniform convergence embodied in this result is crucial in machine learning since one wants to ensure that the prediction system performs well with any data provided to it.  In \S\ref{sec.work_complexity}, we employ a slight variant of this result to discuss computational trade-offs that arise in large-scale learning.\footnote{We also note that considerably better bounds hold when one can collect statistics on actual examples, e.g., by determining gaps dependent on an observed variance of the risk or by considering uniform bounds restricted to families of prediction functions that achieve a risk within a certain threshold of the optimum \cite{Dudl99,Mass00,Bous02}.} 

Interestingly, one quantity that does not enter in \eqref{vcdim} is the number of parameters that distinguish a particular member function $h$ of the family $\Hcal$.  In some settings such as logistic regression, this number is essentially the same as $\VCdim$, which might suggest that the task of optimizing over $h \in \Hcal$ is more cumbersome as $\VCdim$ increases.  However, this is not always the case.  Certain families of functions are amenable to minimization despite having a very large or even infinite number of parameters~\cite[Section~4.11]{Vapn98}.  For example, support vector machines~\cite{CortVapn95} were designed to take advantage of this fact~\cite[Theorem~10.3]{Vapn98}.

All in all, while bounds such as \eqref{vcdim} are theoretically interesting and provide useful insight, they are rarely used directly in practice since, as we have suggested in \S\ref{sec.text_classification} and \S\ref{sec.deep_neural_nets}, it is typically easier to estimate the gap between empirical and expected risk with \emph{cross-validation} experiments.  We now present ideas underlying a practical framework that respects the trade-offs mentioned above.

\paragraph{\textbf{Structural Risk Minimization}}

\begin{figure}[ht]
  \center
  \includegraphics[width=.67\linewidth]{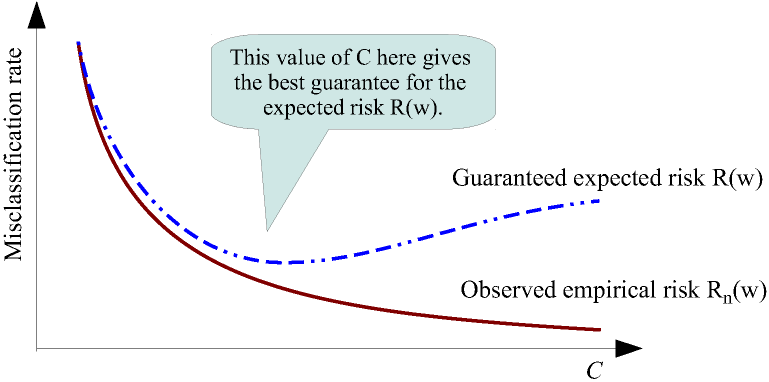}
  \caption{Illustration of structural risk minimization.  Given a set of $n$ examples, a decision function family $\Hcal$, and a relative preference function $\Omega$, the figure illustrates a typical relationship between the expected and empirical risks corresponding to a prediction function obtained by an optimization algorithm that minimizes an empirical risk $R_n(h)$ subject to $\Omega(h) \leq C$.  The optimal empirical risk decreases when~$C$ increases.  Meanwhile, the deviation between empirical and expected risk is bounded above by a quantity---which depends on $\Hcal$ and $\Omega$---that increases with $C$.  While not shown in the figure, the value of~$C$ that offers the best guarantee on the expected risk increases with~$n$, i.e., the number of examples; recall \eqref{vcdim}.}
  \label{fig.sec2.srm}
\end{figure}

An approach for choosing a prediction function that has proved to be widely successful in practice is \emph{structural risk minimization}~\cite{VapnCher74,Vapn98}.  Rather than choose a generic family of prediction functions---over which it would be both difficult to optimize and to estimate the gap between empirical and expected risks---one chooses a \emph{structure}, i.e., a collection of nested function families.  For instance, such a structure can be formed as a collection of subsets of a given family $\Hcal$ in the following manner: given a preference function $\Omega$, choose various values of a \emph{hyperparameter} $C$, according to each of which one obtains the subset $\Hcal_C := \{h \in \Hcal : \Omega(h) \leq C\}$.  Given a fixed number of examples, increasing $C$ reduces the empirical risk (i.e., the minimum of $R_n(h)$ over $h \in \Hcal_C$), but, after some point, it typically increases the gap between expected and empirical risks.  This phenomenon is illustrated in Figure~\ref{fig.sec2.srm}.

Other ways to introduce structures are to consider a regularized empirical risk $R_n(h) + \lambda \Omega(h)$ (an idea introduced in problem~\eqref{eqn.sec2.textsvm}, which may be viewed as the Lagrangian for minimizing $R_n(h)$ subject to $\Omega(h) \leq C$), enlarge the dictionary in a bag-of-words representation, increase the degree of a polynomial model function, or add to the dimension of an inner layer of a DNN.

Given such a set-up, one can avoid estimating the gap between empirical and expected risk by splitting the available data into subsets: a \emph{training set} used to produce a subset of candidate solutions, a \emph{validation set} used to estimate the expected risk for each such candidate, and a \emph{testing set} used to estimate the expected risk for the candidate that is ultimately chosen.  Specifically, over the training set, one minimizes an empirical risk measure $R_n$ over $\Hcal_C$ for various values of~$C$.  This results in a handful of candidate functions.  The validation set is then used to estimate the expected risk corresponding to each candidate solution, after which one chooses the function yielding the lowest estimated risk value.  Assuming a large enough range for~$C$ has been used, one often finds that the best solution does not correspond to the largest value of $C$ considered; again, see Figure~\ref{fig.sec2.srm}.

\begin{figure}[ht]
  \center
  \includegraphics[width=.67\linewidth]{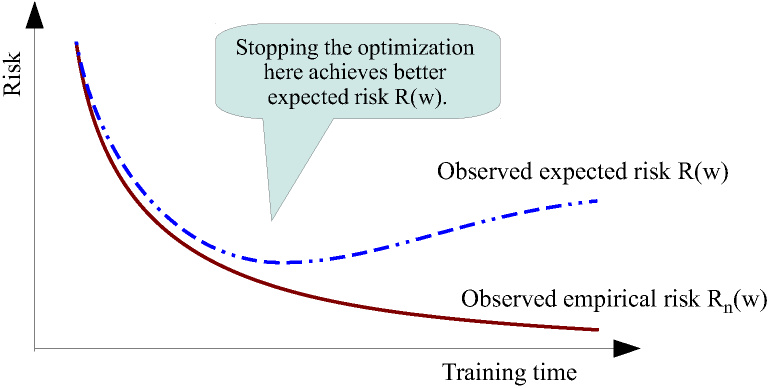}
  \caption{Illustration of early stopping.  Prematurely stopping the optimization of the empirical risk $R_n$ often results in a better expected risk $R$.  In this manner, the stopping time plays a similar role as the hyperparameter $C$ in the illustration of structural risk minimization in Figure~\ref{fig.sec2.srm}.}
  \label{fig.sec4.earlystopping}
\end{figure}

Another, albeit indirect avenue toward risk minimization is to employ an algorithm for minimizing~$R_n$, but terminate the algorithm \emph{early}, i.e., before an actual minimizer of $R_n$ is found.  In this manner, the role of the hyperparameter is played by the training time allowed, according to which one typically finds the relationships illustrated in Figure~\ref{fig.sec4.earlystopping}.  Theoretical analyses related to the idea of early stopping are much more challenging than those for other forms of structural risk minimization.  However, it is worthwhile to mention these effects since early stopping is a popular technique in practice, and is often \emph{essential} due to computational budget limitations.

Overall, the structural risk minimization principle has proved useful for many applications, and can be viewed as an alternative of the approach of employing expert human knowledge mentioned in \S\ref{sec.text_classification}.  Rather than encoding knowledge as formal classification rules, one encodes it via preferences for certain prediction functions over others, then explores the performance of various prediction functions that have been optimized under the influence of such preferences.

\section{Overview of Optimization Methods}\label{sec.overview}
\setcounter{equation}{0}
\setcounter{theorem}{0}
\setcounter{algorithm}{0}
\setcounter{figure}{0}
\setcounter{table}{0}

We now turn our attention to the main focus of our study, namely, numerical algorithms for solving optimization problems that arise in large-scale machine learning.  We begin by formalizing our problems of interest, which can be seen as generic statements of problems of the type described in~\S\ref{sec.case_studies} for minimizing expected and empirical risks.  We then provide an overview of two main classes of optimization methods---\emph{stochastic} and \emph{batch}---that can be applied to solve such problems, emphasizing some of the fundamental reasons why stochastic methods have inherent advantages.  We close this section with a preview of some of the advanced optimization techniques that are discussed in detail in later sections, which borrow ideas from both stochastic and batch methods.

\subsection{Formal Optimization Problem Statements}

As seen in \S\ref{sec.case_studies}, optimization problems in machine learning arise through the definition of prediction and loss functions that appear in measures of expected and empirical risk that one aims to minimize.  Our discussions revolve around the following definitions.

\paragraph{\textbf{Prediction and Loss Functions}}

Rather than consider a variational optimization problem over a generic family of prediction functions, we assume that the prediction function~$h$ has a fixed form and is parameterized by a real vector $w \in \R{d}$ over which the optimization is to be performed.  Formally, for some given $h(\cdot;\cdot) : \R{d_x} \times \R{d} \to \R{d_y}$, we consider the family of prediction functions
\bequationn
  \Hcal := \{h(\cdot;w) : w \in \R{d}\}.
\eequationn
We aim to find the prediction function in this family that minimizes the losses incurred from inaccurate predictions.  For this purpose, we assume a given loss function $\ell : \R{d_y} \times \R{d_y} \to \R{}$ as one that, given an input-output pair $(x,y)$, yields the loss $\ell(h(x;w),y)$ when $h(x;w)$ and $y$ are the predicted and true outputs, respectively.

\paragraph{\textbf{Expected Risk}}

Ideally, the parameter vector $w$ is chosen to minimize the expected loss that would be incurred from \emph{any} input-output pair. To state this idea formally, we assume that losses are measured with respect to a probability distribution $P(x,y)$ representing the true relationship between inputs and outputs.  That is, we assume that the input-output space $\R{d_x} \times \R{d_y}$ is endowed with $P : \R{d_x}\times\R{d_y} \to [0,1]$ and the objective function we wish to minimize is
\bequation\label{eq.exp_risk_w}
   R(w) = \int_{\R{d_x}\times\R{d_y}} \ell(h(x;w),y)\ dP(x,y) = \E[\ell(h(x;w),y)].
\eequation
We say that $R : \R{d} \to \R{}$ yields the \emph{expected risk} (i.e., expected loss) given a parameter vector $w$ with respect to the probability distribution $P$.

\paragraph{\textbf{Empirical Risk}}

While it may be desirable to minimize \eqref{eq.exp_risk_w}, such a goal is untenable when one does not have complete information about $P$.  Thus, in practice, one seeks the solution of a problem that involves an estimate of the expected risk $R$.  In supervised learning, one has access (either all-at-once or incrementally) to a set of $n \in \N{}$ independently drawn input-output samples $\{(x_i,y_i)\}_{i=1}^n \subseteq \R{d_x}\times\R{d_y}$, with which one may define the \emph{empirical risk} function $R_n : \R{d} \to \R{}$ by
\bequation\label{eq.emp_risk_w}
  R_n(w) = \frac{1}{n} \sum_{i=1}^n \ell(h(x_i;w),y_i).
\eequation
Generally speaking, minimization of $R_n$ may be considered the practical optimization problem of interest. For now, we consider the unregularized measure~\eqref{eq.emp_risk_w}, remarking that the optimization methods that we discuss in the subsequent sections can be applied readily when a smooth regularization term is included.
(We leave a discussion of nonsmooth regularizers until~\S\ref{sec.nonsmooth}.)

Note that, in \S\ref{sec.case_studies}, the functions $R$ and $R_n$ represented \emph{misclassification error}; see \eqref{eqn.sec2.empiricalerror} and \eqref{eqn.sec2.expectederror}.  However, these new definitions of $R$ and $R_n$ measure the loss as determined by the function $\ell$.  We use these latter definitions for the rest of the paper.

\paragraph{\textbf{Simplified Notation}}

The expressions \eqref{eq.exp_risk_w} and \eqref{eq.emp_risk_w} show explicitly how the expected and empirical risks depend on the loss function, sample space or sample set, etc.  However, when discussing optimization methods, we will often employ a simplified notation that also offers some avenues for generalizing certain algorithmic ideas.  In particular, let us represent a sample (or set of samples) by a random seed~$\xi$; e.g., one may imagine a realization of $\xi$ as a single sample $(x,y)$ from $\R{d_x}\times\R{d_y}$, or a realization of $\xi$ might be a set of samples $\{(x_i,y_i)\}_{i\in\Scal}$.  In addition, let us refer to the loss incurred for a given $(w,\xi)$ as $f(w;\xi)$, i.e.,
\begin{equation}  \label{composition}
  f \ \mbox{is the composition of the loss function $\ell$ and the prediction function $h$.}
\end{equation}
In this manner, the expected risk for a given $w$ is the expected value of this composite function taken with respect to the distribution of $\xi$:
\bequation\label{def.F}
  \textcolor{blue}{(\text{Expected Risk}) \qquad R(w) = \E[f(w;\xi)]}.
\eequation
In a similar manner, when given a set of realizations $\{\xi_{[i]}\}_{i=1}^n$ of $\xi$ corresponding to a sample set $\{(x_i,y_i)\}_{i=1}^n$, let us define the loss incurred by the parameter vector $w$ with respect to the $i$th sample as
\bequation\label{eq.fi}
  f_i(w) := f(w;\xi_{[i]}),
\eequation
and then write the empirical risk as the average of the sample losses:
\bequation\label{def.f}
  \textcolor{red}{(\text{Empirical Risk}) \qquad R_n(w) = \frac{1}{n} \sum_{i=1}^n f_i(w)}.
\eequation
For future reference, we use $\xi_{[i]}$ to denote the $i$th element of a fixed set of realizations of a random variable $\xi$, whereas, starting in~\S\ref{sec.sg}, we will use $\xi_k$ to denote the $k$th element of a sequence of random variables.

\subsection{Stochastic vs.~Batch Optimization Methods}

Let us now introduce some fundamental optimization algorithms for minimizing risk.  For the moment, since it is the typical setting in practice, we introduce two algorithm classes in the context of minimizing the empirical risk measure~$R_n$ in \eqref{def.f}.  Note, however, that much of our later discussion will focus on the performance of algorithms when considering the true measure of interest, namely, the expected risk $R$ in \eqref{def.F}.

Optimization methods for machine learning fall into two broad categories.  We refer to them as \emph{stochastic} and \emph{batch}.  The prototypical stochastic optimization method is the stochastic gradient method (SG) \cite{RobbMonr51}, which, in the context of minimizing $R_n$ and with $w_1 \in \R{d}$ given, is defined by
\bequation\label{eq.sg_init}
  w_{k+1} \gets w_k - \alpha_k \nabla f_{i_k}(w_k).
\eequation
Here, for all $k \in \N{} := \{1,2,\dots\}$, the index $i_k$ (corresponding to the seed $\xi_{[i_k]}$, i.e., the sample pair $(x_{i_k},y_{i_k})$) is chosen \emph{randomly} from $\{1,\dots,n\}$ and $\alpha_k$ is a positive stepsize.  Each iteration of this method is thus very cheap, involving only the computation of the gradient $\nabla f_{i_k}(w_k)$ corresponding to one sample.  The method is notable in that the iterate sequence is not determined uniquely by the function $R_n$, the starting point~$w_1$, and the sequence of stepsizes $\{\alpha_k\}$, as it would in a deterministic optimization algorithm.  Rather, $\{w_k\}$ is a stochastic process whose behavior is determined by the random sequence $\{i_k\}$.  Still, as we shall see in our analysis in~\S\ref{sec.sg}, while each direction $-\nabla f_{i_k}(w_k)$ might not be one of descent from $w_k$ (in the sense of yielding a negative directional derivative for $R_n$ from $w_k$), if it is a descent direction \emph{in expectation}, then the sequence $\{w_k\}$ can be guided toward a minimizer of $R_n$.

For many in the optimization research community, a \emph{batch} approach is a more natural and well-known idea.  The simplest such method in this class is the steepest descent algorithm---also referred to as the gradient, batch gradient, or full gradient method---which is defined by the iteration
\bequation\label{eq.fg_init}
  w_{k+1} \gets w_k - \alpha_k \nabla R_n(w_k) = w_k - \frac{\alpha_k}{n} \sum_{i=1}^n \nabla f_i(w_k).
\eequation
Computing the step $-\alpha_k \nabla R_n(w_k)$ in such an approach is more expensive than computing the step $-\alpha_k \nabla f_{i_k}(w_k)$ in SG, though one may expect that a better step is computed when all samples are considered in an iteration.

Stochastic and batch approaches offer different trade-offs in terms of per-iteration costs and expected per-iteration improvement in minimizing empirical risk.  Why, then, has SG risen to such prominence in the context of large-scale machine learning?  Understanding the reasoning behind this requires careful consideration of the computational trade-offs between stochastic and batch methods, as well as a deeper look into their abilities to guarantee improvement in the underlying expected risk~$R$.  We start to investigate these topics in the next subsection.

We remark in passing that the stochastic and batch approaches mentioned here have analogues in the simulation and stochastic optimization communities, where they are referred to as \emph{stochastic approximation} (SA) and \emph{sample average approximation} (SAA), respectively \cite{Fu02}.

\renewcommand{\figurename}{Inset}
\setcounter{figtemp}{\value{figure}}
\setcounter{figure}{0}
\begin{figure}[ht]
\begin{adjustbox}{minipage=\textwidth-20pt,margin=8pt 8pt 8pt -4pt,bgcolor=blue!5,frame=2pt}
  \caption{Herbert Robbins and Stochastic Approximation}
  \label{mini.sg}
  \medskip
  \setlength\parindent{.25in}
  
  The paper by Robbins and Monro \cite{RobbMonr51} represents a landmark in the history of numerical optimization methods.  Together with the invention of back propagation \cite{RumeHintWill86a,RumeHintWill86b}, it also represents one of the most notable developments in the field of machine learning.  The SG method was first proposed in \cite{RobbMonr51}, not as a gradient method, but as a Markov chain.

  \indent
  Viewed more broadly, the works by Robbins and Monro \cite{RobbMonr51} and Kalman \cite{Kalm60} mark the beginning of the field of \emph{stochastic approximation}, which studies the behavior of iterative methods that use noisy signals.  The initial focus on optimization led to the study of algorithms that track the solution of the ordinary differential equation $\dot{w} = -\nabla F(w)$.  Stochastic approximation theory has had a major impact in signal processing and in areas closer to the subject of this paper, such as pattern recognition \cite{Amar67} and neural networks \cite{Bott91}.

  After receiving his PhD, Herbert Robbins became a lecturer at New York University, where he co-authored with Richard Courant the popular book \emph{What is Mathematics?} \cite{RobbCour41}, which is still in print after more than seven decades \cite{RobbCour96}.  Robbins went on to become one of the most prominent mathematicians of the second half of the twentieth century, known for his contributions to probability, algebra, and graph theory.
\end{adjustbox}
\end{figure}
\renewcommand{\figurename}{Fig.}
\setcounter{figure}{\value{figtemp}}

\subsection{Motivation for Stochastic Methods}\label{sec.motiv_sg}

Before discussing the strengths of stochastic methods such as SG, one should not lose sight of the fact that batch approaches possess some intrinsic advantages.  First, when one has reduced the stochastic problem of minimizing the expected risk $R$ to focus exclusively on the deterministic problem of minimizing the empirical risk $R_n$, the use of full gradient information at each iterate opens the door for many deterministic gradient-based optimization methods.  That is, in a batch approach, one has at their disposal the wealth of nonlinear optimization techniques that have been developed over the past decades, including the full gradient method \eqref{eq.fg_init}, but also accelerated gradient, conjugate gradient, quasi-Newton, and inexact Newton methods \cite{NoceWrig06}.  (See \S\ref{sec.second_order} and \S\ref{sec.other} for discussion of these techniques.)  Second, due to the sum structure of $R_n$, a batch method can easily benefit from parallelization since the bulk of the computation lies in evaluations of $R_n$ and~$\nabla R_n$.  Calculations of these quantities can even be done in a distributed manner.

Despite these advantages, there are intuitive, practical, and theoretical reasons for following a stochastic approach.  Let us motivate them by contrasting the hallmark SG iteration~\eqref{eq.sg_init} with the full batch gradient iteration~\eqref{eq.fg_init}.

\paragraph{\textbf{Intuitive Motivation}}

On an intuitive level, SG employs information more efficiently than a batch method.  To see this, consider a situation in which a training set, call it $\Scal$, consists of ten copies of a set~$\Scal_{\rm sub}$.  A minimizer of empirical risk for the larger set $\Scal$ is clearly given by a minimizer for the smaller set $\Scal_{\rm sub}$, but if one were to apply a batch approach to minimize $R_n$ over $\Scal$, then each iteration would be \emph{ten times} more expensive than if one only had one copy of~$\Scal_{\rm sub}$.  On the other hand, SG performs the same computations in both scenarios, in the sense that the stochastic gradient computations involve choosing elements from $\Scal_{\rm sub}$ with the same probabilities.  In reality, a training set typically does not consist of exact duplicates of sample data, but in many large-scale applications the data does involve a good deal of (approximate) redundancy.  This suggests that using all of the sample data in every optimization iteration is inefficient.

A similar conclusion can be drawn by recalling the discussion in \S\ref{sec.case_studies} related to the use of training, validation, and testing sets.  If one believes that working with only, say, half of the data in the training set is sufficient to make good predictions on unseen data, then one may argue against working with the entire training set in every optimization iteration.  Repeating this argument, working with only a quarter of the training set may be useful at the start, or even with only an eighth of the data, and so on.  In this manner, we arrive at motivation for the idea that working with small samples, at least initially, can be quite appealing.

\paragraph{\textbf{Practical Motivation}}

The intuitive benefits just described have been observed repeatedly in practice, where one often finds very real advantages of SG in many applications.  As an example, Figure~\ref{wins} compares the performance of a batch L-BFGS method \cite{LiuNoce89,Noce80} (see \S\ref{sec.second_order}) and the SG method \eqref{eq.sg_init} with a constant stepsize (i.e., $\alpha_k=\alpha$ for all $k \in \N{}$) on a binary classification problem using a logistic loss objective function and the data from the RCV1 dataset mentioned in~\S\ref{sec.text_classification}.  The figure plots the empirical risk $R_n$ as a function of the number of accesses of a sample from the training set, i.e., the number of evaluations of a sample gradient $\nabla f_{i_k}(w_k)$.  Each set of $n$ consecutive accesses is called an \emph{epoch}.  The batch method performs only one step per epoch while SG performs~$n$ steps per epoch.  The plot shows the behavior over the first 10 epochs.  The advantage of SG is striking and representative of typical behavior in practice.  (One should note, however, that to obtain such efficient behavior, it was necessary to run SG repeatedly using different choices for the stepsize $\alpha$ until a good choice was identified for this particular problem.  We discuss theoretical and practical issues related to the choice of stepsize in our analysis in~\S\ref{sec.sg}.) \


\bfigure[ht]
  \center
  \includegraphics[width=0.5\linewidth]{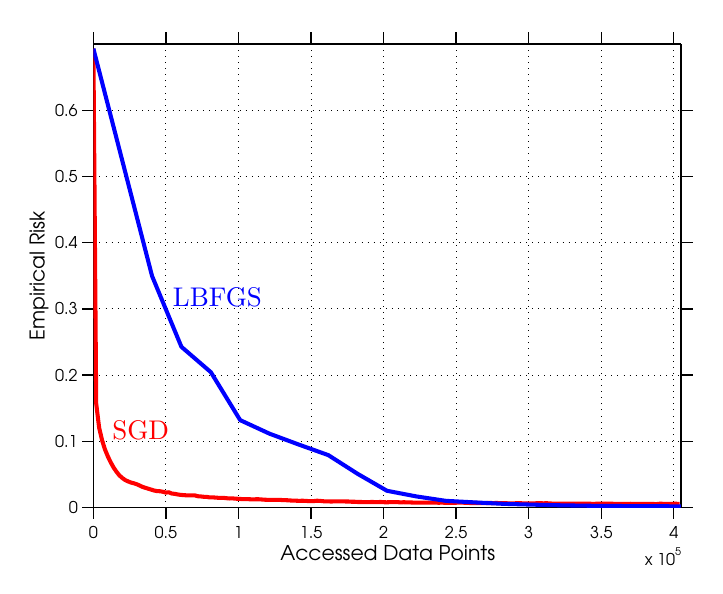}
  \caption{Empirical risk $R_n$ as a function of the number of accessed data points (ADP) for a batch L-BFGS method and the stochastic gradient (SG) method \eqref{eq.sg_init} on a binary classification problem with a logistic loss objective and the RCV1 dataset.  SG was run with a fixed stepsize of $\alpha = 4$.}
  \label{wins}
\efigure

At this point, it is worthwhile to mention that the fast initial improvement achieved by SG, followed by a drastic slowdown after 1~or~2 epochs, is common in practice and is fairly well understood.  An intuitive way to explain this behavior is by considering the following example due to Bertsekas~\cite{Bert15}.

\bexample\label{ex.region_of_confusion}
  Suppose that each $f_i$ in~\eqref{def.f} is a convex quadratic with minimal value at zero and  minimizers $w_{i,*}$ evenly distributed in $[-1,1]$ such that the minimizer of $R_n$ is $w_* = 0$; see Figure~\ref{confusion}.  At $w_1 \ll -1$, SG  will, with certainty, move to the right (toward~$w_*$).  Indeed, even if a subsequent iterate lies slightly to the right of the minimizer $w_{1,*}$ of the ``leftmost'' quadratic, it is likely (but not certain) that SG will continue moving to the right.  However, as iterates near~$w_*$, the algorithm enters a \emph{region of confusion} in which there is a significant chance that a step will not move toward~$w_*$.  In this manner, progress will slow significantly.  Only with more complete gradient information could the method know with certainty how to move toward $w_*$.
\eexample

\bfigure[ht]
  \center
  \includegraphics[width=0.5\linewidth,clip=true,trim=100 200 120 200]{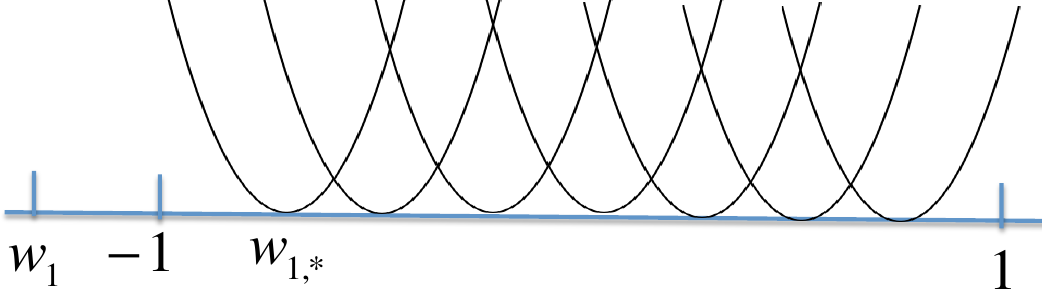}
  \caption{Simple illustration to motivate the fast initial behavior of the SG method for minimizing empirical risk \eqref{def.f}, where each $f_i$ is a convex quadratic. This example is adapted from \cite{Bert15}.}
  \label{confusion}
\efigure

Despite the issues illustrated by this example, we shall see in~\S\ref{sec.sg} that one can nevertheless ensure convergence by employing a sequence of diminishing stepsizes to overcome any oscillatory behavior of the algorithm.

\paragraph{\textbf{Theoretical Motivation}}

One can also cite theoretical arguments for a preference of SG over a batch approach.  Let us give a preview of these arguments now, which are studied in more depth and further detail in \S\ref{sec.sg}.

\bitemize
  \item It is well known that a batch approach can minimize $R_n$ at a fast rate; e.g., if~$R_n$ is strongly convex (see Assumption~\ref{ass.sc}) and one applies a batch gradient method, then there exists a constant $\rho \in (0,1)$ such that, for all $k \in \N{}$, the \emph{training error} satisfies
\bequation\label{eq.new_linear}
  R_n(w_k) - R_n^* \leq \Ocal(\rho^k),
\eequation
where $R_n^*$ denotes the minimal value of $R_n$.  The rate of convergence exhibited here is refereed to as R-linear convergence in the optimization literature \cite{OrteRhei00} and geometric convergence in the machine learning research community; we shall simply refer to it as \emph{linear convergence}.  From \eqref{eq.new_linear}, one can conclude that, in the worst case, the total number of iterations in which the training error can be above a given $\epsilon > 0$ is proportional to $\log(1/\epsilon)$.  This means that, with a per-iteration cost proportional to $n$ (due to the need to compute $\nabla R_n(w_k)$ for all $k \in \N{}$), the total work required to obtain $\epsilon$-optimality for a batch gradient method is proportional to $n\log(1/\epsilon)$.
  \item The rate of convergence of a basic stochastic method is slower than for a batch gradient method; e.g., if $R_n$ is strictly convex and each $i_k$ is drawn uniformly from $\{1,\dots,n\}$, then, for all $k \in \N{}$, the SG iterates defined by \eqref{eq.sg_init} satisfy the \emph{sublinear convergence} property (see Theorem~\ref{th.sg_sc})
  \bequation\label{eq.sg_for_empirical_risk}
    \E[R_n(w_k) - R_n^*] = \Ocal(1/k).
  \eequation
  However, it is crucial to note that \emph{neither the per-iteration cost nor the right-hand side of \eqref{eq.sg_for_empirical_risk} depends on the sample set size $n$.}  This means that the total work required to obtain $\epsilon$-optimality for SG is proportional to $1/\epsilon$.  Admittedly, this can be larger than $n\log(1/\epsilon)$ for moderate values of $n$ and~$\epsilon$, but, as discussed in detail in \S\ref{sec.work_complexity}, the comparison favors SG when one moves to the \emph{big data} regime where $n$ is large and one is merely limited by a computational time budget.
  \item Another important feature of SG is that, in a stochastic optimization setting, it yields the same convergence rate as in \eqref{eq.sg_for_empirical_risk} for the error in expected risk, $R - R^*$, where $R^*$ is the minimal value of $R$.  Specifically, by applying the SG iteration \eqref{eq.sg_init}, but with $\nabla f_{i_k}(w_k)$ replaced by $\nabla f(w_k;\xi_k)$ with each $\xi_k$ drawn independently according to the distribution $P$, one finds that
  \bequation\label{eq.sg_for_expected_risk}
    \E[R(w_k) - R^*] = \Ocal(1/k);
  \eequation
  again a sublinear rate, but on the expected risk.  Moreover, in this context, a batch approach is not even viable without the ability to compute $\nabla R$.  Of course, this represents a different setting than one in which only a finite training set is available, but it reveals that if $n$ is large with respect to~$k$, then the behavior of SG in terms of minimizing the empirical risk $R_n$ or the expected risk $R$ is practically indistinguishable up to iteration $k$.  This property cannot be claimed by a batch method.
\eitemize

In summary, there are intuitive, practical, and theoretical arguments in favor of stochastic over batch approaches in optimization methods for large-scale machine learning.  For these reasons, and since SG is used so pervasively by practitioners, we frame our discussions about optimization methods in the context of their relationship with SG.  We do not claim, however, that batch methods have no place in practice.  For one thing, if Figure~\ref{wins} were to consider a larger number of epochs, then one would see the batch approach eventually overtake the stochastic method and yield a lower training error.  
This motivates why many recently proposed methods try to combine the best properties of batch and stochastic algorithms. Moreover, the SG iteration is difficult to parallelize and requires excessive communication between nodes in a distributed computing setting, providing further impetus for the design of new and improved optimization algorithms.

\subsection{Beyond SG: Noise Reduction and Second-Order Methods}\label{sec.beyond_sg}

Looking forward, one of the main questions being asked by researchers and practitioners alike is: what lies beyond SG that can serve as an efficient, reliable, and easy-to-use optimization method for the kinds of applications discussed in~\S\ref{sec.case_studies}?

To answer this question, we depict in Figure~\ref{view1} methods that aim to improve upon SG as lying on a two-dimensional plane.  At the origin of this organizational scheme is SG, representing the base from which all other methods may be compared.

\bfigure[ht]
  \center
  \includegraphics[width=0.6\linewidth,clip=true,trim=15 55 15 55]{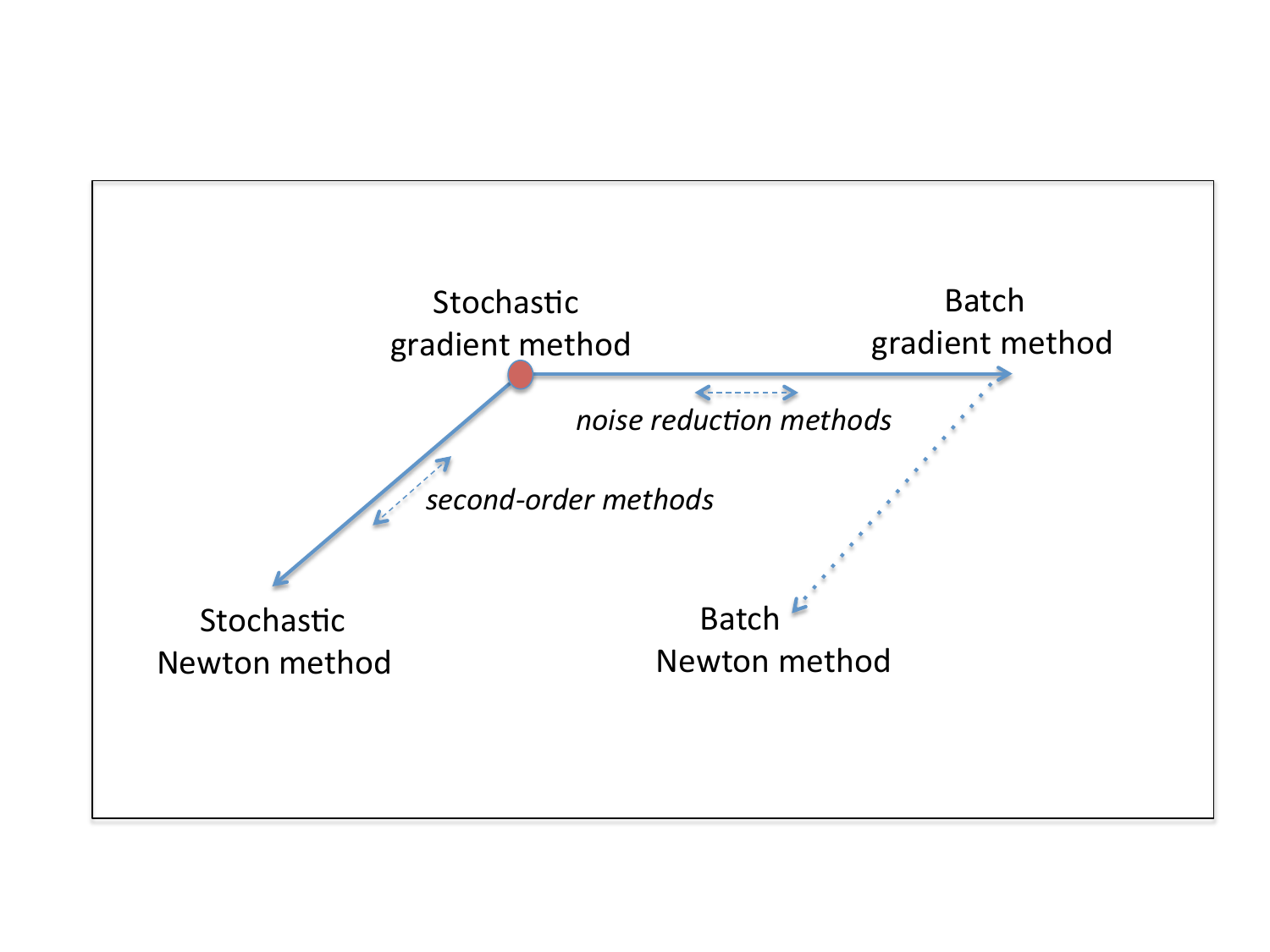}
  \caption{Schematic of a two-dimensional spectrum of optimization methods for machine learning. The horizontal axis represents methods designed to control stochastic noise; the second axis, methods that deal with ill conditioning.}
  \label{view1}
\efigure

From the origin along the horizontal access, we place methods that are neither purely stochastic nor purely batch, but attempt to combine the best properties of both approaches.  For example, observing the iteration~\eqref{eq.sg_init}, one quickly realizes that there is no particular reason to employ information from only one sample point per iteration.  Instead, one can employ a \emph{mini-batch} approach in which a small subset of samples, call it $\Scal_k \subseteq \{1,\dots,n\}$, is chosen randomly in each iteration, leading to
\bequation\label{eq.mini-batch}
  w_{k+1} \gets w_k - \frac{\alpha_k}{|\Scal_k|} \sum_{i \in \Scal_k} \nabla f_i(w_k).
\eequation
Such an approach falls under the framework set out by Robbins and Monro~\cite{RobbMonr51}, and allows some degree of parallelization to be exploited in the computation of mini-batch gradients.  In addition, one often finds that, due to the reduced variance of the stochastic gradient estimates, the method is easier to tune in terms of choosing the stepsizes $\{\alpha_k\}$.  Such a mini-batch SG method has been widely used in practice.

Along this horizontal axis, one finds other methods as well.  In our investigation, we classify two main groups as \emph{dynamic sample size} and \emph{gradient aggregation} methods, both of which aim to improve the rate of convergence from sublinear to linear.  These methods do not simply compute mini-batches of fixed size, nor do they compute full gradients in every iteration.  Instead, they dynamically replace or incorporate new gradient information in order to construct a more reliable step with smaller variance than an SG step.  For this reason, we refer to the methods along the horizontal axis as \emph{noise reduction methods}.  We discuss methods of this type in \S\ref{sec.noise_reduction}.  

Along the second axis in Figure~\ref{view1} are algorithms that, in a broad sense, attempt to overcome the adverse effects of high nonlinearity and ill-conditioning.  For such algorithms, we use the term \emph{second-order methods}, which encompasses a variety of strategies; see \S\ref{sec.second_order}.  We discuss well known inexact Newton and quasi-Newton methods, as well as (generalized) Gauss-Newton methods~\cite{Bert96,Schr01}, the natural gradient method~\cite{Amar98}, and scaled gradient iterations~\cite{TielHint12,DuchHazaSing11}.

We caution that the schematic representation of methods presented in Figure~\ref{view1} should not be taken too literally since it is not possible to truly organize algorithms so simply, or to include all methods along only two such axes.  For example, one could argue that iterate averaging methods do not neatly belong in the category of second-order methods, even though we place them there, and one could argue that gradient methods with momentum \cite{Poly64} or acceleration \cite{Nest83,Nest04} do belong in this category, even though we discuss them separately in \S\ref{sec.other}.  Nevertheless, Figure~\ref{view1} provides a useful road map as we describe and analyze a large collection of optimization methods of various forms and characteristics.  Moreover, our two-dimensional roadmap is useful in that it suggests that optimization methods do not need to exist along the coordinate axes only; e.g., a batch Newton method is placed at the lower-right corner, and one may consider various combinations of second-order and noise reduction schemes.

\section{Analyses of Stochastic Gradient Methods}\label{sec.sg}
\setcounter{equation}{0}
\setcounter{theorem}{0}
\setcounter{algorithm}{0}
\setcounter{figure}{0}
\setcounter{table}{0}

In this section, we provide insights into the behavior of a stochastic gradient method (SG) by establishing its convergence properties and worst-case iteration complexity bounds.  A preview of such properties were given in \eqref{eq.sg_for_empirical_risk}--\eqref{eq.sg_for_expected_risk}, but now we prove these and other interesting results in detail, all within the context of a generalized SG algorithm.  We start by analyzing our SG algorithm when it is invoked to minimize a strongly convex objective function, where it is possible to establish a global rate of convergence to the optimal objective value.  This is followed by analyses when our SG algorithm is employed to minimize a generic nonconvex objective.  To emphasize the generality of the results proved in this section, we remark that the objective function under consideration could be the expected risk~\eqref{def.F} or empirical risk~\eqref{def.f}; i.e., we refer to the objective function
$F : \R{d} \to \R{}$, which represents either
\bequation\label{def.F_general}
  F(w) = \left\{\baligned \textcolor{blue}{R(w)}\ &\textcolor{blue}{= \E[f(w;\xi)]} \\ &\text{or} \\ \textcolor{red}{R_n(w)}\ &\textcolor{red}{= \frac{1}{n} \sum_{i=1}^n f_i(w)}. \ealigned\right. 
\eequation
Our analyses apply equally to both objectives; the only difference lies in the way that one picks the stochastic gradient estimates in the method.\footnote{Picking samples uniformly from a finite training set, replacing them into the set for each iteration, corresponds to sampling from a discrete distribution giving equal weight to every sample.  In this case, the SG algorithm in this section optimizes the empirical risk $F = R_n$.  Alternatively, picking samples in each iteration according to the distribution~$P$, the SG algorithm optimizes the expected risk $F = R$.  One could also imagine picking samples \emph{without replacement} until one exhausts a finite training set.  In this case, the SG algorithm here can be viewed as optimizing either $R_n$ or $R$, but only until the training set is exhausted.  After that point, our analyses no longer apply.  Generally speaking, the analyses of such \emph{incremental algorithms} often requires specialized techniques \cite{Bert15,gurbuzbalaban2015convergence}.}

We define our generalized SG method as Algorithm~\ref{alg.sg}.  The algorithm merely presumes that three computational tools exist: $(i)$ a mechanism for generating a realization of a random variable $\xi_k$ (with $\{\xi_k\}$ representing a sequence of jointly independent random variables); $(ii)$ given an iterate $w_k \in \R{d}$ and the realization of~$\xi_k$, a mechanism for computing a stochastic vector $g(w_k,\xi_k) \in \R{d}$; and $(iii)$ given an iteration number $k \in \N{}$, a mechanism for computing a scalar stepsize~$\alpha_k>0$.

\balgorithm[ht]
  \caption{Stochastic Gradient (SG) Method}
  \label{alg.sg}
  \balgorithmic[1]
    \State Choose an initial iterate $w_1$.
    \For{$k = 1,2,\dots$}
      \State Generate a realization of the random variable $\xi_k$.
      \State Compute a stochastic vector $g(w_k,\xi_k)$.
      \State Choose a stepsize $\alpha_k > 0$.
      \State Set the new iterate as $w_{k+1} \gets w_k - \alpha_k g(w_k,\xi_k)$. \label{step.sg.update}
    \EndFor
  \ealgorithmic
\ealgorithm

The generality of Algorithm~\ref{alg.sg} can be seen in various ways.  First, the value of the random variable $\xi_k$ need only be viewed as a seed for generating a stochastic direction; as such, a realization of it may represent the choice of a single training sample as in the simple SG method stated as~\eqref{eq.sg_init}, or may represent a set of samples as in the mini-batch SG method~\eqref{eq.mini-batch}.  Second, $g(w_k,\xi_k)$ could represent a stochastic gradient---i.e., an unbiased estimator of $\nabla F(w_k)$, as in the classical method of Robbins and Monro \cite{RobbMonr51}---or it could represent a stochastic Newton or quasi-Newton direction; see \S\ref{sec.second_order}.  That is, our analyses cover the choices
\bequation\label{brackii}
  g(w_k,\xi_k) = \left\{\baligned \displaystyle &\nabla f(w_k;\xi_k) \\ \displaystyle \frac{1}{n_k}\sum_{i=1}^{n_k} &\nabla f(w_k;\xi_{k,i}) \\ \displaystyle H_k \frac{1}{n_k} \sum_{i=1}^{n_k} &\nabla f(w_k;\xi_{k,i}), \ealigned\right. 
\eequation
where, for all $k \in \N{}$, one has flexibility in the choice of mini-batch size $n_k$ and symmetric positive definite scaling matrix $H_k$.  No matter what choice is made, we shall come to see that all of our theoretical results hold as long as the expected angle between $g(w_k,\xi_k)$ and $\nabla F(w_k)$ is sufficiently positive.  Third, Algorithm~\ref{alg.sg} allows various choices of the stepsize sequence $\{\alpha_k\}$.  Our analyses focus on two choices, one involving a fixed stepsize and one involving diminishing stepsizes, as both are interesting in theory and in practice.  Finally, we note that Algorithm~\ref{alg.sg} also covers active learning techniques in which the iterate $w_k$ influences the sample selection.\footnote{We have assumed that the elements of the random variable sequence $\{\xi_k\}$ are independent in order to avoid requiring certain machinery from the analyses of stochastic processes.  Viewing~$\xi_k$ as a seed instead of a sample during iteration $k$ makes this restriction minor.  However, it is worthwhile to mention that all of the results in this section still hold if, instead, $\{\xi_k\}$ forms an adapted (non-anticipating) stochastic process and expectations taken with respect to~$\xi_k$ are replaced by expectations taken with respect to the conditional distribution of~$\xi_k$ given $\{\xi_1,\dots,\xi_{k-1}\}$.}

Notwithstanding all of this generality, we henceforth refer to Algorithm~\ref{alg.sg} as \emph{SG}.  The particular instance \eqref{eq.sg_init} will be referred to as \emph{simple SG} or \emph{basic SG}, whereas the instance \eqref{eq.mini-batch} will be referred to as \emph{mini-batch SG}.

Beyond our convergence and complexity analyses, a complete appreciation for the properties of SG is not possible without highlighting its theoretical advantages over batch methods in terms of computational complexity.  Thus, we include in section~\S\ref{sec.work_complexity} a discussion of the trade-offs between rate of convergence and computational effort among prototypical stochastic and batch methods for large-scale learning.

\subsection{Two Fundamental Lemmas}\label{sec.sg_fundamental_lemmas}

Our approach for establishing convergence guarantees for SG is built upon an assumption of smoothness of the objective function.  (Alternative foundations are possible; see Appendix~\ref{app.convex}.)  This, and an assumption about the first and second moments of the stochastic vectors $\{g(w_k,\xi_k)\}$ lead to two fundamental lemmas from which all of our results will be derived.

Our first assumption is formally stated as the following.  Recall that, as already mentioned in \eqref{def.F_general}, $F$ can stand for either expected or empirical risk.

\bassumption[\textbf{Lipschitz-continuous objective gradients}]\label{ass.Lipschitz}
The objective function $F : \R{d}~\to~\R{}$ is continuously differentiable and the gradient function of~$F$, namely, $\nabla F : \R{d} \to \R{d}$, is Lipschitz continuous with Lipschitz constant $L > 0$, i.e.,
  \bequationn
    \|\nabla F(w) - \nabla F(\wbar)\|_2 \leq L \|w - \wbar\|_2\ \ \text{for all}\ \ \{w,\wbar\} \subset \R{d}.
  \eequationn
\eassumption

Intuitively, Assumption~\ref{ass.Lipschitz} ensures that the gradient of $F$ does not change arbitrarily quickly with respect to the parameter vector.  Such an assumption is essential for convergence analyses of most gradient-based methods; without it, the gradient would not provide a good indicator for how far to move to decrease $F$. An important consequence of Assumption~\ref{ass.Lipschitz} is that
\bequation\label{eq.lipschitz_bound}
   F(w) \leq F(\wbar) + \nabla F(\wbar)^T (w-\wbar) + \thalf L \|w-\wbar\|_2^2 \ \ \text{for all}\ \ \{w,\wbar\} \subset \R{d}.
   \eequation
This inequality is proved in Appendix~\ref{sec.proofs}, but note that it also follows immediately if~$F$ is twice continuously differentiable and the Hessian function $\nabla^2 F : \R{d} \to \R{d \times d}$ satisfies $\|\nabla^2 F(w)\|_2 \leq L$ for all $w \in \R{d}$.

Under Assumption~\ref{ass.Lipschitz} alone, we obtain the following lemma.  In the result, we use $\E_{\xi_k}[\cdot]$ to denote an expected value taken with respect to the distribution of the random variable $\xi_k$ given $w_k$.  Therefore, $\E_{\xi_k}[F(w_{k+1})]$ is a meaningful quantity since $w_{k+1}$ depends on $\xi_k$ through the update in Step~\ref{step.sg.update} of Algorithm~\ref{alg.sg}.

\blemma\label{lem.Lipschitz}
  Under Assumption~\ref{ass.Lipschitz}, the iterates of SG (Algorithm~\ref{alg.sg}) satisfy the following inequality for all $k\in\N{}$: 
  \bequation\label{eq.Lipschitz}
    \E_{\xi_k}[F(w_{k+1})] - F(w_k) \leq -\alpha_k\nabla F(w_k)^T\E_{\xi_k}[g(w_k,\xi_k)] + \thalf \alpha_k^2 L \E_{\xi_k}[\|g(w_k,\xi_k)\|_2^2].
  \eequation
\elemma
\bproof
  By Assumption~\ref{ass.Lipschitz}, the iterates generated by SG satisfy
  \bequalin
    F(w_{k+1}) - F(w_k)
      &\leq \nabla F(w_k)^T(w_{k+1} - w_k) + \thalf L\|w_{k+1} - w_k\|_2^2 \\
      &\leq -\alpha_k\nabla F(w_k)^Tg(w_k,\xi_k) + \thalf \alpha_k^2 L \|g(w_k,\xi_k)\|_2^2.
  \eequalin
  Taking expectations in these inequalities with respect to the distribution of $\xi_k$, and noting that $w_{k+1}$---but not $w_k$---depends on $\xi_k$, we obtain the desired bound. 
\eproof

This lemma shows that, regardless of how SG arrived at $w_k$, the expected decrease in the objective function yielded by the $k$th step is bounded above by a quantity involving: $(i)$ the expected directional derivative of $F$ at $w_k$ along $-g(x_k,\xi_k)$ and $(ii)$~the second moment of $g(x_k,\xi_k)$.  For example, if $g(w_k,\xi_k)$ is an unbiased estimate of $\nabla F(w_k)$, then it follows from Lemma~\ref{lem.Lipschitz} that
\bequation\label{orchid}
  \E_{\xi_k}[F(w_{k+1})] - F(w_k) \leq -\alpha_k \| \nabla F(w_k)\|_2^2 + \thalf \alpha_k^2 L \E_{\xi_k}[\|g(w_k,\xi_k)\|_2^2].
\eequation
We shall see that convergence of SG is guaranteed as long as the stochastic directions and stepsizes are chosen such that the right-hand side of \eqref{eq.Lipschitz} is bounded above by a \emph{deterministic} quantity that asymptotically ensures sufficient descent in $F$.  One can ensure this in part by stating additional requirements on the first and second moments of the stochastic directions $\{g(w_k,\xi_k)\}$. In particular, in order to limit the harmful effect of the last term in~\eqref{orchid}, we restrict the variance of $g(w_k,\xi_k)$, i.e.,
\bequation\label{eq.variance} 
  \var_{\xi_k}[g(w_k,\xi_k)] := \E_{\xi_k}[\|g(w_k,\xi_k)\|_2^2] - \|\E_{\xi_k}[g(w_k,\xi_k)]\|_2^2\,.
\eequation

\bassumption[\textbf{First and second moment limits}]\label{ass.sg}
  The objective function and SG (Algorithm~\ref{alg.sg}) satisfy the following:
  \benumerate
    \item[(a)] The sequence of iterates $\{w_k\}$ is contained in an open set over which $F$ is bounded below by a scalar $F_{\inf}$.
    \item[(b)] There exist scalars $\mu_G \geq \mu > 0$ such that, for all $k\in\N{}$,
      \bsubequations
        \begin{align}
          \nabla F(w_k)^T\E_{\xi_k}[g(w_k,\xi_k)] &\geq \mu \|\nabla F(w_k)\|^2_2\ \ \text{and} \label{eq.descent_condition} \\
          \|\E_{\xi_k}[g(w_k,\xi_k)]\|_2 &\leq \mu_G \|\nabla F(w_k)\|_2. \label{eq.mean_condition}
        \end{align}
    \esubequations
    \item[(c)] There exist scalars $M \geq 0$ and $M_{V} \geq 0$ such that, for all $k\in\N{}$,
    \bequation\label{eq.variance_bound}
      \var_{\xi_k}[g(w_k,\xi_k)] \leq M + M_{V} \|\nabla F(w_k)\|_2^2.
    \eequation
  \eenumerate
\eassumption

The first condition, Assumption~\ref{ass.sg}$(a)$, merely requires the objective function to be bounded below over the region explored by the algorithm.  The second requirement, Assumption~\ref{ass.sg}$(b)$, states that, in expectation, the vector $-g(w_k,\xi_k)$ is a direction of sufficient descent for $F$ from $w_k$ with a norm comparable to the norm of the gradient.  The properties in this requirement hold immediately with $\mu_G=\mu=1$ if $g(w_k,\xi_k)$ is an unbiased estimate of $\nabla F(w_k)$, and are maintained if such an unbiased estimate is multiplied by a positive definite matrix $H_k$ that is conditionally uncorrelated with $g(w_k,\xi_k)$ given $w_k$ and whose eigenvalues lie in a fixed positive interval for all $k \in \N{}$.  The third requirement, Assumption~\ref{ass.sg}$(c)$, states that the variance of $g(w_k,\xi_k)$ is restricted, but in a relatively minor manner.  For example, if~$F$ is a convex quadratic function, then the variance is allowed to be nonzero at any stationary point for $F$ and is allowed to grow quadratically in any direction.

All together, Assumption~\ref{ass.sg}, combined with the definition~\eqref{eq.variance}, requires that the second moment of $g(w_k,\xi_k)$ satisfies
\bequation\label{eq.second_moment}
  \E_{\xi_k}[\|g(w_k,\xi_k)\|_2^2] \leq M + M_G \|\nabla F(w_k)\|_2^2\ \ \text{with}\ \ M_G := M_V + \mu_G^2 \geq \mu^2 > 0.
\eequation
In fact, all of our analyses in this section hold if this bound on the second moment were to be assumed directly.  (We have stated Assumption~\ref{ass.sg} in the form above merely to facilitate our discussion in \S\ref{sec.noise_reduction}.)

The following lemma builds on Lemma~\ref{lem.Lipschitz} under the
additional conditions now set forth in Assumption~\ref{ass.sg}.

\blemma\label{lem.expected_decrease}
  Under Assumptions~\ref{ass.Lipschitz} and~\ref{ass.sg}, the iterates of SG (Algorithm~\ref{alg.sg}) satisfy the following inequalities for all $k\in\N{}$: 
  \bsubequations
    \begin{align}
      \E_{\xi_k}[F(w_{k+1})] - F(w_k)
        &\leq - \mu \alpha_k \|\nabla F(w_k)\|_2^2 + \thalf \alpha_k^2 L \E_{\xi_k}[\|g(w_k,\xi_k)\|_2^2] \label{eq.expected_decrease_pre} \\
        &\leq - (\mu - \thalf\alpha_kLM_G)\alpha_k \|\nabla F(w_k)\|_2^2 + \thalf \alpha_k^2 L M. \label{eq.expected_decrease}
    \end{align}
  \esubequations
\elemma
\bproof
  By Lemma~\ref{lem.Lipschitz} and \eqref{eq.descent_condition}, it follows that
  \bequalin
    \E_{\xi_k}[F(w_{k+1})] - F(w_k)
      &\leq -\alpha_k \nabla F(w_k)^T\E_{\xi_k}[g(w_k,\xi_k)] + \thalf \alpha_k^2 L \E_{\xi_k}[\|g(w_k,\xi_k)\|_2^2] \nonumber \\
      &\leq - \mu \alpha_k \|\nabla F(w_k)\|_2^2 + \thalf \alpha_k^2 L \E_{\xi_k}[\|g(w_k,\xi_k)\|_2^2], \\
  \eequalin
  which is \eqref{eq.expected_decrease_pre}. Assumption~\ref{ass.sg}, giving \eqref{eq.second_moment}, then yields \eqref{eq.expected_decrease}.
\eproof

As mentioned, this lemma reveals that regardless of how the method arrived at the iterate~$w_k$, the optimization process continues in a \emph{Markovian} manner in the sense that $w_{k+1}$ is a random variable that depends only on the iterate $w_k$, the seed~$\xi_k$, and the stepsize $\alpha_k$ \emph{and not on any past iterates}.  This can be seen in the fact that the difference $\E_{\xi_k}[F(w_{k+1})] - F(w_k)$ is bounded above by a deterministic quantity.  Note also that the first term in  \eqref{eq.expected_decrease} is strictly negative for small~$\alpha_k$ and suggests a decrease in the objective function by a magnitude proportional to $\|\nabla F(w_k)\|_2^2$.  However, the second term in \eqref{eq.expected_decrease} could be large enough to allow the objective value to increase.  Balancing these terms is critical in the design of SG methods.

\subsection{SG for Strongly Convex Objectives}\label{sec.sg_sc}

The most benign setting for analyzing the SG method is in the context of minimizing a strongly convex objective function.  For the reasons described in Inset~\ref{mini.aside}, when not considering a generic nonconvex objective $F$, we focus on the strongly convex case and only briefly mention the (not strongly) convex case in certain occasions.

\renewcommand{\figurename}{Inset}
\setcounter{figtemp}{\value{figure}}
\setcounter{figure}{1}
\begin{figure}[ht]
\begin{adjustbox}{minipage=\textwidth-20pt,margin=8pt 8pt 8pt -4pt,bgcolor=blue!5,frame=2pt}
  \caption{Perspectives on SG Analyses}
  \label{mini.aside}
  \medskip
  \setlength\parindent{.25in}

  All of the convergence rate and complexity results presented in this paper relate to the minimizaton of \emph{strongly convex} functions.  This is in contrast with a large portion of the literature on optimization methods for machine learning, in which much effort is placed to strengthen convergence guarantees for methods applied to functions that are convex, but not strongly convex.  We have made this choice for a few reasons.  First, it leads to a focus on results that are relevant to actual machine learning practice, since in many situations when a convex model is employed---such as in logistic regression---it is often regularized by a strongly convex function to facilitate the solution process.  Second, there exist a variety of situations in which the objective function is not globally (strongly) convex, but is so in the neighborhood of local minimizers, meaning that our results can represent the behavior of the algorithm in such regions of the search space.  Third, one can argue that related results when minimizing non-strongly convex models can be derived as extensions of the results presented here \cite{HazanReductions16}, making our analyses a starting point for deriving a more general theory.
    
  We have also taken a pragmatic approach in the types of convergence guarantees that we provide.  In particular, in our analyses, we focus on results that reveal the properties of SG iterates \emph{in expectation}.  The stochastic approximation literature, on the other hand, often relies on martingale techniques to establish \emph{almost sure convergence} \cite{Glad65,RobbSieg71} under the same assumptions~\cite{Bott98}.  For our purposes, we omit these complications since, in our view, they do not provide significant additional insights into the forces driving convergence of the method.
\end{adjustbox}
\end{figure}
\renewcommand{\figurename}{Fig.}
\setcounter{figure}{\value{figtemp}}

We formalize a strong convexity assumption as the following.

\bassumption[\textbf{Strong convexity}]\label{ass.sc}
  The objective function $F : \R{d} \to \R{}$ is strongly convex in that there exists a constant $c > 0$ such that
  \bequation\label{eq.sc}
     F(\wbar) \geq F(w) + \nabla F(w)^T(\wbar-w)
         + \thalf c \|\wbar-w\|_2^2\ \ \text{for all}\ \ (\wbar,w) \in \R{d} \times \R{d}.
  \eequation
  Hence, $F$ has a unique minimizer, denoted as $w_* \in \R{d}$ with $F_* := F(w_*)$.
\eassumption

A useful fact from convex analysis (proved in Appendix~\ref{sec.proofs}) is that, under Assumption~\ref{ass.sc}, one can bound the optimality gap at a given point in terms of the squared $\ell_2$-norm of the gradient of the objective at that point:
\bequation\label{eq.sc2}
  2c(F(w) - F_*) \leq \|\nabla F(w)\|_2^2\ \ \text{for all}\ \ w \in \R{d}.
\eequation
We use this inequality in several proofs.  We also observe that, from \eqref{eq.lipschitz_bound} and \eqref{eq.sc}, the constants in Assumptions~\ref{ass.Lipschitz} and \ref{ass.sc} must satisfy $c \leq L$.

We now state our first convergence theorem for SG, describing its behavior when minimizing a strongly convex objective function when employing a fixed stepsize.  In this case, it will not be possible to prove convergence to the solution, but only to a neighborhood of the optimal value.  (Intuitively, this limitation should be clear from \eqref{eq.expected_decrease} since the first term on the right-hand side decreases in magnitude as the solution is approached---i.e., as $\nabla F(w_k)$ tends to zero---but the last term remains constant.  Thus, after some point, a reduction in the objective cannot be expected.)  We use $\E[\cdot]$ to denote an expected value taken with respect to the joint distribution of all random variables.  For example, since $w_k$ is completely determined by the realizations of the independent random variables $\{\xi_1,\xi_2,\dots,\xi_{k-1}\}$, the \emph{total expectation} of $F(w_k)$ for any $k \in \N{}$ can be taken as
\bequationn
  \E[F(w_k)] = \E_{\xi_1} \E_{\xi_2} \dots \E_{\xi_{k-1}} [F(w_k)].
\eequationn
The theorem shows that if the stepsize is not too large, then, in expectation, the sequence of function values $\{F(w_k)\}$ converges near the optimal value. 

\btheorem[\textbf{Strongly Convex Objective, Fixed Stepsize}]\label{th.sg_sc_fixed}
  Under Assumptions~\ref{ass.Lipschitz}, \ref{ass.sg}, and~\ref{ass.sc} (with $F_{\inf}=F_*$), suppose that the SG method (Algorithm~\ref{alg.sg}) is run with a fixed stepsize, $\alpha_k = \bar \alpha$ for all $k\in\N{}$, satisfying
  \bequation\label{eq.alpha_fixed}
    0 < \bar \alpha \leq \frac{\mu}{LM_G}.
  \eequation
  Then, the expected optimality gap satisfies the following inequality for all $k \in \N{}$\,:
  \bequation\label{uxmal}
    \baligned
    \E[F(w_{k}) - F_*] \leq&\ \frac{\bar \alpha LM}{2c\mu} + \(1 - \bar \alpha c\mu\)^{k-1} \(F(w_1) - F_* - \frac{\bar \alpha LM}{2c\mu}\) \\
    \xrightarrow{k\rightarrow\infty} &\ \frac{\bar \alpha LM}{2c\mu}.
    \ealigned
  \eequation
\etheorem

\bproof 
  Using Lemma~\ref{lem.expected_decrease} with~\eqref{eq.alpha_fixed} and \eqref{eq.sc2}, we have for all $k\in\N{}$ that
  \bequalin
    \E_{\xi_k}[F(w_{k+1})] - F(w_k)]
      &\leq -(\mu - \thalf \bar \alpha LM_G) \bar \alpha\|\nabla F(w_k)\|_2^2 + \thalf \bar \alpha^2 L M \\
      &\leq -\thalf\bar \alpha\mu\|\nabla F(w_k)\|_2^2 + \thalf \bar \alpha^2 L M \\
      &\leq -\bar \alpha c \mu (F(w_k) - F_*) + \thalf \bar \alpha^2 L M.
  \eequalin
  Subtracting $F_*$ from both sides, taking total expectations, and rearranging, this yields
  \bequationn
    \E[F(w_{k+1}) - F_*] \leq (1 - \bar \alpha c\mu)\E[F(w_k) - F_*] + \thalf\bar \alpha^2LM.
  \eequationn
  Subtracting the constant $\bar \alpha LM/(2c\mu)$ from both sides, one obtains
  \begin{align}
    \E[F(w_{k+1}) - F_*] - \frac{\bar \alpha LM}{2c\mu}
      &\leq \(1 - \bar \alpha c\mu\) \E[F(w_k) - F_*] + \frac{\bar \alpha^2 LM}{2} - \frac{\bar \alpha LM}{2c\mu} \nonumber\\
      &=    \(1 - \bar \alpha c\mu\) \(\E[F(w_k) - F_*] - \frac{\bar \alpha LM}{2c\mu}\). \label{eq.contract_fixed}
  \end{align}
  Observe that~\eqref{eq.contract_fixed} is a contraction inequality since, by \eqref{eq.alpha_fixed} and \eqref{eq.second_moment},
  \bequation\label{mgu}
    0 < \bar\alpha c\mu \leq \frac{c\mu^2}{L M_G} \leq \frac{c\mu^2}{L\mu^2} = \frac{c}{L} \leq 1.
  \eequation
  The result thus follows by applying \eqref{eq.contract_fixed} repeatedly through iteration $k \in \N{}$.
\eproof

If $g(w_k,\xi_k)$ is an unbiased estimate of $\nabla F(w_k)$, then $\mu=1$, and if there is no noise in $g(w_k,\xi_k)$, then we may presume that $M_G=1$ (due to \eqref{eq.second_moment}).  In this case, \eqref{eq.alpha_fixed} reduces to $\bar\alpha \in (0,1/L]$, a classical stepsize requirement of interest for a steepest descent method.

Theorem~\ref{th.sg_sc_fixed} illustrates the interplay between the stepsizes and bound on the variance of the stochastic directions.  If there were no noise in the gradient computation or if noise were to decay with $\|\nabla F(w_k)\|_2^2$ (i.e., if $M=0$ in~\eqref{eq.variance_bound} and~\eqref{eq.second_moment}), then one can obtain linear convergence to the optimal value.  This is a standard result for the full gradient method with a sufficiently small positive stepsize.  On the other hand, when the gradient computation is noisy, one clearly loses this property.  One can still use a fixed stepsize and be sure that the expected objective values will converge linearly to a neighborhood of the optimal value, but, after some point, the noise in the gradient estimates prevent further progress; recall Example~\ref{ex.region_of_confusion}.  It is apparent from \eqref{uxmal} that selecting a smaller stepsize worsens the contraction constant in the convergence rate, but allows one to arrive closer to the optimal value.

\begin{figure}[ht]
\center
\includegraphics[width=0.75\linewidth]{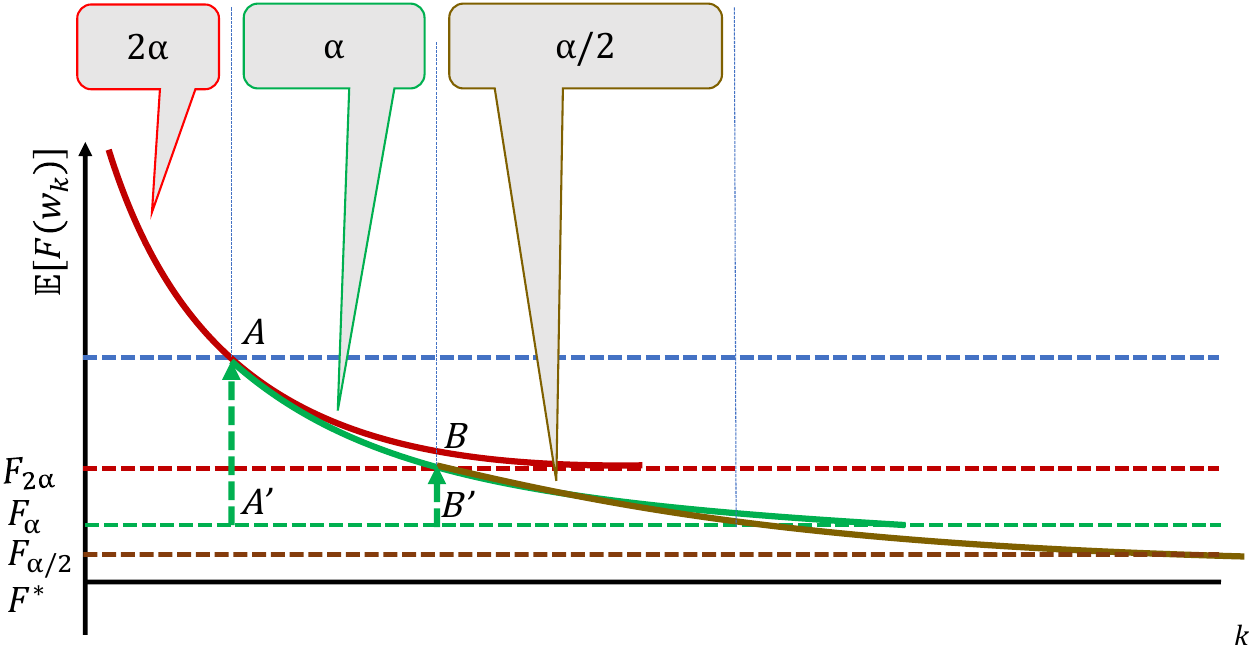}
\caption{\label{fig.sec4.decrate}\relax
Depiction of the strategy of halving the stepsize $\alpha$ when the expected suboptimality gap is smaller than twice the asymptotic limit $F_\alpha$.  In the figure, the segment $B$--$B'$ has one third of the length of $A$--$A'$.  This is the amount of decrease that must be made in the exponential term in \eqref{uxmal} by raising the contraction factor to the power of the number of steps during which one maintains a given constant stepsize; see~\eqref{eq.third}.  Since the contraction factor is $(1 - \alpha c \mu)$, the number of steps must be proportional to $\alpha$. Therefore, whenever the stepsize is halved, one must maintain it twice as long.  Overall, doubling the number of iterations halves the suboptimality gap each time, yielding an effective rate of $\Ocal(1/k)$.}
\end{figure}

These observations provide a foundation for a strategy often employed in practice by which SG is run with a fixed stepsize, and, if progress appears to stall, a smaller stepsize is selected and the process is repeated.  A straightforward instance of such an approach can be motivated with the following sketch.  Suppose that $\alpha_1 \in (0,\frac{\mu}{LM_G}]$ is chosen as in \eqref{eq.alpha_fixed} and the SG method is run with this stepsize from iteration $k_1 = 1$ until iteration $k_2$, where $w_{k_2}$ is the first iterate at which the expected suboptimality gap is smaller than twice the asymptotic value in \eqref{uxmal}, i.e., $\E[F(w_{k_2})-F_*] \leq 2F_{\alpha_1}$, where $F_{\alpha} := \frac{\alpha LM}{2c\mu}$.  Suppose further that, at this point, the stepsize is halved and the process is repeated; see Figure~\ref{fig.sec4.decrate}.   This leads to the stepsize schedule $\{\alpha_{r+1}\} = \{\alpha_1 2^{-r}\}$, index sequence $\{k_r\}$, and bound sequence $\{F_{\alpha_r}\} = \{\frac{\alpha_r LM}{2c\mu}\} \searrow 0$ such that, for all $r \in \{2,3,\dots\}$,
\bequation\label{eq.halving}
  \E[F(w_{k_{r+1}})-F_*] \leq 2F_{\alpha_r}\ \ \text{where}\ \ \E[F(w_{k_r})-F_*] \approx 2F_{\alpha_{r-1}} = 4F_{\alpha_r}.
\eequation
In this manner, the expected suboptimality gap converges to zero.

However, this does \emph{not} occur 
by halving the stepsize in every iteration, but only once the gap itself has been cut in half from a previous threshold.  To see what is the appropriate \emph{effective rate} of stepsize decrease, we may invoke Theorem~\ref{th.sg_sc_fixed}, from which it follows that to achieve the first bound in \eqref{eq.halving} one needs
\bequation\label{eq.third}
  \begin{aligned}
  & (1-\alpha_rc\mu)^{(k_{r+1}-k_r)} (4F_{\alpha_r}-F_{\alpha_r}) \leq F_{\alpha_r} \\
  \implies & k_{r+1}-k_r \geq \frac{\log(1/3)}{\log(1-\alpha_rc\mu)} \approx \frac{\log(3)}{\alpha_rc\mu} = \Ocal(2^r).
  \end{aligned}
\eequation
In other words, each time the stepsize is cut in half, double the number of iterations are required.  This is a \emph{sublinear} rate of stepsize decrease---e.g., if $\{k_r\} = \{2^{r-1}\}$, then $\alpha_k = \alpha_1/k$ for all $k \in \{2^r\}$---which, from $\{F_{\alpha_r}\} = \{\frac{\alpha_r LM}{2c\mu}\}$ and \eqref{eq.halving}, means that a sublinear convergence rate of the suboptimality gap is achieved.

  
In fact, these conclusions can be obtained in a more rigorous manner that also allows more flexibility in the choice of stepsize sequence.  The following result harks back to the seminal work of Robbins and Monro \cite{RobbMonr51}, where the stepsize requirement takes the form
\bequation\label{eq.alphas}
  \sum_{k=1}^\infty \alpha_k = \infty\ \ \text{and}\ \ \sum_{k=1}^\infty \alpha_k^2 < \infty.
\eequation
  
\btheorem[\textbf{Strongly Convex Objective, Diminishing Stepsizes}]\label{th.sg_sc}
  Under Assumptions~\ref{ass.Lipschitz}, \ref{ass.sg}, and~\ref{ass.sc} (with $F_{\inf}=F_*$), suppose that the SG method (Algorithm~\ref{alg.sg}) is run with a stepsize sequence such that, for all $k \in \N{}$,
  \bequation\label{eq.alpha_diminishing_specific}
    \alpha_k = \frac{\beta}{\gamma + k}\ \ \text{for some}\ \ \beta > \frac{1}{c\mu}\ \ \text{and}\ \ \gamma > 0\ \ \text{such that}\ \ \alpha_1\leq\frac{\mu}{L M_G}.
  \eequation
  Then, for all $k \in \N{}$, the expected optimality gap satisfies
  \begin{equation}
    \E[F(w_k) - F_*] \leq \ \frac{\nu}{\gamma + k}, \label{sublin} 
  \end{equation}
  where
  \bequation\label{nubeta}
    \nu := \max\left\{\frac{\beta^2LM}{2(\beta c\mu-1)},(\gamma+1)(F(w_1) - F_*)\right\}.
  \eequation
\etheorem
\bproof
  By \eqref{eq.alpha_diminishing_specific}, the inequality $\alpha_k L M_G \leq \alpha_1 L M_G \leq \mu$ holds for all $k\in\N{}$.  Hence, along with Lemma~\ref{lem.expected_decrease} and \eqref{eq.sc2}, one has for all $k\in\N{}$ that
  \bequalin
    \E_{\xi_k}[F(w_{k+1})] - F(w_k)
      &\leq -(\mu - \thalf \alpha_k LM_G)\alpha_k\|\nabla F(w_k)\|_2^2 + \thalf \alpha_k^2 L M \\
      &\leq -\thalf\alpha_k\mu\|\nabla F(w_k)\|_2^2 + \thalf \alpha_k^2 L M \\
      &\leq -\alpha_k c \mu (F(w_k) - F(w_*)) + \thalf \alpha_k^2 L M.
  \eequalin
  Subtracting $F_*$ from both sides, taking total expectations, and rearranging, this yields
  \bequation\label{eq.contract}
    \E[F(w_{k+1}) - F_*] \leq (1 - \alpha_k c\mu)\E[F(w_k) - F_*] + \thalf \alpha_k^2 LM.
  \eequation
  We now prove \eqref{sublin} by induction.  First, the definition of $\nu$ ensures that it holds for $k = 1$.  Then, assuming \eqref{sublin} holds for some $k\geq1$, it follows from \eqref{eq.contract} that
  \bequalin
    \E[F(w_{k+1}) - F_*]
    &\leq \(1 - \frac{\beta c\mu}{\khat}\) \frac{\nu}{\khat} + \frac{\beta^2LM}{2\khat^2}
         \qquad \text{(with $\khat := \gamma+k$)}\\
       &= \(\frac{\khat-\beta c\mu}{\khat^2}\)\nu + \frac{\beta^2LM}{2\khat^2} \\
       &= \(\frac{\khat-1}{\khat^2}\)\nu \underbrace{- \(\frac{\beta c\mu - 1}{\khat^2}\)\nu + \frac{\beta^2LM}{2\khat^2}}_{\text{nonpositive by the definition of $\nu$}} \leq \frac{\nu}{\khat+1},
  \eequalin
  where the last inequality follows because $\khat^2 \geq (\khat+1)(\khat-1)$.
\eproof

Let us now remark on what can be learned from Theorems~\ref{th.sg_sc_fixed} and~\ref{th.sg_sc}.

\paragraph{\textbf{Role of Strong Convexity}}

Observe the crucial role played by the strong convexity parameter $c>0$, the positivity of which is needed to argue that~\eqref{eq.contract_fixed} and~\eqref{eq.contract} contract the expected optimality gap.  However, the strong convexity constant impacts the stepsizes in different ways in Theorems~\ref{th.sg_sc_fixed} and~\ref{th.sg_sc}.  In the case of constant stepsizes, the possible values of $\bar \alpha$ are constrained by the upper bound~\eqref{eq.alpha_fixed} that does not depend on $c$. In the case of diminishing stepsizes, the initial stepsize $\alpha_1$ is subject to the same upper bound \eqref{eq.alpha_diminishing_specific}, but the stepsize parameter $\beta$ must be larger than~$1/(c\mu)$.  This additional requirement is critical to ensure the $\mathcal{O}(1/k)$ convergence rate.  How critical?  Consider, e.g., \cite{NemiJudiLanShap09} in which the authors provide a simple example (with unbiased gradient estimates and $\mu=1$) involving the minimization of a deterministic quadratic function \emph{with only one optimization variable} in which $c$ is overestimated, which results in $\beta$ being underestimated.  In the example, even after $10^9$ iterations, the distance to the solution remains greater than $10^{-2}$.

\paragraph{\textbf{Role of the Initial Point}}

Also observe the role played by the initial point, which determines the initial optimality gap, namely, $F(w_1)-F_*$. When using a fixed stepsize, the initial gap appears with an exponentially decreasing factor; see~\eqref{uxmal}.  In the case of diminishing stepsizes, the gap appears prominently in the second term defining $\nu$ in~\eqref{nubeta}. However, with an appropriate initialization phase, one can easily diminish the role played by this term.\footnote{In fact, the bound \eqref{sublin} slightly overstates the asymptotic influence of the initial optimality gap. Applying Chung's lemma~\cite{Chun54} to the contraction equation \eqref{eq.contract} shows that the first term in the definition of $\nu$ effectively determines the asymptotic convergence rate of $\E[F(w_k)-F_*]$.}  For example, suppose that one begins by running SG with a fixed stepsize $\bar\alpha$ until one (approximately) obtains a point, call it~$w_1$, with $F(w_1) - F_* \leq \bar\alpha L M / (2c\mu)$.  A guarantee for this bound can be argued from \eqref{uxmal}.  Starting here with $\alpha_1 = \bar\alpha$, the choices for $\beta$ and $\gamma$ in Theorem~\ref{th.sg_sc} yield
\bequationn
  (\gamma + 1)\E[F(w_1) - F_*] \leq \beta\alpha_1^{-1} \frac{\alpha_1 LM}{2 c\mu} = \frac{\beta LM}{2 c\mu} < \frac{\beta^2 LM}{2(\beta c \mu - 1)},
\eequationn
meaning that the value for $\nu$ is dominated by the first term in~\eqref{nubeta}.

On a related note, we claim that for practical purposes the initial stepsize should be chosen as large as allowed, i.e., $\alpha_1 = \mu/(LM_G)$. Given this choice of $\alpha_1$, the best asymptotic regime with decreasing stepsizes \eqref{sublin} is achieved by making $\nu$ as small as possible. Since we have argued that only the first term matters in the definition of~$\nu$, this leads to choosing $\beta=2/(c\mu)$. Under these conditions, one has
\bequation\label{eq.nu}
  \nu = \frac{\beta^2LM}{2(\beta c\mu-1)} = \frac{2}{\mu^2}\(\frac{L}{c}\)\(\frac{M}{c}\)\,.
\eequation
We shall see the (potentially large) ratios $L/c$ and $M/c$ arise again later.

\paragraph{\textbf{Trade-Offs of (Mini-)Batching}}

As a final observation about what can be learned from Theorems~\ref{th.sg_sc_fixed} and~\ref{th.sg_sc}, let us take a moment to compare the theoretical performance of two fundamental algorithms---the simple SG iteration \eqref{eq.sg_init} and the mini-batch SG iteration \eqref{eq.mini-batch}---when these results are applied for minimizing empirical risk, i.e., when $F = R_n$.  This provides a glimpse into how such results can be used to compare algorithms in terms of their computational trade-offs.

The most elementary instance of our SG algorithm is simple SG, which, as we have seen, consists of picking a random sample index $i_k$ at each iteration and computing
\bequation\label{sgcost}
   g(w_k,\xi_k) = \nabla f_{i_k}(w_k).
\eequation
By contrast, instead of picking a single sample, mini-batch SG consists of randomly selecting a subset $\Scal_k$ of the sample indices and computing
\bequation \label{minicost}
  g(w_k,\xi_k) = \frac{1}{|\Scal_k|} \sum_{i\in \Scal_k} \nabla f_{i} (w_k) \,.
\eequation
To compare these methods, let us assume for simplicity that we employ the same number of samples in each iteration so that the mini-batches are of constant size, i.e., $|S_k|=\nmb$.  There are then two distinct regimes to consider, namely, when $\nmb \ll n$ and when $\nmb \approx n$.  Our goal here is to use the results of Theorems~\ref{th.sg_sc_fixed} and~\ref{th.sg_sc} to show that, in the former scenario, the theoretical benefit of mini-batching can appear to be somewhat ambiguous, meaning that one must leverage certain computational tools to benefit from mini-batching in practice.  As for the scenario when $\nmb \approx n$, the comparison is more complex due to a trade-off between per-iteration costs and overall convergence rate of the method (recall \S\ref{sec.motiv_sg}).  We leave a more formal treatment of this scenario, specifically with the goals of large-scale machine learning in mind, for~\S\ref{sec.work_complexity}.

Suppose then that the mini-batch size is $\nmb \ll n$.  The computation of the stochastic direction $g(w_k,\xi_k)$ in~\eqref{minicost} is clearly $\nmb$ times more expensive than in \eqref{sgcost}.  In return, the variance of the direction is reduced by a factor of $1/\nmb$.  (See~\S\ref{sec.dynamic_sample_size} for further discussion of this fact.)  That is, with respect to our analysis, the constants $M$ and $M_V$ that appear in Assumption~\ref{ass.sg} (see \eqref{eq.variance_bound}) are reduced by the same factor, becoming $M/\nmb$ and $M_V/\nmb$ for mini-batch SG.  It is natural to ask whether this reduction of the variance pays for the higher per-iteration cost.

Consider, for instance, the case of employing a sufficiently small constant stepsize $\bar\alpha > 0$.  For mini-batch SG, Theorem~\ref{th.sg_sc_fixed} leads to
 \bequationn
    \E[F(w_{k}) - F_*] \leq \frac{\bar \alpha LM}{2 c\mu\,\nmb}
  + \left[1 - \bar \alpha c\mu \right]^{k-1} \(F(w_1) - F_* - \frac{\bar \alpha LM}{2c\mu\,\nmb}\).
  \eequationn
  Using the simple SG method with stepsize $\bar \alpha/\nmb$ leads to a similar asymptotic gap:
  \bequationn
  \E[F(w_{k}) - F_*] \leq \frac{\bar \alpha LM}{2 c\mu\,\nmb}
   + \left[1 - \frac{\bar \alpha c\mu}{\nmb} \right]^{k-1} \({F(w_1)} - F_* - \frac{\bar \alpha LM}{2c\mu\,\nmb}\).
  \eequationn
The worse contraction constant (indicated using square brackets) means that one needs to run $\nmb$ times more iterations of the simple SG algorithm to obtain an equivalent optimality gap.  That said, since the computation in a simple SG iteration is $\nmb$ times cheaper, this amounts to effectively the same total computation as for the mini-batch SG method.  A similar analysis employing the result of Theorem~\ref{th.sg_sc} can be performed when decreasing stepsizes are used.

These observations suggest that the methods can be comparable.  However, an important consideration remains.  In particular, the convergence theorems require that the initial stepsize be smaller than $\mu/(L M_G)$.  Since~\eqref{eq.second_moment} shows that $M_G \geq \mu^2$, the largest this stepsize can be is $1/(\mu L)$.  Therefore, one cannot simply assume that the mini-batch SG method is allowed to employ a stepsize that is $\nmb$ times larger than the one used by SG.  In other words, one cannot always compensate for the higher per-iteration cost of a mini-batch SG method by selecting a larger stepsize.

One can, however, realize benefits of mini-batching in practice since it offers important opportunities for software optimization and parallelization; e.g., using sizeable mini-batches is often the only way to fully leverage a GPU processor.  Dynamic mini-batch sizes can also be used as a substitute for decreasing stepsizes; see \S\ref{sec.dynamic_sample_size}.

\subsection{SG for General Objectives}\label{sec.sg_nc}

As mentioned in our case study of deep neural networks in \S\ref{sec.deep_neural_nets}, many important machine learning models lead to nonconvex optimization problems, which are currently having a profound impact in practice.  Analyzing the SG method when minimizing nonconvex objectives is more challenging than in the convex case since such functions may possess multiple local minima and other stationary points.  Still, we show in this subsection that one can provide meaningful guarantees for the SG method in nonconvex settings.

Paralleling \S\ref{sec.sg_sc}, we present two results---one for employing a fixed positive stepsize and one for diminishing stepsizes.  We maintain the same assumptions about the stochastic directions $g(w_k, \xi_k)$, but do not assume convexity of $F$.  As before, the results in this section still apply to a wide class of methods since $g(w_k, \xi_k)$ could be defined as a (mini-batch) stochastic gradient or a Newton-like direction; recall \eqref{brackii}.

Our first result describes the behavior of the sequence of gradients of $F$ when fixed stepsizes are employed.  Recall from Assumption~\ref{ass.sg} that the sequence of function values $\{ F(w_k) \}$ is assumed to be bounded below by a scalar $F_{\inf}$.

\btheorem[\textbf{Nonconvex Objective, Fixed Stepsize}]\label{th.sg_nc_fixed}
  Under Assumptions~\ref{ass.Lipschitz} and \ref{ass.sg}, suppose that the SG method (Algorithm~\ref{alg.sg}) is run with a fixed stepsize, $\alpha_k = \bar\alpha$ for all $k \in \N{}$, satisfying
  \bequation\label{eq.alpha_fixed_nc}
    0 < \bar\alpha \leq \frac{\mu}{L M_G}.
  \eequation
  Then, the expected sum-of-squares and average-squared gradients of $F$ corresponding to the SG iterates satisfy the following inequalities for all $K\in\N{}$:
  \bsubequations
    \begin{align}
      \E\left[\sum_{k=1}^{K} \|\nabla F(w_{k})\|_2^2\right]
      \leq&\ \frac{K\bar\alpha LM}{\mu} + \frac{2(F(w_1)-F_{\inf})}{\mu\bar\alpha} \label{eq.sg_nc_fixed_sum} \\
      \text{and therefore}\ \  \E\left[\frac{1}{K} \sum_{k=1}^{K} \|\nabla F(w_{k})\|_2^2\right]
      \leq&\ \frac{\bar\alpha LM}{\mu} + \frac{2(F(w_1)-F_{\inf})}{K\mu\bar\alpha} \label{eq.sg_nc_fixed} \\
      \xrightarrow{K\rightarrow\infty} &\ \frac{\bar\alpha LM}{\mu}. \nonumber
    \end{align}
  \esubequations
\etheorem

\bproof
  Taking the total expectation of \eqref{eq.expected_decrease} and from \eqref{eq.alpha_fixed_nc},
  \bequalin
  \E[F(w_{k+1})] - \E[F(w_k)]
      & \leq - (\mu - \thalf\bar\alpha LM_G)\bar\alpha\E[\|\nabla F(w_k)\|_2^2] + \thalf \bar\alpha^2 L M \\
      & \leq - \thalf\mu\bar\alpha\E[\|\nabla F(w_k)\|_2^2] + \thalf \bar\alpha^2 L M.
  \eequalin
  Summing both sides of this inequality for $k\in\{1,\dots,K\}$ and recalling Assumption~\ref{ass.sg}(a) gives
  \bequationn
    F_{\inf} - F(w_1) \leq \E[F(w_{K+1})]-F(w_1)
      \leq - \thalf \mu \bar\alpha \sum_{k=1}^{K} \E[\|\nabla F(w_k)\|_2^2] + \thalf K \bar\alpha^2 L M.
  \eequationn
  Rearranging yields \eqref{eq.sg_nc_fixed_sum}, and dividing further by $K$ yields \eqref{eq.sg_nc_fixed}.
\eproof

If $M=0$, meaning that there is no noise or that noise reduces proportionally to $\| \nabla F(w_k)\|_2^2$ (see equations~\eqref{eq.variance_bound} and~\eqref{eq.second_moment}), then \eqref{eq.sg_nc_fixed_sum} captures a classical result for the full gradient method applied to nonconvex functions, namely, that  the sum of squared gradients remains finite, implying that $\{\|\nabla F(w_k)\|_2\} \rightarrow 0$.  In the presence of noise (i.e., $M>0$), Theorem~\ref{th.sg_nc_fixed} illustrates the interplay between the stepsize $\bar\alpha$ and the variance of the stochastic directions.  While one cannot bound the expected optimality gap as in the convex case, inequality \eqref{eq.sg_nc_fixed} bounds the average norm of the gradient of the objective function observed on $\{w_k\}$ visited during the first~$K$ iterations.  This quantity gets smaller when $K$ increases, indicating that the SG method spends increasingly more time in regions where the objective function has a (relatively) small gradient.  Moreover, the asymptotic result that one obtains from~\eqref{eq.sg_nc_fixed} illustrates that noise in the gradients inhibits further progress, as in \eqref{uxmal} for the convex case. The average norm of the gradients can be made arbitrarily small by selecting a small stepsize, but doing so reduces the speed at which the norm of the gradient approaches its limiting distribution.

We now turn to the case when the SG method is applied to a nonconvex objective with a decreasing sequence of stepsizes satisfying the classical conditions \eqref{eq.alphas}.  While not the strongest result that one can prove in this context---and, in fact, we prove a stronger result below---the following theorem is perhaps the easiest to interpret and remember.  Hence, we state it first.

\btheorem[\textbf{Nonconvex Objective, Diminishing Stepsizes}]\label{th.sg_liminf}
  Under Assumptions~\ref{ass.Lipschitz} and \ref{ass.sg}, suppose that the SG method (Algorithm~\ref{alg.sg}) is run with a stepsize sequence satisfying \eqref{eq.alphas}. Then
  \begin{equation}  \label{cocoi}
      \liminf_{k\rightarrow\infty}\,\E[ \|\nabla F(w_k)\|^2_2 ] = 0\, .
  \end{equation}
\etheorem

The proof of this theorem follows based on the results given in Theorem~\ref{th.sg_nc} below.  A ``$\liminf$'' result of this type should be familiar  to those knowledgeable of the nonlinear optimization literature.  After all, such a result is all that can be shown for certain important methods, such as the nonlinear conjugate gradient method \cite{NoceWrig06}.  The intuition that one should gain from the statement of Theorem~\ref{th.sg_liminf} is that, for the SG method with diminishing stepsizes, the expected gradient norms cannot stay bounded away from zero.

The following result characterizes more precisely the convergence property of SG. 

\btheorem[\textbf{Nonconvex Objective, Diminishing Stepsizes}]\label{th.sg_nc}
  Under Assumptions~\ref{ass.Lipschitz} and \ref{ass.sg}, suppose that the SG method (Algorithm~\ref{alg.sg}) is run with a stepsize sequence satisfying \eqref{eq.alphas}.  Then, with $A_K := \sum_{k=1}^{K}\alpha_k$,
  \bsubequations
    \begin{align}
    \lim_{K\to\infty} \E\left[\sum_{k=1}^{K} \alpha_k \|\nabla F(w_k)\|^2_2 \right] &< \infty \label{eq.zout} \\
    \text{and therefore}\ \ \E\left[\frac{1}{A_K} \sum_{k=1}^{K} \alpha_k \|\nabla F(w_k)\|^2_2 \right] &\xrightarrow{K\rightarrow\infty} 0. \label{eq.zout2}
    \end{align}
  \esubequations
\etheorem
\bproof
The second condition in \eqref{eq.alphas} ensures that $\{\alpha_k\} \rightarrow 0$, meaning that, without loss of generality, we may assume that $\alpha_k L M_G \leq\mu$ for all $k\in\N{}$.  Then, taking the total expectation of \eqref{eq.expected_decrease},
\bequalin
\E[F(w_{k+1})] - \E[F(w_k)]
      & \leq - (\mu - \thalf\alpha_k LM_G) \alpha_k \,\E[\|\nabla F(w_k)\|_2^2] + \thalf \alpha_k^2 L M \\
      & \leq  - \thalf \mu \alpha_k \E[\|\nabla F(w_k)\|_2^2] + \thalf \alpha_k^2 L M \,.
\eequalin
Summing both sides of this inequality for $k\in\{1,\dots,K\}$ gives
\bequationn
   F_{\inf} - \E[F(w_1)] \leq \E[F(w_{K+1})]-\E[F(w_1)] \leq
     - \thalf \mu \sum_{k=1}^{K} \alpha_k \E[\|\nabla F(w_k)\|_2^2] + \thalf LM \sum_{k=1}^{K}\alpha_k^2\,.
\eequationn
Dividing by $\mu/2$ and rearranging the terms, we obtain
\bequationn
  \sum_{k=1}^{K} \alpha_k \E[\|\nabla F(w_k)\|_2^2] ~\leq~
          \frac{2(\E[F(w_1)]-F_{\inf})}{\mu} + \frac{LM}{\mu} \sum_{k=1}^{K}\alpha_k^2\,.
\eequationn
The second condition in \eqref{eq.alphas} implies that the right-hand side of this inequality converges to a finite limit when $K$ increases, proving \eqref{eq.zout}.  Then, \eqref{eq.zout2} follows since the first condition in \eqref{eq.alphas} ensures that $A_K\rightarrow\infty$ as $K \to \infty$. 
\eproof

Theorem~\ref{th.sg_nc} establishes results about a weighted sum-of-squares and a weighted average of squared gradients of $F$ similar to those in Theorem~\ref{th.sg_nc_fixed}. However, unlike~\eqref{eq.sg_nc_fixed}, the conclusion~\eqref{eq.zout2} states that the weighted average norm of the squared gradients converges to zero even if the gradients are noisy, i.e.,  if $M>0$.  The fact that \eqref{eq.zout2} only specifies a property of a weighted average (with weights dictated by $\{\alpha_k\}$) is only of minor importance since one can still conclude that the expected gradient norms cannot asymptotically stay far from zero.

We can now see that Theorem~\ref{th.sg_liminf} is a direct consequence of   Theorem~\ref{th.sg_nc}, for if \eqref{cocoi} did not hold, it would contradict Theorem~\ref{th.sg_nc}.


The next result gives a stronger conclusion than Theorem~\ref{th.sg_liminf}, at the expense of only showing a property of the gradient of $F$ at a randomly selected iterate.

\bcorollary\label{th.sg_limprob}
  Suppose the conditions of Theorem~\ref{th.sg_nc} hold.  For any $K \in \N{}$, let $k(K)\in\{1,\dots,K\}$ represent a random index chosen with probabilities proportional to $\{\alpha_k\}_{k=1}^K$.  Then, $\|\nabla F(w_{k(K)})\|_2\:\xrightarrow{K\rightarrow\infty}\:0 $ in probability.
\ecorollary
\bproof
Using Markov's inequality and \eqref{eq.zout}, for any $\varepsilon>0$, we can write
\bequationn
\P\{ \|\nabla F(w_k)\|_2 \geq \varepsilon \} = \P\{ \|\nabla F(w_k)\|_2^2 \geq \varepsilon^2 \} 
\leq \varepsilon^{-2}\, \E[\,\E_{k}[ \|\nabla F(w_k)\|_2^2 ]\,] ~ \xrightarrow{K\rightarrow\infty} ~ 0 \,,
\eequationn
which is the definition of convergence in probability.
\eproof

Finally, we present the following result (with proof in Appendix~\ref{sec.proofs}) to illustrate that stronger convergence results also follow under additional regularity conditions.

\bcorollary\label{th.sg_dd}
Under the conditions of Theorem~\ref{th.sg_nc}, if we further assume that the objective function $F$ is twice differentiable, and that the mapping $w \mapsto \|\nabla F(w)\|_2^2$ has Lipschitz-continuous derivatives, then
\[
   \lim_{k\to\infty} \E[\|\nabla F(w_k)\|_2^2] = 0.
\]
\ecorollary

\subsection{Work Complexity for Large-Scale Learning}\label{sec.work_complexity}   \label{sec.work_complex}

Our discussion thus far has focused on the convergence properties of SG when minimizing a given objective function representing either expected or empirical risk.  However, our investigation would be incomplete without considering how these properties impact the computational workload associated with solving an underlying machine learning problem.  As previewed in \S\ref{sec.overview}, there are arguments that a more slowly convergent algorithm such as SG, with its sublinear rate of convergence, is more efficient for large-scale learning than (full, batch) gradient-based methods that have a linear rate of convergence.  The purpose of this section is to present these arguments in further detail.

As a first attempt for setting up a framework in which one may compare optimization algorithms for large-scale learning, one might be tempted to consider the situation of having a given training set size~$n$ and asking what type of algorithm---e.g., a simple SG or batch gradient method---would provide the best guarantees in terms of achieving a low expected risk.  However, such a comparison is difficult to make when one cannot determine the precise trade-off between per-iteration costs and overall progress of the optimization method that one can guarantee.

An easier way to approach the issue is to consider a \emph{big data} scenario with an infinite supply of training examples, but a limited computational time budget.  One can then ask whether running a simple optimization algorithm such as SG works better than running a more sophisticated batch optimization algorithm.  We follow such an approach now, following the work in \cite{Bott10,BottBous08}.

Suppose that both the expected risk $R$ and the empirical risk $R_n$ attain their minima with parameter vectors $w_*\in\argmin R(w)$ and $w_n\in\argmin R_n(w)$, respectively.  In addition, let $\wtilde_n$ be the approximate empirical risk minimizer returned by a given optimization algorithm when the time budget $\Tcal_{\max}$ is exhausted.  The tradeoffs associated with this scenario can be formalized as choosing the family of prediction functions~$\Hcal$, the number of examples~$n$, and the optimization accuracy~$\epsilon := \E[R_n(\wtilde_n)-R_n(w_n)]$ in order to minimize, within time $\Tcal_{max}$, the \emph{total error}
\bequation\label{eq.error_decomposition}
  \E[R(\wtilde_n)] = \underbrace{R(w_*)}_{\displaystyle\Ecal_{app}(\Hcal)} + ~ \underbrace{\E[R(w_n)-R(w_*)]}_{\displaystyle\Ecal_{est}(\Hcal,n)} ~ + ~ \underbrace{\E[R(\wtilde_n)-R(w_n)]}_{\displaystyle\Ecal_{opt}(\Hcal,n,\epsilon)}.
\eequation
To minimize this error, one needs to balance the contributions from each of the three terms on the right-hand side.  For instance, if one decides to make the optimization more accurate---i.e., reducing $\epsilon$ in the hope of also reducing the \emph{optimization error} $\Ecal_{opt}(\Hcal,n,\epsilon)$ (evaluated with respect to $R$ rather than $R_n$)---one might need to make up for the additional computing time by: $(i)$ reducing the sample size $n$, potentially increasing the \emph{estimation error} $\Ecal_{est}(\Hcal,n)$; or $(ii)$ simplifying the function family $\Hcal$, potentially increasing the \emph{approximation error} $\Ecal_{app}(\Hcal)$.

Useful guidelines for achieving an optimal balance between these errors can be obtained by setting aside the choice of $\Hcal$ and carrying out a worst-case analysis on the influence of the sample size $n$ and optimization tolerance $\epsilon$, which together only influence the estimation and optimization errors. This simplified set-up can be formalized in terms of the macroscopic optimization problem
\bequation\label{prob.simplified}
  \min_{n,\epsilon}\ \Ecal(n,\epsilon) = \E[R(\wtilde_n) - R(w_*)]\ \st\ \Tcal(n,\epsilon)\leq\Tcal_{\max}.
\eequation
The computing time $\Tcal(n,\epsilon)$ depends on the details of the optimization algorithm in interesting ways.  For example, the computing time of a batch algorithm increases linearly (at least) with the number of examples $n$, whereas, crucially, the computing time of a stochastic algorithm is independent of~$n$.  With a batch optimization algorithm, one could consider increasing~$\epsilon$ in order to make time to use more training examples. However, with a stochastic algorithm, one should always use as many examples as possible because the per-iteration computing time is independent of $n$.

To be specific, let us compare the solutions of \eqref{prob.simplified} for prototypical stochastic and batch methods---namely, simple SG and a batch gradient method---using simplified forms for the worst-case of the error function $\Ecal$ and the time function $\Tcal$.  For the error function, a direct application of the uniform laws of large numbers~\cite{Vapn83} yields
\bequationn
 \begin{array}{ccccccc}
   \Ecal(n,\epsilon) = \E[ R(\wtilde_n) - R(w_*) ]
      & = &  \underbrace{\E[R(\wtilde_n)-R_n(\wtilde_n)]}
      & + &  \underbrace{\E[R_n(\wtilde_n) - R_n(w_n)]} \\
      &&  =\Ocal\(\sqrt{\log(n)/n}\) &&      =\epsilon \\[1ex]
      & + &  \underbrace{\E[R_n(w_n) - R_n(w_*)]}
      & + &  \underbrace{\E[R_n(w_*) - R(w_*)]}, \\
      &&  \leq0  &&   =\Ocal\(\sqrt{\log(n)/n}\) 
\end{array}
\eequationn
which leads to the upper bound
\bequation\label{eq.initial_rate}
   \Ecal(n,\epsilon) = \mathcal{O}\left( \sqrt{\frac{\log(n)}{n}} + \epsilon \right).
\eequation
The inverse-square-root dependence on the number of examples~$n$ that appears here is typical of statistical problems.  However, even faster convergence rates for reducing these terms with respect to $n$ can be established under specific conditions.  For instance, when the loss function is strongly convex~\cite{LeeBartWill98} or when the data distribution satisfies certain assumptions~\cite{Tsyb04}, it is possible to show that
\bequationn
   \Ecal(n,\epsilon) = \mathcal{O}\left( \frac{\log(n)}{n}+\epsilon \right).
\eequationn
To simplify further, let us work with the asymptotic (i.e., for large $n$) equivalence
\bequation\label{eq.fastest_rate}
   \Ecal(n,\epsilon) \sim \frac{1}{n} + \epsilon,
\eequation
which is the fastest rate that remains compatible with elementary statistical results.\footnote{For example, suppose that one is estimating the mean of a distribution $P$ defined on the real line by minimizing the risk $R(\mu)=\int (x-\mu)^2 dP(x)$.  The convergence rate~\eqref{eq.fastest_rate} amounts to estimating the distribution mean with a variance proportional to $1/n$. A faster convergence rate would violate the Cramer-Rao bound.}  Under this assumption, noting that the time constraint in \eqref{prob.simplified} will always be active (since one can always lower $\epsilon$, and hence $\Ecal(n,\epsilon)$, by giving more time to the optimization algorithm), and recalling the worst-case computing time bounds introduced in~\S\ref{sec.motiv_sg}, one arrives at the following conclusions.
\bitemize
  \item
  A simple SG method can achieve $\epsilon$-optimality with a computing time of $\Tcal_{\rm stoch}(n,\epsilon) \sim 1/\epsilon$.  Hence, within the time budget $\Tcal_{\max}$, the accuracy achieved is proportional to $1/\Tcal_{\max}$, regardless of the $n$.  This means that, to minimize the error~$\Ecal(n,\epsilon)$, one should simply choose~$n$ as large as possible.  Since the maximum number of examples that can be processed by SG during the time budget is proportional to $\Tcal_{\max}$, it follows that the optimal error is proportional to $1/\Tcal_{\max}$.
  \item
  A batch gradient method can achieve $\epsilon$-optimality with a computing time of $\Tcal_{\rm batch}(n,\epsilon) \sim n\log(1/\epsilon)$.  This means that, within the time budget $\Tcal_{\max}$, it can achieve $\epsilon$-optimality by processing $n \sim \Tcal_{\max}/\log(1/\epsilon)$ examples.  One now finds that the optimal error is not necessarily achieved by choosing $n$ as large as possible, but rather by choosing $\epsilon$ (which dictates $n$) to minimize~\eqref{eq.fastest_rate}.  Differentiating $\Ecal(n,\epsilon) \sim \log(1/\epsilon)/\Tcal_{\max} + \epsilon$ with respect to $\epsilon$ and setting the result equal to zero, one finds that optimality is achieved with $\epsilon \sim 1/\Tcal_{\max}$, from which it follows that the optimal error is proportional to $\log(\Tcal_{\max})/\Tcal_{\max} + 1/\Tcal_{\max}$.
\eitemize

These results are summarized in Table~\ref{table.trade_offs}. Even though a batch approach possesses a better dependency on $\epsilon$, this advantage does not make up for its dependence on $n$.  This is true even though we have employed \eqref{eq.fastest_rate}, the most favorable rate that one may reasonably assume.  In conclusion, we have found that a stochastic optimization algorithm performs better in terms of expected error, and, hence, makes a better learning algorithm in the sense considered here.  These observations are supported by practical experience (recall Figure~\ref{wins} in \S\ref{sec.motiv_sg}).


\btable[ht]
\centering\par\medskip
\def\ds{\displaystyle}
\btabular{| l | c| c|}\hline
\rule{0pt}{2.2ex} & Batch & Stochastic \\
\hline
\rule[-3ex]{0pt}{7ex}$\Tcal(n,\epsilon)$&$\sim$
 $\ds n \log\left(\frac{1}{\epsilon}\right)$ & $\ds \frac{1}{\epsilon}$ \\
\rule[-3ex]{0pt}{7ex}$\Ecal^*$&$\sim$
 $\ds \frac{\log(\Tcal_{\max})}{\Tcal_{\max}} + \frac{1}{\Tcal_{\max}}$ & $\ds \frac{1}{\Tcal_{\max}}$ \\
\hline
\etabular
\caption{The first row displays the computing times of idealized batch and stochastic optimization algorithms. The second row gives the corresponding solutions of~\eqref{prob.simplified}, assuming~\eqref{eq.fastest_rate}.}
\label{table.trade_offs}
\etable

\paragraph{\textbf{A Lower Bound}}

The results reported in Table~\ref{table.trade_offs} are also notable because the SG algorithm matches a lower complexity bound that has been established for any optimization method employing a \emph{noisy oracle}.  To be specific, in the widely employed model for studying optimization algorithms  proposed by Nemirovsky and Yudin \cite{NemiYudi78}, one assumes that information regarding the objective function is acquired by querying an \emph{oracle}, ignoring the computational demands of doing so.  Using such a model, it has been established, e.g., that the full gradient method applied to minimize a strongly convex objective function is not optimal in terms of the accuracy that can be achieved within a given number of calls to the oracle, but that one can achieve an optimal method through acceleration techniques; see~\S\ref{sec.accelerated}.

The case when only gradient estimates are available through a noisy oracle has been studied, e.g. 
in~\cite{AgarBartRaviWain12,raginsky2011information}.  Roughly speaking, these investigations show that, again when minimizing a strongly convex function, no algorithm that performs $k$ calls to the oracle can guarantee accuracy better than $\Omega(\smash{1/k})$. As we have seen, SG achieves this lower bound up to constant factors.  This analysis applies both for the optimization of expected risk and empirical risk.

\subsection{Commentary}\label{sec.critique}

Although the analysis presented in~\S\ref{sec.work_complexity} can be quite compelling, it would be premature  to conclude that SG is a perfect solution for large-scale machine learning problems. There is, in fact, a large gap between asymptotical behavior and practical realities.  Next, we discuss issues related to this gap.

\paragraph{\textbf{Fragility of the Asymptotic Performance of SG}}

The convergence speed given, e.g., by Theorem~\ref{th.sg_sc}, holds when the stepsize constant $\beta$ exceeds a quantity inversely proportional to the strong convexity parameter $c$ (see \eqref{eq.alpha_diminishing_specific}).  In some settings, determining such a value is relatively easy, such as when the objective function includes a squared $\ell_2$-norm regularizer (e.g., as in \eqref{eqn.sec2.textsvm}), in which case the regularization parameter provides a lower bound for $c$.  However, despite the fact that this can work well in practice, it is not completely satisfactory because one should reduce the regularization parameter when the number of samples increases.  It is therefore desirable to design algorithms that adapt to local convexity properties of the objective, so as to avoid having to place cumbersome restrictions on the stepsizes.

\paragraph{\textbf{SG and Ill-Conditioning}}

The analysis of \S\ref{sec.work_complexity} is compelling since, as long as the optimization problem is reasonably well-conditioned, the constant factors favor the SG algorithm. In particular, the minimalism of the SG algorithm allows for very efficient implementations that either fully leverage the sparsity of training examples (as in the case study on text classification in~\S\ref{sec.text_classification}) or harness the computational power of GPU processors (as in the case study on deep neural network in~\S\ref{sec.deep_neural_nets}). In contrast, state-of-the-art batch optimization algorithms often carry more overhead.  However, this advantage erodes when the conditioning of the objective function worsens.  Again, consider Theorem~\ref{th.sg_sc}.  This result involves constant factors that grow with both the condition number $L/c$ and the ratio $M/c$. Both of these ratios can be improved greatly by adaptively rescaling the stochastic directions based on matrices that capture local curvature information of the objective function; see~\ref{sec.second_order}.

\paragraph{\textbf{Opportunities for Distributed Computing}}

Distributing the SG step computation can potentially reduce the computing time by a constant factor equal to the number of machines.  However, such an improvement is difficult to realize.  The SG algorithm is notoriously difficult to distribute efficiently because it accesses the shared parameter vector $w$ with relatively high frequency.  Consequently, even though it is very robust to additional noise and can be run with very relaxed synchronization \cite{NiuRechReWrig11,DeanCorrMongChenDeviLeMaoRanzSeniTuckYangNg12}, distributed SG algorithms suffer from large communication overhead.  Since this overhead is potentially much larger than the additional work associated with mini-batch and other methods with higher per-iteration costs, distributed computing offers new opportunities for the success of such methods for machine learning.

\paragraph{\textbf{Alternatives with Faster Convergence}}

As mentioned above, \cite{AgarBartRaviWain12,raginsky2011information} establish lower complexity bounds for optimization algorithms that only access information about the objective function through  noisy estimates of $F(w_k)$ and $\nabla F(w_k)$ acquired in each iteration.  The bounds apply, e.g., when SG is employed to minimize the expected risk $R$ using gradient estimates evaluated on samples drawn from the distribution $P$.  However, an algorithm that optimizes the empirical risk $R_n$ has access to an additional piece of information: it knows when a gradient estimate is evaluated on a training example that has already been visited during previous iterations.  Recent \emph{gradient aggregation} methods (see \S\ref{sec.variancereduction}) make use of this information and improve upon the lower bound in \cite{AgarBartRaviWain12} for the optimization of the empirical risk (though not for the expected risk). These algorithms enjoy linear convergence with low computing times in practice.  Another avenue for improving the convergence rate is to employ a \emph{dynamic sampling} approach (see~\S\ref{sec.dynamic_sample_size}), which, as we shall see, can match the optimal asymptotic efficiency of SG in big data settings.

\renewcommand{\figurename}{Inset}
\setcounter{figtemp}{\value{figure}}
\setcounter{figure}{2}
\begin{figure}[ht]
\begin{adjustbox}{minipage=\textwidth-20pt,margin=8pt 8pt 8pt -4pt,bgcolor=blue!5,frame=2pt}
  \caption{Regret Bounds}
  \label{mini.regret}
  \setlength\parindent{.25in}
  \medskip

  Convergence results for SG and its variants are occasionally established using \emph{regret bounds} as an intermediate step \cite{Zink03,ShalSingSrebCott11,DuchHazaSing11}.  Regret bounds can be traced to Novikoff's analysis of the Perceptron \cite{Novi62} and to Cover's universal portfolios~\cite{Cove91}.  To illustrate, suppose that one aims to minimize a convex expected risk measure $R(w) = \E[f(w;\xi)]$ over $w \in \R{d}$ with minimizer $w_* \in \R{d}$.  At a given iterate $w_k$, one obtains by convexity of $f(w;\xi_k)$ (recall~\eqref{eq.wbound}) that
\bequationn
  \|w_{k+1}-w_*\|^2 - \|w_{k}-w_*\|^2  \leq - 2 \alpha_k (f(w_k;\xi_k)-f(w_*;\xi_k)) + \alpha_k^2 \|\nabla f(w_k;\xi_k)\|_2^2.
\eequationn
Following \cite{Zink03}, assuming that $\|\nabla f(w_k;\xi_k)\|_2^2 \leq M$ and $\|w_k-w_*\|_2^2 < B$ for some constants $M>0$ and $B>0$ for all $k \in \N{}$, one finds that
\bequalin
      &\ \alpha_{k+1}^{-1} \|w_{k+1}\!-\!w_*\|_2^2 - \alpha_{k}^{-1}\|w_{k}\!-\!w_*\|_2^2 \\
  \leq&\ - 2 ( f(w_k;\xi_k) - f(w_*;\xi_k) ) + \alpha_k M + (\alpha_{k+1}^{-1} - \alpha_{k}^{-1}) \|w_k - w_*\|_2^2 \\
  \leq&\ - 2 ( f(w_k;\xi_k) - f(w_*;\xi_k) ) + \alpha_k M + (\alpha_{k+1}^{-1} - \alpha_{k}^{-1}) B.
\eequalin
Summing for $k=\{1\dots K\}$ with stepsizes $\alpha_k=1/\sqrt{k}$ leads to the regret bound
\bequation\label{eq.regretbound}
   \left(\sum_{k=1}^{K} f(w_k;\xi_k)\right) \leq \left( \sum_{k=1}^{K} f(w_*;\xi_k) \right) + M\sqrt{K} + o(\sqrt{K}),
\eequation
which bounds the losses incurred from $\{w_k\}$ compared to those yielded by the fixed vector $w_*$.  Such a bound is remarkable because its derivation holds for any sequence of noise variables~$\{\xi_k\}$. This means that the average loss observed during the execution of the SG algorithm is \emph{never} much worse than the best average loss one would have obtained if the optimal parameter $w_*$ were known in advance. Further assuming that the noise variables are independent and using a martingale argument \cite{CesaConcGent04} leads to more traditional results of the form
\bequationn
  \E\left[ \frac{1}{K}\sum_{k=1}^{K} F(w_k) \right] \leq  F_* + {\cal O}\left(\frac{1}{\sqrt{K}}\right).
\eequationn
As long as one makes the same independent noise assumption, results obtained with this technique cannot be fundamentally different from the results that we have established. However, one should appreciate that the regret bound~\eqref{eq.regretbound} itself remains meaningful when the noise variables are dependent or even adversarial. Such results reveals interesting connections between probability theory, machine learning, and game theory~\cite{ShafVovk05,CesaConcGent04}.
\end{adjustbox}
\end{figure}
\renewcommand{\figurename}{Fig.}
\setcounter{figure}{\value{figtemp}}

\FloatBarrier

\section{Noise Reduction Methods}\label{sec.noise_reduction}
\setcounter{equation}{0}
\setcounter{theorem}{0}
\setcounter{algorithm}{0}
\setcounter{figure}{0}
\setcounter{table}{0}

The theoretical arguments in the previous section, together with extensive computational experience, have led many in the machine learning community to view  SG as the ideal optimization approach for large-scale applications.  We argue, however, that this is far from settled.  SG suffers from, amongst other things, the adverse effect of noisy gradient estimates.  This prevents it from converging to the solution when fixed stepsizes are used and leads to a slow, sublinear rate of convergence when a diminishing stepsize sequence $\{\alpha_k\}$ is employed. 

To address this limitation, methods endowed with \emph{noise reduction} capabilities have been developed. These methods, which reduce the errors in the gradient estimates and/or iterate sequence, have proved to be effective in practice and enjoy attractive theoretical properties.  Recalling the schematic of optimization methods in Figure~\ref{view1}, we depict these methods on the horizontal axis given in Figure~\ref{view2}.

\bfigure[ht]
  \center
   \includegraphics[width=0.6\linewidth,clip=true,trim=15 88 5 55]{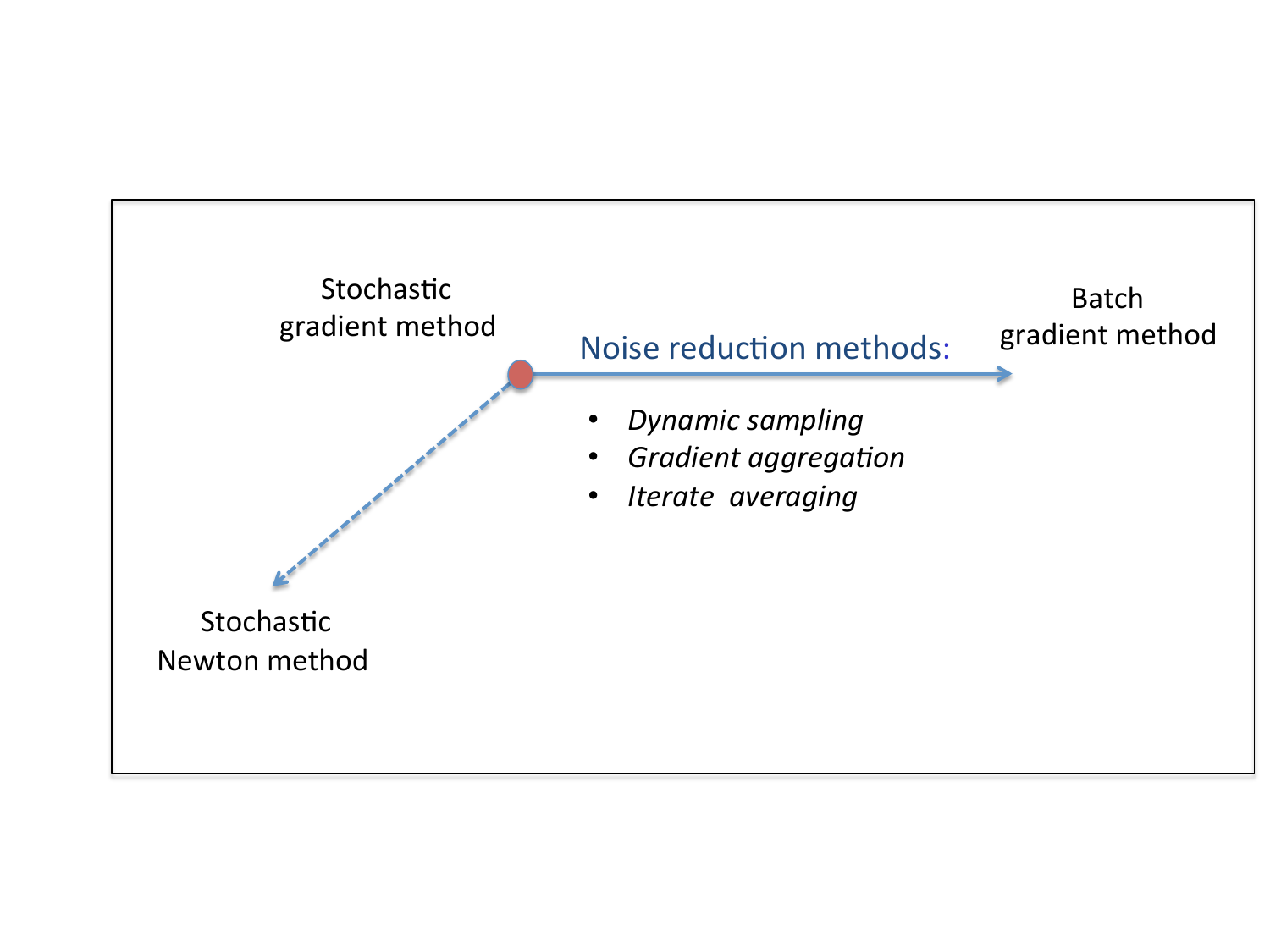}
  \caption{View of the schematic from Figure~\ref{view1} with a focus on noise reduction methods.}
  \label{view2}
\efigure

The first two classes of methods that we consider achieve noise reduction in a manner that allows them to possess a linear rate of convergence to the optimal value using a fixed stepsize.  The first type, \emph{dynamic sampling} methods, achieve noise reduction by gradually increasing the mini-batch size used in the gradient computation, thus employing increasingly more accurate gradient estimates as the optimization process proceeds. \emph{Gradient aggregation} methods, on the other hand, improve the quality of the search directions by storing gradient estimates corresponding to samples employed in previous iterations, updating one (or some) of these estimates in each iteration, and defining the search direction as a weighted average of these estimates.

The third class of methods that we consider, \emph{iterate averaging} methods, accomplish noise reduction not by averaging gradient estimates, but by maintaining an average of iterates computed during the optimization process.  Employing a more aggressive stepsize sequence---of order $\Ocal(1/\sqrt{k})$ rather than $\Ocal(1/k)$, which is appealing in itself---it is this sequence of averaged iterates that converges to the solution. These methods are closer in spirit to SG and their rate of convergence remains sublinear,  but it can be shown that the variance of the sequence of average iterates is smaller than the variance of the SG iterates.

To formally motivate a concept of noise reduction, we begin this section by discussing a fundamental result that stipulates a rate of decrease in noise that allows a stochastic-gradient-type method to converge at a linear rate.   We then show that a dynamic sampling method that enforces such noise reduction enjoys optimal complexity bounds, as defined in \S\ref{sec.work_complexity}.  Next, we discuss three gradient aggregation methods---SVRG, SAGA, and SAG---the first of which can be viewed as a bridge between methods that control errors in the gradient with methods like SAGA and SAG in which noise reduction is accomplished in a more subtle manner.  We conclude with a discussion of the practical and theoretical properties of iterate averaging methods.

\subsection{Reducing Noise at a Geometric Rate}\label{sec.dss}

Let us recall the fundamental inequality \eqref{eq.Lipschitz}, which we restate here for convenience:
\bequationn
  \E_{\xi_k}[F(w_{k+1})] - F(w_k) \leq -\alpha_k\nabla F(w_k)^T\E_{\xi_k}[g(w_k,\xi_k)] + \thalf \alpha_k^2 L \E_{\xi_k}[\|g(w_k,\xi_k)\|_2^2].
\eequationn
(Recall that, as stated in~\eqref{def.F_general}, the objective $F$ could stand for the expected risk $R$ or empirical risk $R_n$.)  It is intuitively clear that if $-g(w_k,\xi_k)$ is a descent direction in expectation (making the first term on the right hand side negative) and if we are able to decrease $\E_{\xi_k}[\|g(w_k,\xi_k)\|_2^2$ fast enough (along with $\|\nabla F(w_k)\|_2^2$), then the effect of having noisy directions will not impede a fast rate of convergence.  From another point of view, we can expect such behavior if, in Assumption~\ref{ass.sg}, we suppose instead that the variance of $g(w_k,\xi_k)$ vanishes sufficiently quickly.

We formalize this intuitive argument by considering the SG method with a fixed stepsize and showing that the sequence of expected optimality gaps vanishes at a linear rate as long as the variance of the stochastic vectors, denoted by $\var_{\xi_k}[g(w_k,\xi_k)]$ (recall \eqref{eq.variance}), decreases \emph{geometrically}.

\btheorem[\textbf{Strongly Convex Objective, Noise Reduction}]\label{th.geometric_decrease}
  Suppose that Assumptions~\ref{ass.Lipschitz}, \ref{ass.sg}, and \ref{ass.sc} (with $F_{\inf}=F_*$) hold, but with \eqref{eq.variance_bound} refined to the existence of constants $M \geq 0$ and $\zeta \in (0,1)$ such that, for all $k \in \N{}$,
  \bequation\label{eq.geometric_decrease}
    \var_{\xi_k}[g(w_k,\xi_k)] \leq M\zeta^{k-1}.
  \eequation
  In addition, suppose that the SG method (Algorithm~\ref{alg.sg}) is run with a fixed stepsize, $\alpha_k = \bar\alpha$ for all $k \in \N{}$, satisfying
  \bequation\label{eq.beta_geometric}
    0 < \bar\alpha \leq \min\left\{\frac{\mu}{L\mu_G^2},\frac{1}{c\mu}\right\}.
  \eequation
  Then, for all $k \in \N{}$, the expected optimality gap satisfies
  \bequation\label{eq.geometric_linear}
    \E[F(w_k) - F_*] \leq \omega \rho^{k-1},
  \eequation
  where
  \bsubequations
    \begin{align}
      \omega &:= \max\{\tfrac{\bar\alpha LM}{c\mu},F(w_1)-F_*\} \label{eq.nu_new} \\
      \text{and}\ \ \rho &:= \max\{1 - \tfrac{\bar\alpha c\mu}{2},\zeta\} < 1. \label{eq.rho}
    \end{align}
  \esubequations
\etheorem
\bproof
  By Lemma~\ref{lem.expected_decrease} (specifically, \eqref{eq.expected_decrease_pre}), we have
  \bequationn
    \E_{\xi_k}[F(w_{k+1})] - F(w_k) \leq -\mu\bar\alpha\|\nabla F(w_k)\|_2^2 + \thalf\bar\alpha^2L\E_{\xi_k}[\|g(w_k,\xi_k)\|_2^2].
  \eequationn
  Hence, from \eqref{eq.variance}, \eqref{eq.mean_condition}, \eqref{eq.geometric_decrease}, \eqref{eq.beta_geometric}, and \eqref{eq.sc2}, we have
  \bequalin
    \E_{\xi_k}[F(w_{k+1})] - F(w_k)
      &\leq -\mu\bar\alpha\|\nabla F(w_k)\|_2^2 + \thalf\bar\alpha^2L\(\mu_G^2\|\nabla F(w_k)\|_2^2 + M\zeta^{k-1}\) \\
      &\leq -(\mu - \thalf\bar\alpha L \mu_G^2)\bar\alpha\|\nabla F(w_k)\|_2^2 + \thalf\bar\alpha^2LM\zeta^{k-1} \\
      &\leq -\thalf\mu\bar\alpha\|\nabla F(w_k)\|_2^2 + \thalf\bar\alpha^2LM\zeta^{k-1} \\
      &\leq -\bar\alpha c\mu(F(w_k) - F_*) + \thalf\bar\alpha^2LM\zeta^{k-1}.
  \eequalin
  Adding and subtracting $F_*$ and taking total expectations, this yields
  \bequation\label{eq.fuji_apple}
    \E[F(w_{k+1}) - F_*] \leq (1 - \bar\alpha c\mu)\E[F(w_k) - F_*] + \thalf\bar\alpha^2LM\zeta^{k-1}.
  \eequation
  
  We now use induction to prove \eqref{eq.geometric_linear}.  By the definition of $\omega$, it holds for $k=1$.  Then, assuming it holds for $k \geq 1$, we have from \eqref{eq.fuji_apple}, \eqref{eq.nu_new}, and \eqref{eq.rho} that
  \bequalin
    \E[F(w_{k+1}) - F_*]
      &\leq \(1 - \bar\alpha c\mu\)\omega\rho^{k-1} + \thalf\bar\alpha^2LM\zeta^{k-1} \\
      &=    \omega\rho^{k-1}\(1 - \bar\alpha c\mu + \frac{\bar\alpha^2LM}{2\omega}\(\frac{\zeta}{\rho}\)^{k-1}\) \\
      &\leq \omega\rho^{k-1}\(1 - \bar\alpha c\mu + \frac{\bar\alpha^2LM}{2\omega}\) \\
      &\leq \omega\rho^{k-1}\(1 - \bar\alpha c\mu + \frac{\bar\alpha c\mu}{2}\) \\
      &=    \omega\rho^{k-1}\(1 - \frac{\bar\alpha c\mu}{2}\) \\
      &\leq \omega\rho^k,
  \eequalin
  as desired.
\eproof

Consideration of the typical magnitudes of the constants $\mu$, $L$, $\mu_G$, and $c$ in \eqref{eq.beta_geometric} reveals that the admissible range of values of $\bar\alpha$ is large, i.e., the restriction on the stepsize $\bar\alpha$ is not unrealistic in practical situations.

Now, a natural question is whether one can design efficient optimization methods for attaining the critical bound \eqref{eq.geometric_decrease} on the variance of the stochastic directions.  We show next that a dynamic sampling method is one such approach.

\subsection{Dynamic Sample Size Methods}\label{sec.dynamic_sample_size}

Consider the iteration
\bequation\label{gets}
  w_{k+1} \gets w_k - \bar\alpha g(w_k,\xi_k),
\eequation
where the stochastic directions are computed for some $\tau > 1$ as
\bequation\label{eq.geometric_growth}
  g(w_k,\xi_k) := \frac{1}{n_k} \sum_{i\in\Scal_k} \nabla f(w_k;\xi_{k,i})\ \ \text{with}\ \ n_k := |\Scal_k| = \lceil\tau^{k-1}\rceil.
\eequation
That is, consider a mini-batch 
SG iteration with a fixed stepsize in which the mini-batch size used to compute unbiased stochastic gradient estimates increases geometrically as a function of the iteration counter~$k$.  To show that such an approach can fall under the class of methods considered in Theorem~\ref{th.geometric_decrease}, suppose that the samples represented by the random variables $\{\xi_{k,i}\}_{i\in\Scal_k}$ are drawn independently according to $P$ for all $k \in \N{}$.  If we assume that each stochastic gradient $\nabla f(w_k;\xi_{k,i})$ has an expectation equal to the true gradient $\nabla F(w_k)$, then \eqref{eq.descent_condition} holds with $\mu_G = \mu = 1$.  If, in addition, the variance of each such stochastic gradient is equal and is bounded by $M \geq 0$, then for arbitrary $i \in \Scal_k$ we have (see \cite[p.183]{Freu62})
\bequation\label{eq.variance_red}
  \var_{\xi_k}[g(w_k,\xi_k)] \leq \frac{\var_{\xi_k}[\nabla f(w_k;\xi_{k,i})]}{n_k} \leq \frac{M}{n_k}. 
\eequation
This bound, when combined with the rate of increase in $n_k$ given in \eqref{eq.geometric_growth}, yields \eqref{eq.geometric_decrease}. We have thus shown that if one employs a mini-batch SG method with (unbiased) gradient estimates computed as in \eqref{eq.geometric_growth}, then, by Theorem~\ref{th.geometric_decrease}, one obtains linear convergence to the optimal value of a strongly convex function.  We state this formally as the following corollary to~Theorem~\ref{th.geometric_decrease}.

\bcorollary
 Let $\{w_k\}$ be the iterates generated by \eqref{gets}--\eqref{eq.geometric_growth} with unbiased gradient estimates, i.e., $\E_{\xi_{k,i}}[\nabla f(w_k;\xi_{k,i})] = \nabla F(w_k)$ for all $k \in \N{}$ and $i \in \Scal_k$.  Then, the variance condition \eqref{eq.geometric_decrease} is satisfied, and if all other assumptions of Theorem~\ref{th.geometric_decrease} hold, then the expected optimality gap vanishes linearly in the sense of \eqref{eq.geometric_linear}.
\ecorollary

The reader may question whether it is meaningful to describe a method as linearly convergent if the per-iteration cost increases without bound. In other words, it is not immediately apparent that such an algorithm is competitive with a classical SG approach even though the desired reduction in the gradient variance is achieved.  To address this question, let us estimate the total work complexity of the dynamic sampling algorithm, defined as the number of evaluations of the individual gradients $\nabla f(w_k;\xi_{k,i})$ required to compute an $\epsilon$-optimal solution, i.e., to achieve
\bequation\label{wiles}
  \E[F(w_k) - F_*] \leq \epsilon .
\eequation
We have seen that the simple SG method \eqref{eq.sg_init} requires one such evaluation per iteration, and that its rate of convergence for diminishing stepsizes (i.e., the only set-up in which convergence to the solution can be guaranteed) is given by \eqref{sublin}.  Therefore, as previously mentioned, the number of stochastic gradient evaluations required by the SG method to guarantee \eqref{wiles} is $\Ocal(\epsilon^{-1})$.  We now show that the method \eqref{gets}--\eqref{eq.geometric_growth} can attain the same complexity.

\btheorem
  Suppose that the dynamic sampling SG method \eqref{gets}--\eqref{eq.geometric_growth} is run with a stepsize~$\bar\alpha$ satisfying \eqref{eq.beta_geometric} and some $\tau \in (1,(1-\frac{\bar\alpha c\mu}{2})^{-1}]$.  In addition, suppose that Assumptions~\ref{ass.Lipschitz}, \ref{ass.sg}, and Assumption~\ref{ass.sc} (with $F_{\inf} = F_*$) hold.  Then, the total number of evaluations of a stochastic gradient of the form $\nabla f(w_k;\xi_{k,i})$  required to obtain \eqref{wiles} is $\Ocal(\epsilon^{-1})$.
\etheorem
\bproof
  We have that the conditions of Theorem~\ref{th.geometric_decrease} hold with $\zeta = 1/\tau$.  Hence, we have from \eqref{eq.geometric_linear} that there exists $\kbar \in \N{}$ such that \eqref{wiles} holds for all $k \geq \kbar$.   We can then use $\omega\rho^{\kbar-1} \leq \epsilon$ to write $(\kbar-1)\log\rho \leq \log(\epsilon/\omega)$, which along with $\rho \in (0,1)$ (recall \eqref{eq.rho}) implies that
  \bequation\label{sbarro}
    \kbar-1 \geq \left\lceil\frac{\log(\epsilon/\omega)}{\log\rho}\right\rceil = \left\lceil\frac{\log(\omega/\epsilon)}{-\log\rho}\right\rceil.
  \eequation
  
  Let us now estimate the total number of sample gradient evaluations required up through iteration $\kbar$.  We claim that, without loss of generality, we may assume that $\log(\omega/\epsilon)/(-\log\rho)$ is integer-valued and that \eqref{sbarro} holds at equality.  Then, by \eqref{eq.geometric_growth}, the number of sample gradients required in iteration $\kbar$ is $\lceil\tau^{\kbar-1}\rceil$ where
  \bequalin
    \tau^{\kbar-1}
     &= \tau^{\frac{\log(\omega/\epsilon)}{-\log\rho}} \\
      &= \exp\(\log\(\tau^{\frac{\log(\omega/\epsilon)}{-\log\rho}}\)\) \\
      &= \exp\(\(\frac{\log(\omega/\epsilon)}{-\log\rho}\)\log\tau\) \\
      &= \(\exp(\log(\omega/\epsilon))\)^{\frac{\log\tau}{-\log\rho}} \\
      &= \(\frac{\omega}{\epsilon}\)^\theta\ \ \text{with}\ \ \theta := \frac{\log\tau}{-\log\rho}.
  \eequalin
  Therefore, the total number of sample gradient evaluations for the first $\kbar$ iterations is
  \bequationn
    \sum_{j=1}^{\kbar} \lceil \tau^{j-1} \rceil \leq 2 \sum_{j=1}^{\kbar} \tau^{j-1} = 2\(\frac{\tau^{\kbar} - \tau}{\tau - 1}\) = 2\(\frac{\tau(\omega/\epsilon)^\theta - \tau}{\tau - 1}\) \leq 2\(\frac{\omega}{\epsilon}\)^\theta\(\frac{1}{1-1/\tau}\).
  \eequationn
  In fact, since $\tau \leq (1-\frac{\bar\alpha c\mu}{2})^{-1}$, it follows from \eqref{eq.rho} that $\rho = \zeta = \tau^{-1}$, which implies that $\theta = 1$.  Specifically, with $\tau = (1-\sigma(\frac{\bar\alpha c\mu}{2}))^{-1}$ for some $\sigma \in [0,1]$, then $\theta = 1$ and
  \bequationn
    \sum_{j=1}^{\kbar} \lceil \tau^{j-1} \rceil \leq \frac{4\omega}{\sigma\epsilon\bar\alpha c\mu},
  \eequationn
  as desired.
\eproof

The discussion so far has focused on dynamic sampling strategies for a gradient method, but these ideas also apply for second-order methods that incorporate the matrix $H$ as in \eqref{brackii}.

This leads to the following question: given the rate of convergence of a batch optimization algorithm on strongly convex functions (i.e., linear, superlinear, etc.),~what should be the sampling rate so that the overall algorithm is \emph{efficient} in the sense that it results in the lowest computational complexity?   To answer this question, certain results have been established \cite{PasuGlynGhosHash15}: $(i)$~if the optimization method has a sublinear rate of convergence, then there is no sampling rate that makes the algorithm ``efficient''; $(ii)$ if the optimization algorithm is linearly convergent, then the sampling rate must be geometric (with restrictions on the constant in the rate) for the algorithm to be ``efficient''; $(iii)$  for superlinearly convergent methods, increasing the sample size at a rate that is slightly faster than geometric will yield an ``efficient'' method.

\subsubsection{Practical Implementation}

We now address the question of how to design practical algorithms that enjoy the theoretical benefits of noise reduction through the use of increasing sample sizes.  

One approach is to follow the theoretical results described above and tie the rate of growth in the sample size $n_k = |{\cal S}_k |$ to the rate of convergence of the  optimization iteration \cite{FrieSchm12,HomeShap98}.  Such an approach, which \emph{presets} the sampling rate before running the optimization algorithm, requires some experimentation. For example, for the iteration \eqref{gets}, one needs to find a value of the parameter $\tau >1$ in \eqref{eq.geometric_growth} that yields good performance for the application at hand.  In addition, one may want to delay the application of dynamic sampling to prevent the full sample set from being employed too soon (or at all). Such heuristic adaptations could be difficult in practice.

An alternative is to consider mechanisms for choosing the sample sizes not according to a prescribed sequence, but \emph{adaptively} according to information gathered during the optimization process. One avenue that has been explored along these lines has been to design techniques that produce descent directions sufficiently often \cite{ByrdChinNoceWu12,HashGhosPasu14}.  Such an idea is based on two observations.  First, one can show that \emph{any} direction $g(w_k,\xi_k)$ is a descent direction for $F$ at $w_k$ if, for some $\chi \in [0,1)$, one has
\bequation\label{honest}
  \delta(w_k,\xi_k) := \|g(w_k,\xi_k) - \nabla F(w_k)\|_2 \leq \chi \|g(w_k,\xi_k)\|_2.
\eequation
To see this, note that  $\|\nabla F(w_k)\|_2 \geq (1 - \chi)\|g(w_k,\xi_k)\|_2$, which after squaring \eqref{honest} implies
\bequalin
  \nabla F(w_k)^Tg(w_k,\xi_k)
    &\geq \thalf(1 - \chi^2)\|g(w_k,\xi_k)\|_2^2 + \|\nabla F(w_k)\|_2^2 \\
    &\geq \thalf(1 - \chi^2 +(1-\chi)^2) \|g(w_k,\xi_k)\|_2^2 \\
    &\geq (1 - \chi) \|g(w_k,\xi_k)\|_2^2.
\eequalin
The second observation is that while one cannot cheaply verify the inequality in \eqref{honest} because it involves the evaluation of $\nabla F(w_k)$, one can estimate the left-hand side $ \delta(w_k,\xi_k)$, and then choose $n_k$ so that \eqref{honest} holds sufficiently often. 

Specifically, if we assume that $g(w_k,\xi_k)$ is an unbiased gradient estimate, then 
\bequationn
  \E[\delta(w_k,\xi_k)^2] = \E[\|g(w_k,\xi_k) - \nabla F(w_k)\|_2^2] = \var_{\xi_k}[g(w_k,\xi_k)].
\eequationn
This variance is expensive to compute, but one can approximate it with a sample variance.  For example, if the samples are drawn without replacement from a set of (very large) size $n_k$, then one has the approximation
\bequationn
  \var_{\xi_k}[g(w_k,\xi_k)] \approx \frac{\trace({\rm Cov}(\{\nabla f(w_k;\xi_{k,i})\}_{i\in\Scal_k}))}{n_k} =: \varphi_k.
\eequationn
An \emph{adaptive sampling} algorithm thus tests the following condition in place of \eqref{honest}:
\begin{equation}\label{darte}
  \varphi_k \leq \chi^2 \|g(w_k,\xi_k)\|_2^2.
\end{equation}
If this condition is not satisfied, then one may consider increasing the sample size---either immediately in iteration $k$ or in a subsequent iteration---to a size that one might predict would satisfy such a condition.  This technique is algorithmically attractive, but does not guarantee that the sample size $n_k$ increases at a geometric rate. One can, however, employ a backup \cite{ByrdChinNoceWu12,HashGhosPasu14}: if \eqref{darte} increases the sampling rate more slowly than a preset geometric sequence, then a growth in the sample size is imposed. 
 
Dynamic sampling techniques are not yet widely used in machine learning, and we suspect that the practical technique presented here might merely serve as a starting point for further investigations.  Ultimately, an algorithm that performs like SG at the start and transitions to a regime of reduced variance in an efficient manner could prove to be a very powerful method for large-scale learning.
 
\subsection{Gradient Aggregation}
\label{sec.variancereduction}

The dynamic sample size methods described in the previous subsection require a careful balance in order to achieve the desired linear rate of convergence without jeopardizing per-iteration costs.  Alternatively, one can attempt to achieve an improved rate by asking a different question: rather than compute increasingly more \emph{new} stochastic gradient information in each iteration, is it possible to achieve a lower variance by \emph{reusing} and/or \emph{revising} previously computed information?  After all, if the current iterate has not been displaced too far from previous iterates, then stochastic gradient information from previous iterates may still be useful.  In addition, if one maintains indexed gradient estimates in storage, then one can revise specific estimates as new information is collected.  Ideas such as these lead to concepts of \emph{gradient aggregation}.  In this subsection, we present ideas for gradient aggregation in the context of minimizing a finite sum such as an empirical risk measure~$R_n$, for which they were originally designed.


Gradient aggregation algorithms for minimizing finite sums that possess cheap per-iteration costs have a long history.  For example, Bertsekas and co-authors \cite{Bert95} have proposed \emph{incremental gradient} methods, the randomized versions of which can be viewed as instances of a basic SG method for minimizing a finite sum.  However, the basic variants of these methods only achieve a sublinear rate of convergence.  By contrast, the methods on which we focus in this section are able to achieve a linear rate of convergence on strongly convex problems.  This improved rate is achieved primarily by either an increase in computation or an increase in storage.

\subsubsection{SVRG}

The first method we consider operates in cycles.  At the beginning of each cycle, an iterate $w_k$ is available at which the algorithm computes a batch gradient $\nabla R_n(w_k) = \frac{1}{n} \sum_{i=1}^n \nabla f_i(w_k)$.  Then, after initializing $\wtilde_1 \gets w_k$, a set of $m$ inner iterations indexed by $j$ with an update  $\wtilde_{j+1} \gets \wtilde_j - \alpha \gtilde_j$ are performed, where
\bequation\label{gjz}
  \gtilde_j \gets \nabla f_{i_j}(\wtilde_j) - (\nabla f_{i_j}(w_k) - \nabla R_n(w_k))
\eequation
and $i_j \in \{1,\dots,n\}$ is chosen at random.  This formula has a simple interpretation.  Since the expected value of $\nabla f_{i_j}(w_k)$ over all possible $i_j \in \{1,\dots,n\}$ is equal to $\nabla R_n(w_k)$, one can view $\nabla f_{i_j}(w_k) - \nabla R_n(w_k)$ as the bias in the gradient estimate $\nabla f_{i_j}(w_k)$.  Thus, in every iteration, the algorithm randomly draws a stochastic gradient $\nabla f_{i_j}(\wtilde_j)$ evaluated at the current inner iterate~$\wtilde_j$ and corrects it based on a perceived bias.  Overall, $\gtilde_j$ represents an unbiased estimator of $\nabla R_n(\wtilde_j)$, but with a variance that one can expect to be smaller than if one were simply to chose $\gtilde_j = \nabla f_{i_j}(\wtilde_j)$ (as in simple SG).  This is the reason that the method is referred to as the \emph{stochastic variance reduced gradient} (SVRG) method.

A formal description of a few variants of SVRG is presented as Algorithm~\ref{alg.svrg}. 
 For both options $(b)$ and $(c)$, it has been shown that when applied to minimize a strongly convex $R_n$, 
Algorithm~\ref{alg.svrg} can achieve a linear rate of convergence \cite{johnson2013accelerating}.  More precisely, if the stepsize $\alpha$ and the length of the inner cycle $m$ are chosen so that
\bequationn
  \rho := \frac{1}{1-2 \alpha L} \left(\frac{1}{mc \alpha} + 2L \alpha \right) < 1,
\eequationn
then, given that the algorithm has reached $w_k$, one obtains
\bequationn
  \E[R_n(w_{k+1}) - R_n (w_*)] \leq \rho \E[R_n(w_k) - R_n (w_*)]
\eequationn
(where expectation is taken with respect to the random variables in the inner cycle).  It should be emphasized that resulting linear convergence rate applies to the outer iterates $\{w_k\}$, where each step from $w_k$ to $w_{k+1}$ requires $2m + n$ evaluations of component gradients: Step~\ref{svrg.sg} requires two stochastic gradient evaluations while Step~\ref{svrg.full} requires $n$ (a full gradient).  Therefore, one iteration of SVRG is much more expensive than one in SG, and in fact is comparable to a full gradient iteration.

\balgorithm[ht]
  \caption{SVRG Methods for Minimizing an Empirical Risk $R_n$}
  \label{alg.svrg}
  \balgorithmic[1]
    \State Choose an initial iterate $w_1 \in \R{d}$, stepsize $\alpha > 0$, and positive integer $m$.
    \For{$k = 1,2,\dots$}
      \State Compute the batch gradient $\nabla R_n(w_k)$. \label{svrg.full}
      \State Initialize $\wtilde_1 \gets w_k$.
      \For{$j = 1,\dots,m$}
      \State Chose $i_j$ uniformly from $\{1,\dots,n\}$.
      \State Set $\gtilde_j \gets \nabla f_{i_j}(\wtilde_j) - (\nabla f_{i_j}(w_k) - \nabla R_n(w_k))$. \label{svrg.sg}
      \State Set $\wtilde_{j+1} \gets \wtilde_j - \alpha \gtilde_j$.
      \EndFor
      \State {Option $(a)$:} Set $w_{k+1} = \wtilde_{m+1}$
      \State {Option $(b)$:} Set $w_{k+1} = \frac{1}{m} \sum_{j=1}^m \wtilde_{j+1}$
      \State {Option $(c)$:} Choose $j$ uniformly from $\{1,\dots,m\}$ and set $w_{k+1} = \wtilde_{j+1}$.
    \EndFor
  \ealgorithmic
\ealgorithm

In practice, SVRG appears to be quite effective in certain applications compared with SG if one requires high training accuracy.  For the first epochs, SG is more efficient, but once the iterates approach the solution the benefits of the fast convergence rate of SVRG can be observed.
Without explicit knowledge of $L$ and $c$, the length of the inner cycle~$m$ and the stepsize $\alpha$ are typically both chosen by experimentation.

\subsubsection{SAGA}

The second method we consider employs an iteration that is closer in form to SG in that it does not operate in cycles, nor does it compute batch gradients (except possibly at the initial point).  Instead, in each iteration, it computes a stochastic vector $g_k$ as the average of stochastic gradients evaluated at previous iterates.  Specifically, in iteration $k$, the method will have stored $\nabla f_i(w_{[i]})$ for all $i \in \{1,\dots,n\}$ where $w_{[i]}$ represents the latest iterate at which $\nabla f_i$ was evaluated.  An integer $j \in \{1,\dots,n\}$ is then chosen at random and the stochastic vector is set by
\bequation\label{up.saga}
  g_k \gets \nabla f_j(w_k) - \nabla f_j(w_{[j]}) + \frac{1}{n} \sum_{i=1}^n \nabla f_i(w_{[i]}).
\eequation
Taking the expectation of $g_k$ with respect to all choices of $j \in \{1,\dots,n\}$, one again has that $\E[g_k] = \nabla R_n(w_k)$.  Thus, the method employs unbiased gradient estimates, but with variances that are expected to be less than the stochastic gradients that would be employed in a basic SG routine. A precise algorithm employing \eqref{up.saga}, referred to as SAGA  \cite{DefaBachLaco14}, is given in Algorithm~\ref{alg.saga}.

\balgorithm[ht]
  \caption{SAGA Method for Minimizing an Empirical Risk $R_n$}
  \label{alg.saga}
  \balgorithmic[1]
    \State Choose an initial iterate $w_1 \in \R{d}$ and stepsize $\alpha > 0$.
    \For{$i=1,\dots,n$}
      \State Compute $\nabla f_i(w_1)$.
      \State Store $\nabla f_i(w_{[i]}) \gets \nabla f_i(w_1)$.
    \EndFor
    \For{$k = 1,2,\dots$}
      \State Choose $j$ uniformly in $\{1,\dots,n\}$.
      \State Compute $\nabla f_j(w_k)$.
      \State Set $g_k \gets \nabla f_j(w_k) - \nabla f_j(w_{[j]}) + \frac{1}{n} \sum_{i=1}^n \nabla f_i(w_{[i]})$.
      \State Store $\nabla f_j(w_{[j]}) \gets \nabla f_j(w_k)$.
      \State Set $w_{k+1} \gets w_k - \alpha g_k$.
    \EndFor
  \ealgorithmic
\ealgorithm

Beyond its initialization phase, the per-iteration cost of Algorithm~\ref{alg.saga} is the same as in a basic SG method.  However, it has been shown that the method can achieve a linear rate of convergence when minimizing a strongly convex $R_n$.  Specifically, with $\alpha = 1/(2(cn+L))$, one can show that
\bequalin
  \E[\|w_k - w_*\|_2^2] \leq&\ \left(1- \frac{c}{2(cn + L)}\right)^k \(\|w_1 - w_*\|_2^2 + \frac{nD}{cn + L} \),\\\text{where}\ \ 
  D :=&\ R_n(w_1) - \nabla R_n(w_*)^T(w_1 - w_*) - R_n(w_*).
\eequalin
Of course, attaining such a result requires knowledge of the strong convexity constant~$c$ and Lipschitz constant~$L$.  If $c$ is not known, then the stepsize can instead be chosen to be $\alpha = 1/(3L)$ and a similar convergence result can be established; see \cite{DefaBachLaco14}.

Alternative initialization techniques could be used in practice, which may be more effective than evaluating all the gradients $\{\nabla f_i\}_{i=1}^n$ at the initial point. For example, one could perform one epoch of simple SG, or one can assimilate iterates one-by-one and compute $g_k$ only using the gradients available up to that point.

One important drawback of Algorithm~\ref{alg.saga} is the need to store $n$ stochastic gradient vectors, which would be prohibitive in many large-scale applications.  
Note, however, that  if the component functions are of the form $f_i(w_k) = \fhat(x_i^Tw_k)$, then
\bequationn
  \nabla f_i(w_k) = \fhat'(x_i^Tw_k) x_i.
\eequationn
That is, when the feature vectors $\{x_i\}$ are already available in storage, one need only store the scalar $\fhat'(x_i^Tw_k)$ in order to construct $\nabla f_i(w_k)$ at a later iteration.  Such a functional form of $f_i$ occurs in logistic and least squares regression.

Algorithm~\ref{alg.saga} has its origins in the \emph{stochastic average gradient} (SAG)  algorithm \cite{SchmLeRoBach13,LeRoSchmBach12}, where the stochastic direction is defined as
\bequation\label{up.sag}
  g_k \gets \frac{1}{n} \( \nabla f_j(w_k) - \nabla f_j(w_{[j]}) +  \sum_{i=1}^n \nabla f_i(w_{[i]}) \).
\eequation
Although this $g_k$ is not an unbiased estimator of $\nabla R_n(w_k)$, the method enjoys a linear rate of convergence.  One finds that the SAG algorithm is a randomized version of the Incremental Aggregated Gradient (IAG) method proposed in \cite{BlatHeroGauc07} and analyzed in~\cite{gurbuzbalaban2015convergence}, where the index $j$ of the component function updated at every iteration is chosen cyclically.  Interestingly, randomizing this choice yields good practical benefits.

\subsubsection{Commentary}

Although the gradient aggregation methods discussed in this section enjoy a faster rate of convergence than SG (i.e., linear vs.~sublinear), they should not be regarded as clearly superior to SG.  After all, following similar analysis as in \S\ref{sec.work_complex}, the computing time for SG can be shown to be $T(n,\epsilon) \sim \kappa^2/\epsilon$ with $\kappa := L/c$.  (In fact, a computing time of $\kappa/\epsilon$ is often observed in practice.)  On the other hand, the computing times for SVRG, SAGA, and SAG are
\bequationn
  \Tcal(n,\epsilon) \sim (n+\kappa)\log(1/\epsilon),
\eequationn
which grows with the number of examples $n$.  
Thus, following similar analysis as in \S\ref{sec.work_complex}, one finds that, for very large $n$, gradient aggregation methods are comparable to batch algorithms and therefore cannot beat SG in this regime.  For example, 
if $\kappa$ is close to 1, then SG is clearly more efficient since within a single epoch it reaches the optimal testing error \cite{BottLeCu04}.  On the other hand, there exists a regime with $\kappa \gg n$ in which gradient aggregation methods may be superior, and perhaps even easier to tune.  At present, it is not known how useful gradient aggregation methods will prove to be in the future of large-scale machine learning.  That being said, they certainly represent a class of optimization methods of interest due to their clever use of past information.
  
\subsection{Iterate Averaging Methods}

Since its inception, it has been observed that SG generates noisy iterate sequences that tend to oscillate around minimizers during the optimization process.  Hence, a natural idea is to compute a corresponding sequence of \emph{iterate averages} that would automatically possess less noisy behavior.  Specifically, for minimizing a continuously differentiable $F$ with unbiased gradient estimates, the idea is to employ the iteration
\bequation\label{polyak-ruppert}
  \begin{aligned}
    w_{k+1} &\gets w_k - \alpha_k g(w_k,\xi_k) \\
    \text{and}\ \ \tilde w_{k+1} &\gets \frac{1}{k+1} \sum_{j=1}^{k+1} w_j,
  \end{aligned}
\eequation
where the averaged sequence $\{\wtilde_k\}$ has no effect on the computation of the SG iterate sequence~$\{w_k\}$.  Early hopes were that this auxiliary averaged sequence might possess better convergence properties than the SG iterates themselves.  However, such improved behavior was found to be elusive when using classical stepsize sequences that diminished with a rate of $\Ocal(1/k)$ \cite{Poly77}.

A fundamental advancement in the use of iterate averaging came with the work of Polyak~\cite{Poly91}, which was subsequently advanced with Juditsky \cite{PolyJudi92}; see also the work of Ruppert \cite{Rupp88} and Nemirovski and Yudin \cite{NemiYudi78}.  Here, the idea remains to employ the iteration \eqref{polyak-ruppert}, but with stepsizes diminishing at a slower rate of $\Ocal(1/(k^a))$ for some $a \in (\thalf,1)$.  When minimizing strongly convex objectives, it follows from this choice that $\E[\|w_k - w_*\|_2^2] = \Ocal(1/(k^a))$ while $\E[\|\wtilde_k - w_*\|_2^2] = \Ocal(1/k)$.  What is interesting, however, is that in certain cases this combination of  \emph{long steps} and \emph{averaging} yields an optimal constant in the latter rate (for the iterate averages) in the sense that no rescaling of the steps---through multiplication with a positive definite matrix (see~\eqref{eq.sd_rescaled} in the next section)---can improve the asymptotic rate or constant.  This shows that, due to averaging, the adverse effects caused by ill-conditioning disappear.  (In this respect, the effect of averaging has been characterized as being similar to that of using a second-order method in which the Hessian approximations approach the Hessian of the objective at the minimizer \cite{Poly91,Bott10}; see \S\ref{sec.qn_stochastic}.)  This asymptotic behavior is difficult to achieve in practice, but is possible in some circumstances with careful selection of the stepsize sequence \cite{Xu11}.


Iterate averaging has since been incorporated into various schemes in order to allow longer steps while maintaining desired rates of convergence.  Examples include the \emph{robust SA} and \emph{mirror descent SA} methods presented in \cite{NemiJudiLanShap09}, as well as Nesterov's \emph{primal-dual averaging} method proposed in \cite[\S6]{Nest09}.  This latter method is notable in this section as it employs gradient aggregation and yields a $\Ocal(1/k)$ rate of convergence for the averaged iterate sequence.



\section{Second-Order Methods}\label{sec.second_order}
\setcounter{equation}{0}
\setcounter{theorem}{0}
\setcounter{algorithm}{0}
\setcounter{figure}{0}
\setcounter{table}{0}

In \S\ref{sec.noise_reduction}, we looked beyond classical SG to methods that are less affected by noise in the stochastic directions.  Another manner to move beyond classical SG is to address the adverse effects of high nonlinearity and ill-conditioning of the objective function through the use of second-order information.  As we shall see, these methods improve convergence rates of batch methods or the constants involved in the sublinear convergence rate of stochastic methods.

A common way to motivate second-order algorithms is to observe that first-order methods, such as SG and the full gradient method, are not \emph{scale invariant}.  Consider, for example, the full gradient iteration for minimizing a continuously differentiable function $F : \R{d} \to \R{}$, namely
\bequation\label{eq.sd_unscaled}
  w_{k+1} \gets w_k - \alpha_k \nabla F(w_k).
\eequation
An alternative iteration is obtained by applying a full gradient approach after a linear transformation of variables, i.e., by considering $\min_{\wtilde} F(B\wtilde)$ for some symmetric positive definite matrix~$B$.  The full gradient iteration for this problem has the form
\bequationn
  \wtilde_{k+1} \gets \wtilde_k - \alpha_k B \nabla F(B\wtilde_k),
\eequationn
which, after scaling by $B$ and defining $\{w_k\} := \{B\wtilde_k\}$, corresponds to the iteration
\bequation\label{eq.sd_rescaled}
  w_{k+1} \gets w_k - \alpha_k B^2 \nabla F(w_k).
\eequation
Comparing \eqref{eq.sd_rescaled} with \eqref{eq.sd_unscaled}, it is clear that the behavior of the algorithm will be different under this change of variables.  For instance, when $F$ is a strongly convex quadratic with unique minimizer~$w_*$, the full gradient method \eqref{eq.sd_unscaled} generally requires many iterations to approach the minimizer, but from any initial point~$w_1$ the iteration~\eqref{eq.sd_rescaled} with $B = (\nabla^2 F(w_1))^{-1/2}$ and $\alpha_1 = 1$ will yield $w_2 = w_*$.  These latter choices correspond to a single iteration of \emph{Newton's method} \cite{DennSchn96}.  In general, it is natural to seek transformations that perform well in theory and in practice. 

Another motivation for second-order algorithms comes from the observation that each iteration of the form \eqref{eq.sd_unscaled} 
or \eqref{eq.sd_rescaled} chooses the subsequent iterate by first computing the minimizer of a second-order Taylor series approximation $q_k : \R{d} \to \R{}$ of~$F$ at $w_k$, which has the form
\bequation\label{eq.q_model}
  q_k(w) = F(w_k) + \nabla F(w_k)^T(w-w_k) + \tfrac{1}{2} (w- w_k)^T B^{-2} (w - w_k).
\eequation
The full gradient iteration corresponds to $B^{-2} = I$ while Newton's method corresponds to $B^{-2} = \nabla^2 F(w_k)$ (assuming this Hessian is positive definite).  Thus, in general, a full gradient iteration works with a model that is only first-order accurate while Newton's method applies successive local re-scalings based on minimizing an exact second-order Taylor model of $F$ at each iterate.

Deterministic (i.e., batch) methods are known to benefit from the use of second-order information; e.g., Newton's method achieves a quadratic rate of convergence if $w_1$ is sufficiently close to a strong minimizer \cite{DennSchn96}.  On the other hand, stochastic methods like the SG method in \S\ref{sec.sg} cannot achieve a convergence rate that is faster than sublinear, regardless of the choice of $B$; see \cite{AgarBartRaviWain12,Mura98}.  (More on this in \S\ref{sec.qn_stochastic}.) Therefore, it is natural to ask: can there be a benefit to incorporating second-order information in stochastic methods?  We address this question throughout this section by showing that the careful use of successive re-scalings based on (approximate) second-order derivatives can be beneficial between the stochastic and batch regimes.

We begin this section by considering a \emph{Hessian-free Newton} method that employs exact second-order information, but in a judicious manner that exploits the stochastic nature of the objective function.  We then describe methods that attempt to mimic the behavior of a Newton algorithm through first-order information computed over sequences of iterates; these include \emph{quasi-Newton}, \emph{Gauss-Newton}, and related algorithms that employ only diagonal re-scalings. We also discuss the \emph{natural gradient} method, which defines a search direction in the space of realizable distributions, rather than in the space of the real parameter vector $w$.   Whereas Newton's method is invariant to linear transformations of the variables, the natural gradient method is invariant with respect to more general invertible transformations.

We depict the methods of interest in this section on the downward axis illustrated in Figure~\ref{view3}.  We use double-sided arrows for the methods that can be effective throughout the spectrum between the stochastic and batch regimes.  Single-sided arrows are used for those methods that one might consider to be effective only with at least a moderate batch size in the stochastic gradient estimates.  We explain these distinctions as we describe the methods.

\bfigure[ht]
  \center
  \includegraphics[width=0.7\linewidth,clip=true,trim=10 55 15 55]{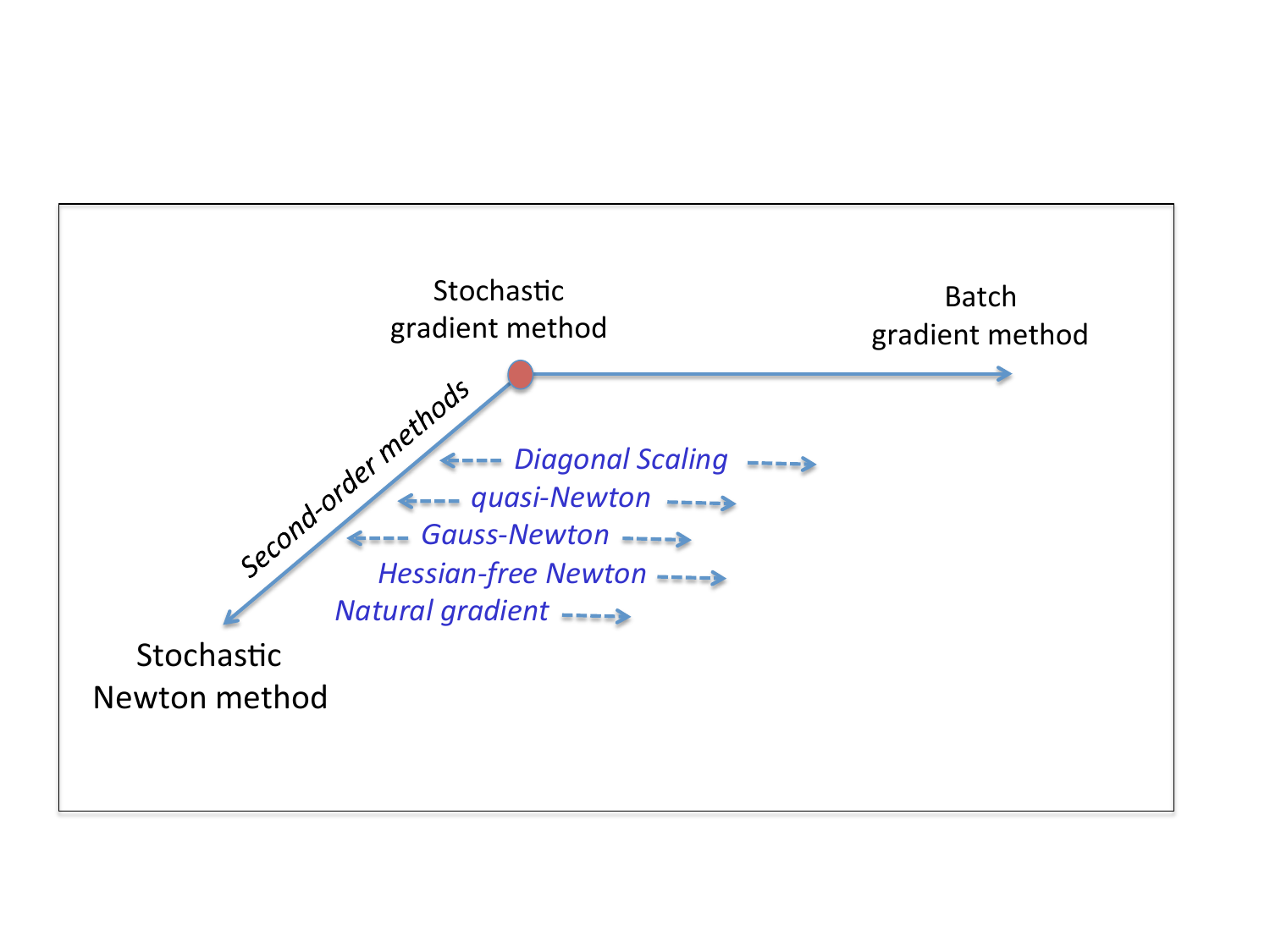}
  \caption{View of the schematic from Figure~\ref{view1} with a focus on second-order methods.  The dotted arrows indicate the effective regime of each method: the first three methods can employ mini-batches of any size, whereas the last two methods are efficient only for moderate-to-large mini-batch sizes.}
  \label{view3}
\efigure

\subsection{Hessian-Free Inexact Newton Methods}\label{sec:hfn}

Due to its scale invariance properties and its ability to achieve a quadratic rate of convergence in the neighborhood of a strong local minimizer, Newton's method represents an ideal in terms of optimization algorithms.  It does not scale well with the dimension $d$ of the optimization problem, but there are variants that can scale well while also being able to deal with nonconvexity.

When minimizing a twice-continuously differentiable $F$, a Newton iteration is
\bsubequations\label{newtoncg}
  \begin{align}
    w_{k+1} &\gets w_k + \alpha_k s_k, \\
    \text{where}\   s_k  \ \mbox{satisfies}\ \nabla^2 F(w_k) s_k &= -\nabla F(w_k). \label{eq.newton_system}
  \end{align}
\esubequations
This iteration demands much in terms of computation and storage. 
However, rather than solve the Newton system \eqref{eq.newton_system} exactly through matrix factorization techniques, one can instead only solve it inexactly through an iterative approach such as the conjugate gradient (CG) method \cite{GoluVanL12}.  
By ensuring that the linear solves are accurate enough, such an \emph{inexact Newton-CG} method can enjoy a superlinear rate of convergence \cite{DembEiseStei82}.

In fact, the computational benefits of inexact Newton-CG go beyond its ability to maintain classical convergence rate guarantees.  Like many iterative linear system techniques, CG applied to \eqref{eq.newton_system} does not require access to the Hessian itself, only Hessian-vector products \cite{Pear94}.  It is in this sense that such a method may be called \emph{Hessian-free}.  This is ideal when such products can be coded directly without having to form an explicit Hessian, as Example~\ref{ex.hess_free} below demonstrates.  Each product is at least as expensive as a gradient evaluation, but as long as the number of products---one per CG iteration---is not too large, the improved rate of convergence can compensate for the extra per-iteration work required over a simple full gradient method.  

\begin{example}\label{ex.hess_free}
Consider the function of the parameter vector $w = (w_1,w_2)$ given by $F(w)= \exp(w_1w_2)$. Let us define, for any $d \in \R{2}$,  the function
\[
     \phi(w;d)=  \nabla F(w)^Td= w_2 \exp(w_1 w_2) d_1 + w_1 \exp(w_1 w_2) d_2.
\]
Computing the gradient of $\phi$ with respect to $w$, we have
\[
      \nabla_w \phi(w;d)= \nabla^2 F(w) d=  
      \left[  \begin{array}{c}
        w_2^2 \exp(w_1 w_2) d_1 + (\exp(w_1 w_2) + w_1 w_2 \exp(w_1 w_2) )d_2  \\
        ( \exp (w_1 w_2) + w_1 w_2 \exp(w_1 w_2) )d_1 + w_1^2 \exp(w_1 w_2) d_2
        \end{array} \right].
\]
We have thus obtained, for any $d \in \R{2}$, a formula for computing $ \nabla^2 F(w) d$ that does not require $\nabla^2 F(w)$ explicitly. Note that by storing the scalars $w_1 w_2$ and $\exp(w_1 w_2)$ from the evaluation of $F$, the additional costs of computing the gradient-vector and Hessian-vector products are small. 
\end{example}

The idea illustrated in this example can be applied in general; e.g., see also Example~\ref{ex.hess_free_2} on page \pageref{ex.hess_free_2}. For a smooth objective function~$F$, one can compute $ \nabla^2 F(w) d$ at a cost that is a small multiple of the cost of evaluating $\nabla F$, and without forming the Hessian, which would require $\Ocal(d^2)$ storage. The savings in computation come at the expense of storage of some additional quantities, as explained in Example~\ref{ex.hess_free}. 

In  machine learning applications, including those involving multinomial logistic regression and deep neural networks, Hessian-vector products can be computed in this manner, and an inexact Newton-CG method can be applied.  A concern is that, in certain cases, the  cost of the CG iterations may render such a method uncompetitive with alternative approaches, such as an SG method or a limited memory BFGS method (see \S\ref{sec:stochqn}), which have small computational overhead.  Interestingly, however, the structure of the risk measures \eqref{eq.exp_risk_w} and \eqref{eq.emp_risk_w} can be exploited so that the resulting method has lighter computational overheads, as described next.

\subsubsection{Subsampled Hessian-Free Newton Methods}  \label{ssnewt}

The motivation for the method we now describe stems from the observation that, in inexact Newton methods, the Hessian matrix need not be as accurate as the gradient to yield an effective iteration. Translated to the context of large-scale machine learning applications, this means that the iteration is more tolerant to noise in the Hessian estimate than it is to noise in the gradient estimate.

Based on this idea, the technique we state here employs a smaller sample for defining the Hessian than for the stochastic gradient estimate.  Following similar notation as introduced in \S\ref{sec.dynamic_sample_size}, let the stochastic gradient estimate be
\bequationn
  \nabla f_{\Scal_k}(w_k;\xi_k) = \frac{1}{|\Scal_k|} \sum_{i\in\Scal_k} \nabla f(w_k;\xi_{k,i})
\eequationn
and let the stochastic Hessian estimate be
\bequation\label{eq:subsampledhessian}
  \nabla^2 f_{\Scal_k^H}(w_k;\xi_k^H) = \frac{1}{|\Scal_k^H|} \sum_{i\in\Scal_k^H} \nabla^2 f(w_k;\xi_{k,i}),
\eequation
where $\Scal_k^H$ is conditionally uncorrelated with $\Scal_k$ given $w_k$.  If one chooses the \emph{subsample} size $|\Scal_k^H|$ small enough, then the cost of each product involving the Hessian approximation can be reduced significantly, thus reducing the cost of each CG iteration. On the other hand, one should choose $|\Scal_k^H|$ large enough so that the  curvature information captured through the Hessian-vector products is productive.  If done appropriately, \emph{Hessian subsampling} is robust and effective \cite{agarwal2016second,byrd2011use,pilanci2015newton,roosta2016sub}.  An inexact Newton method that incorporates this techniques is outlined in Algorithm~\ref{newton-cg}.  The algorithm is stated with a backtracking (Armijo) line search \cite{NoceWrig06}, though other stepsize-selection techniques could be considered as well.  

\newlength{\continueindent}
\setlength{\continueindent}{0.2em}
\makeatletter
\newcommand*{\ALG@customparshape}{\parshape 2 \leftmargin \linewidth \dimexpr\ALG@tlm+\continueindent\relax \dimexpr\linewidth+\leftmargin-\ALG@tlm-\continueindent\relax}
\apptocmd{\ALG@beginblock}{\ALG@customparshape}{}{\errmessage{failed to patch}}
\makeatother

\balgorithm
  \caption{Subsampled Hessian-Free Inexact Newton Method}
  \label{newton-cg}
  \balgorithmic[1]
    \State Choose an initial iterate $w_1$.
    \State Choose constants $\rho \in (0,1)$, $\gamma \in (0,1)$, $\eta \in (0,1)$, and $\max_{cg} \in \N{}$.
    \For{$k=1,2,\dots$}
      \State Generate realizations of $\xi_k$ and $\xi_k^H$ corresponding to $\Scal_k$ and $\Scal_k^H$.
      \State \label{alg.newton-cg.step} Compute $s_k$ by applying Hessian-free CG to solve
      \bequation   \label{jamaica}
         \nabla^2 f_{\Scal_k^H}(w_k;\xi_k^H) s = -\nabla f_{\Scal_k}(w_k;\xi_k)
      \eequation
      until $\max_{cg}$ iterations have been performed or a trial solution yields
      \bequationn
        \|r_k\|_2 := \|\nabla^2 f_{\Scal_k^H}(w_k;\xi_k^H) s + \nabla f_{\Scal_k}(w_k;\xi_k)\|_2 \leq \rho\|\nabla f_{\Scal_k}(w_k;\xi_k)\|_2.
      \eequationn
      \State Set $w_{k+1} \gets w_k + \alpha_k s_k$, where $\alpha_k \in \{\gamma^0,\gamma^1,\gamma^2,\dots\}$ is the largest element with
      \bequation\label{gull}
        f_{\Scal_k}(w_{k+1};\xi_k) \leq f_{\Scal_k}(w_k;\xi_k) + \eta \alpha_k \nabla f_{\Scal_k}(w_k;\xi_k)^Ts_k.
      \eequation
    \EndFor
  \end{algorithmic}
\end{algorithm}

As previously mentioned, the (subsampled) Hessian-vector products required in Algorithm~\ref{newton-cg} can be computed efficiently in the context of many machine learning applications.  For instance, one such case in the following.
\begin{example}\label{ex.hess_free_2}
Consider a binary classification problem where the training  function is given by the logistic loss  with an $\ell_2$-norm regularization parameterized by $\lambda > 0$:
\begin{equation} \label{logistic-loss}
  R_n(w)=\frac{1}{n} \sum_{i=1}^{n}\log(1 + \exp(-y_iw^Tx_i)) + \frac{\lambda}{2}\|w\|^2.
\end{equation} 
A (subsampled) Hessian-vector product can be computed efficiently by observing that
\begin{align*}
 \nabla^2 f_{\Scal_k^H}(w_k;\xi_k^H)d = \frac{1}{|\Scal_k^H|} \sum_{i\in\Scal_k^H}\frac{\exp(-y_iw^Tx_i)}{(1 + \exp(-y_iw^Tx_i))^2}(x_i^Td)x_i + \lambda d .
\end{align*} 
\end{example}

To quantify the step computation cost in an inexact Newton-CG framework such as Algorithm~\ref{newton-cg}, let ${\rm g}_{cost}$ be the cost of computing a gradient estimate $\nabla f_{\Scal_k}(w_k;\xi_k)$ and factor$\times {\rm g}_{cost}$ denote the cost of one Hessian-vector product.  If the maximum number of CG iterations, $\textstyle\max_{cg}$,  is performed for every outer iteration, then the step computation cost in Algorithm~\ref{newton-cg} is
\bequationn
  \textstyle\max_{cg} \times {\rm factor} \times {\rm g}_{cost} + {\rm g}_{cost}.
\eequationn
In a deterministic inexact Newton-CG method for minimizing the empirical risk $R_n$, i.e., when $|\Scal_k^H| = |\Scal_k| = n$ for all $k \in \N{}$, the factor is at least 1 and $\max_{cg}$ would typically be chosen as 5, 20, or more, leading to an iteration that is many times the cost of an SG iteration.  However, in a stochastic framework using Hessian sub-sampling, the factor can be chosen to be sufficiently small such that $\max_{cg} \times {\rm factor} \approx 1$, leading to a per-iteration cost proportional to that of SG.

Implicit in this discussion is the assumption that the gradient sample size $|\Scal_k|$ is large enough so that taking subsamples for the Hessian estimate is sensible.  If, by contrast, the algorithm were to operate in the stochastic regime of SG where $|\Scal_k|$ is small and gradients are very noisy, then it may be necessary to choose $|\Scal_k^H| > |\Scal_k|$ so that Hessian approximations do not corrupt the step. In such circumstances, the method would be far less attractive than SG. Therefore, the subsampled Hessian-free Newton method outlined here is only recommended when $\Scal_k$ is large.  This is why, in Figure~\ref{view3}, the Hessian-free Newton method is illustrated only with an arrow to the right, i.e., in the direction of larger sample sizes.

Convergence of Algorithm~\ref{newton-cg} is easy to establish when minimizing a strongly convex empirical risk measure $F= R_n$ when $\Scal_k \gets \{1,\dots,n\}$ for all $k \in \N{}$, i.e., when full gradients are always used.  In this case, a benefit of employing CG to solve \eqref{jamaica} is that it immediately improves upon the direction employed in a steepest descent iteration.  Specifically, when initialized at zero, it produces in its first iteration a scalar multiple of the steepest descent direction $-\nabla F(w_k)$, and further iterations monotonically improve upon this step (in terms of minimizing a quadratic model of the form in \eqref{eq.q_model}) until the Newton step is obtained, which is done so in at most $d$ iterations of CG (in exact arithmetic).  Therefore, by using any number of CG iterations, convergence can be established using standard techniques to choose the stepsize $\alpha_k$ \cite{NoceWrig06}.  When exact Hessians are also used, the rate of convergence can be controlled through the accuracy with which the systems \eqref{eq.newton_system} are solved.  Defining $r_k := \nabla^2 F(w_k) s_k + \nabla F(w_k)$ for all $k \in \N{}$, the iteration can enjoy a linear, superlinear, or quadratic rate of convergence by controlling $\|r_k\|_2$, where for the superlinear rates one must have $\{\|r_k\|_2/\|\nabla F(w_k)\|_2\} \to 0$ \cite{DembEiseStei82}.

When the Hessians are subsampled (i.e., $\Scal_k^H \subset \Scal_k$ for all $k \in \N{}$), it has not been shown that the rate of convergence is faster than linear; nevertheless, the reduction in the number of iterations required to produce a good approximate solution is often significantly lower than if no Hessian information is used in the algorithm.


\subsubsection{Dealing with Nonconvexity}\label{notcon}

Hessian-free Newton methods are routinely applied  for the solution of nonconvex problems.  In such cases, it is common to employ a \emph{trust region} \cite{ConnGoulToin00} instead of a line search and to introduce an additional condition in Step~\ref{alg.newton-cg.step} of Algorithm~\ref{newton-cg}: terminate CG if a candidate solution $s_k$ is a direction of negative curvature, i.e., $s_k^T \nabla^2 f_{\Scal_k^H}(w_k;\xi_k^H)s_k < 0$ \cite{Stei83}.  A number of more sophisticated strategies have been proposed throughout the years with some success, but none have proved to be totally satisfactory or universally accepted. 

Instead of coping with indefiniteness, one can focus on strategies for ensuring positive (semi)definite Hessian approximations. One of the most attractive ways of doing this in the context of machine learning is to employ a (subsampled) Gauss-Newton approximation to the Hessian, which is a matrix of the form
\bequation\label{eq.gn_prelim}
  G_{\Scal_k^H}(w_k;\xi_{k}^H)= \frac{1}{|\Scal_k^H|} \sum_{i \in \Scal_k^H} J_h(w_k,\xi_{k,i})^T H_\ell(w_k,\xi_{k,i}) J_h(w_k,\xi_{k,i}).
\eequation
Here, the matrix $J_h$ captures the stochastic gradient information for the prediction function $h(x;w)$, whereas the matrix $H_\ell$ only captures the second order information for the (convex) loss function $\ell(h,y)$; see~\S\ref{sec:gauss} for a detailed explanation.  As before, one can directly code the product of this matrix times a vector without forming the matrix components explicitly.  This approach has been applied successfully in the training of deep neural networks~\cite{BeckLeCu89,Mart11}.

We mention in passing that there has been much discussion about the role that negative curvature and saddle points play in the optimization of deep neural networks; see e.g., \cite{DaupPasc14,GoodViny14,choromanska2015loss}.  Numerical tests designed to probe the geometry of the objective function in the neighborhood of a minimizer when training a deep neural network have shown the presence of negative curvature.  It is believed that the inherent stochasticity of the SG method allows it to navigate efficiently through this complex landscape, but it is not known whether classical techniques to avoid approaching saddle points will prove to be successful for either batch or stochastic methods.

\subsection{Stochastic Quasi-Newton Methods}\label{sec:stochqn}

One of the most important developments in the field of nonlinear optimization came with the advent of \emph{quasi-Newton} methods.
These methods construct approximations to the Hessian using only gradient information, and are applicable for convex and nonconvex problems. Versions that scale well with the number of variables, such as \emph{limited memory} methods, have proved to be effective in a wide range of applications  where the number of variables can be in the millions. It is therefore natural to ask whether quasi-Newton methods can be extended to the stochastic setting arising in machine learning. Before we embark on this discussion, let us review the basic principles underlying quasi-Newton methods, focusing on the most popular scheme, namely BFGS \cite{Broy70,Flet70,Gold70,Shan70}.

In the spirit of  Newton's method \eqref{newtoncg}, the  BFGS iteration for minimizing a twice continuously differentiable function $F$ has the form
\bequation\label{eq.bfgs}
  w_{k+1} \gets w_k - \alpha_k H_k \nabla F(w_k),
\eequation
where $H_k$ is a symmetric positive definite approximation of $(\nabla^2 F(w_k))^{-1}$.  This form of the iteration is consistent with \eqref{newtoncg}, but the signifying feature of a quasi-Newton scheme is that the sequence $\{H_k\}$ is updated dynamically by the algorithm rather than through a second-order derivative computation at each iterate.  Specifically, in the BFGS method, the new inverse Hessian approximation is obtained by defining the iterate and gradient displacements
\bequationn
  s_k := w_{k+1} - w_k\ \ \text{and}\ \ v_k := \nabla F(w_{k+1}) - \nabla F(w_k),
\eequationn
then setting
\bequation\label{eq.bfgs_update}
  H_{k+1} \gets \(I - \frac{v_ks_k^T}{s_k^Tv_k}\)^TH_k\(I - \frac{v_ks_k^T}{s_k^Tv_k}\) + \frac{s_ks_k^T}{s_k^Tv_k}.
\eequation
One important aspect of this update is that it ensures that the secant equation $H_{k+1}^{-1}s_k = v_k$ holds, meaning that a second-order Taylor expansion is satisfied along the most recent displacement (though not necessarily along other directions).

A remarkable fact about the BFGS iteration \eqref{eq.bfgs}--\eqref{eq.bfgs_update} is that it enjoys a local superlinear rate of convergence \cite{DennMore74}, and this with only first-order information and without the need for any linear system solves (which are required by Newton's method for it to be quadratically convergent).  However, a number of issues need to be addressed to have an effective method in practice.  For one thing, the update \eqref{eq.bfgs_update} yields dense matrices, even when the exact Hessians are sparse, restricting its use to small and midsize problems.  A common solution for this  is to employ a \emph{limited memory} scheme, leading to a method such as L-BFGS \cite{LiuNoce89,Noce80}.   A key feature of this approach is that the matrices $\{H_k\}$ need not be formed explicitly; instead, each product of the form $H_k \nabla F(w_k)$ can be computed using a formula that only requires recent elements of the sequence of displacement pairs $\{(s_k,v_k)\}$ that have been saved in storage. Such an approach incurs per-iteration costs of order $\Ocal(d)$, and delivers practical performance that is significantly better than an full gradient iteration, though the rate of convergence is only provably linear.

\subsubsection{Deterministic to Stochastic}\label{sec.qn_stochastic}

Let us consider the extension of a quasi-Newton approach from the deterministic to the stochastic setting arising in machine learning. The iteration now takes the form
\begin{equation}\label{directqn}
     w_{k+1} \gets w_k - \alpha_k H_k g(w_k,\xi_k).
\end{equation}
Since we are interested in large-scale problems, we assume that \eqref{directqn} implements an L-BFGS scheme, which avoids the explicit construction of $H_k$. A number of questions arise when considering~\eqref{directqn}.  We list them now with some proposed solutions.

\paragraph{\textbf{Theoretical Limitations}}

The convergence rate of a stochastic iteration such as \eqref{directqn} cannot be faster than sublinear \cite{AgarBartRaviWain12}. Given that SG also has a sublinear rate of convergence, what benefit, if any, could come from incorporating $H_k$ in \eqref{directqn}?  This is an important question.  As it happens, one can see a benefit of $H_k$ in terms of the \emph{constant} that appears in the sublinear rate.  Recall that for the SG method (Algorithm~\ref{alg.sg}), the constant depends on $L/c$, which in turn depends on the conditioning of $\{\nabla^2 F(w_k)\}$.  This is typical of first-order methods.  In contrast,  one can show \cite{BottLeCu04} that if the sequence of Hessian approximations in \eqref{directqn} satisfies $\{H_k\} \rightarrow \nabla^2 F(w_*)^{-1}$, then the constant is independent of the conditioning of the Hessian. Although constructing Hessian approximations with this property might not be viable in practice, this fact suggests that  stochastic quasi-Newton methods could be better equipped to cope with ill-conditioning than SG. 

\paragraph{\textbf{Additional Per-Iteration Costs}}

The SG iteration is very inexpensive, requiring only the evaluation of $g(w_k, \xi_k)$.  The iteration \eqref{directqn}, on the other hand, also requires the product $H_k g(w_k, \xi_k)$, which is known to require $4md$ operations where $m$ is the \emph{memory} in the L-BFGS updating scheme. Assuming for concreteness that the cost of evaluating $g(w_k, \xi_k)$ is exactly $d$ operations (using only one sample) and  that the memory parameter is set to the typical value of $m=5$, one finds that the stochastic quasi-Newton method is 20 times more expensive than SG.  Can the iteration \eqref{directqn} yield fast enough progress as to offset this additional per-iteration cost?  To address this question, one need only observe that the calculation just mentioned focuses on the gradient $g(w_k, \xi_k)$ being based on a single sample. However, when employing mini-batch gradient estimates, the additional cost of the iteration \eqref{directqn} is only marginal. (Mini-batches of size 256 are common in practice.) The use of mini-batches may therefore be considered essential when one contemplates the use of a stochastic quasi-Newton method. This mini-batch need not be large, as in the Hessian-free Newton method discussed in the previous section, but it should not be less than, say, 20 or 50 in light of the additional costs of computing the matrix-vector products.

\paragraph{\textbf{Conditioning of the Scaling Matrices}}

The BFGS formula \eqref{eq.bfgs_update} for updating~$H_k$ involves differences in gradient estimates computed in consecutive iterations. In stochastic settings, the gradients $\{g(w_k,\xi_k)\}$ are noisy estimates of $\{\nabla F(w_k)\}$.  This can cause the updating process to yield poor curvature estimates, which may have a detrimental rather than beneficial effect on the quality of the computed steps.  Since BFGS, like all quasi-Newton schemes, is an \emph{overwriting} process, the effects of even a single bad update may linger for numerous iterations.  How could such effects be avoided in the stochastic regime?  There have been various proposals to avoid differencing noisy gradient estimates.  One possibility is to employ the same sample when computing gradient differences \cite{BordBottGall09,SchrYuGuen07}.  An alternative approach that allows greater freedom in the choice of the stochastic gradient is to \emph{decouple} the step computation \eqref{directqn} and the Hessian update.  In this manner, one can employ a larger sample, if necessary, when computing the gradient displacement vector.  We discuss these ideas further in \S\ref{sec.qn_algorithms}.
  
It is worthwhile to note that if the gradient estimate $g(w_k,\xi_k)$ does not have high variance, then standard BFGS updating can be applied without concerns. Therefore, in the rest of this section, we focus on algorithms that employ noisy gradient estimates in the step computation \eqref{directqn}. This means, e.g., that we are not considering the potential to tie the method with noise reduction techniques described in \S\ref{sec.noise_reduction}, though such an idea is natural and could be effective in practice. 
  
\subsubsection{Algorithms}\label{sec.qn_algorithms}

A straightforward adaptation of L-BFGS  involves only the replacement of deterministic gradients with stochastic gradients throughout the iterative process.  The displacement pairs might then be defined as
\begin{equation}\label{cambert}
  s_k := w_{k+1} - w_k\ \ \text{and}\ \ v_k := \nabla f_{\Scal_k}(w_{k+1},\xi_k) - \nabla f_{\Scal_k}(w_k, \xi_k).
\end{equation}
Note the use of the same seed $\xi_k$ in the two gradient estimates, in order to address the issues related to noise mentioned above.   If each $f_i$ is strongly convex, then $s_k^Tv_k > 0$, and positive definiteness of the updates is also maintained.  Such an approach is sometimes referred to as \emph{online L-BFGS} \cite{SchrYuGuen07,MokhRibe14}.

One disadvantage of this method is the need to compute two, as opposed to only one, gradient estimate per iteration: one to compute the gradient displacement (namely, $g(w_{k+1},\xi_k))$ and another (namely, $g(w_{k+1},\xi_{k+1}))$ to compute the subsequent step.  This is not too onerous, at least as long as the per-iteration improvement outweighs the extra per-iteration cost. A more worrisome feature is that updating the inverse Hessian approximations with \emph{every} step may not be warranted, and may even be detrimental when the gradient displacement is based on a small sample, since it could easily represent a poor approximation of the action of the true Hessian of $F$.

An alternative strategy, which might better represent the action of the true Hessian even when $g(w_k,\xi_k)$ has high variance, is to employ an alternative $v_k$.  In particular, 
since  $\nabla F(w_{k+1}) - \nabla F(w_k) \approx \nabla^2 F(w_k)(w_{k+1} - w_k)$, one could define
\bequation  \label{bowl}
  v_k := \nabla^2 f_{\Scal_k^H}(w_k;\xi_k^H) s_k,
\eequation
where $\nabla^2 f_{\Scal_k^H}(w_k;\xi_k^H)$ is a subsampled Hessian and $| \Scal_k^H |$  is large enough to provide useful curvature information. As in the case of Hessian-free Newton from \S\ref{ssnewt}, the product \eqref{bowl} can be performed without explicitly constructing $\nabla^2 f_{\Scal_k^H}(w_k;\xi_k^H)$.
 
Regardless of the definition of $v_k$, when $| \Scal_k^H |$ is much larger than $| \Scal_k |$, the cost of quasi-Newton updating is excessive due to the cost of computing $v_k$. To address this issue, the computation of $v_k$ can be performed only after a sequence of iterations, so as to amortize costs. This leads to the idea of decoupling the step computation from the quasi-Newton update. This approach, which we refer to for convenience as SQN, performs a sequence of iterations of \eqref{directqn} with $H_k$ fixed, then  computes a new displacement pair $(s_k,v_k)$ with $s_k$ defined as in \eqref{cambert} and $v_k$ set using one of the strategies outlined above. This pair replaces one of the old pairs in storage, which in turn defines the limited memory BFGS step.
 
To formalize all of these alternatives, we state the general stochastic quasi-Newton method presented as Algorithm~\ref{alg.sqn}, with some notation borrowed from Algorithm~\ref{newton-cg}.  In the method, the step computation is based on a collection of $m$ displacement pairs $\Pcal = \{s_j,v_j\}$ in storage and the current stochastic gradient $\nabla f_{\Scal_k}(w_k;\xi_k)$, where the matrix-vector product in \eqref{directqn} can be computed through a \emph{two-loop recursion} \cite{Noce80,NoceWrig06}.  To demonstrate the generality of the method, we note that the \emph{online L-BFGS} method sets $\Scal_k^H \gets \Scal_k$ and {\tt update pairs = true} in every iteration.  In \emph{SQN} using \eqref{bowl}, on the other hand, $|\Scal_k^H |$ should be chosen larger than $| \Scal_k|$ and one sets {\tt update pairs = true} only every, say, 10 or 20 iterations.

\begin{algorithm}
  \caption{Stochastic Quasi-Newton Framework}
  \label{alg.sqn}
  \begin{algorithmic}[1]
    \State Choose an initial iterate $w_1$ and initialize $\Pcal \gets \emptyset$.
    \State Choose a constant $m \in \N{}$.
    \State Choose a stepsize sequence $\{\alpha_k\} \subset \R{}_{++}$.
    \For{$k=1,2,\dots,$}
      \State Generate realizations of $\xi_k$ and $\xi_k^H$ corresponding to $\Scal_k$ and $\Scal_k^H$.
      \State Compute $\shat_k = H_kg(w_k,\xi_k)$ using the two-loop recursion based on the set $\Pcal$.
      \State Set $s_k \gets - \alpha_k \shat_k$.
      \State Set $w_{k+1} \gets w_k + s_k$.
      \If{\texttt{update pairs}}
        \State Compute $s_k$ and $v_k$ (based on the sample $\Scal_k^H$).
        \State Add the new displacement pair $(s_k,v_k)$ to $\Pcal$.
        \State If $|\Pcal| > m$, then remove eldest pair from $\Pcal$.
      \EndIf
    \EndFor
  \end{algorithmic}
\end{algorithm}

To guarantee that the BFGS update is well defined, each displacement pair $(s_j,v_j)$ must satisfy $s_j^Tv_j >0$.  In deterministic optimization, this issue is commonly addressed by either performing a line search (involving exact gradient computations) or modifying the displacement vectors (e.g., through \emph{damping}) so that $s_j^Tv_j > 0$, in which case one does ensure that \eqref{eq.bfgs_update} maintains positive definite approximations.  However, these mechanisms have not been fully developed in the stochastic regime when exact gradient information is unavailable and the gradient displacement vectors are noisy.  Simple ways to overcome these difficulties is to replace the Hessian matrix with a Gauss-Newton approximation or to introduce a combination of damping and regularization (say, through the addition of simple positive definite matrices).


There remains much to be explored in terms of stochastic quasi-Newton methods for machine learning applications.  Experience has shown that some gains in performance can be achieved, but the full potential of the quasi-Newton schemes discussed above (and potentially others) is not yet known. 

\subsection{Gauss-Newton Methods}\label{sec:gauss}

The Gauss-Newton method is a classical approach for nonlinear least squares, i.e., minimization problems in which the objective function is a sum of squares.  This method readily applies for optimization problems arising in machine learning involving a least squares loss function, but the idea generalizes for other popular loss functions as well.  The primary advantage of Gauss-Newton is that it constructs an approximation to the Hessian using only first-order information, and this approximation is guaranteed to be positive semidefinite, even when the full Hessian itself may be indefinite. The price to pay for this convenient representation is that it ignores second-order interactions between elements of the parameter vector $w$, which might mean a loss of curvature information that could be useful for the optimization process.

\paragraph{\textbf{Classical Gauss-Newton}}

Let us introduce the classical Gauss-Newton approach by considering a situation in which, for a given input-output pair $(x,y)$, the loss incurred by a parameter vector~$w$ is measured via a squared norm discrepancy between $h(x;w)\in\R{d}$ and $y\in\R{d}$.  Representing the input-output pair being chosen randomly via the subscript $\xi$, we may thus write
\bequationn
    f(w;\xi) = \ell(h(x_\xi;w),y_\xi) = \thalf \|h(x_\xi;w) - y_\xi\|_2^2.
\eequationn
Writing a second-order Taylor series model of this function in the vicinity of parameter vector $w_k$ would involve its gradient and Hessian at $w_k$, and minimizing the resulting model (recall \eqref{eq.q_model}) would lead to a Newton iteration.  Alternatively, a Gauss-Newton approximation of the function is obtained by making an affine approximation of the prediction function inside the quadratic loss function. Letting $J_h(\cdot;\xi)$ represent the Jacobian of $h(x_\xi;\cdot)$ with respect to $w$, we have the approximation
\[
    h(x_\xi; w) \approx h(x_\xi;w_k) + J_h(w_k;\xi)(w-w_k),
\]
which leads to
\bequalin
   f(w;\xi)
    &\approx \thalf \|h(x_\xi;w_k) + J_h(w_k;\xi)(w - w_k) - y_\xi\|_2^2 \\
    &= \thalf \|h(x_\xi;w_k) - y_\xi\|_2^2 + (h(x_\xi;w_k) - y_\xi)^T J_h(w_k;\xi) (w - w_k) \\
    &+ \thalf (w-w_k)^T J_h(w_k;\xi)^T J_h(w_k;\xi) (w-w_k).
\eequalin
In fact, this approximation is similar to a second-order Taylor series model, except that the terms involving the second derivatives of the prediction function $h$ with respect to the parameter vector have been dropped. The remaining second-order terms are those resulting from the positive curvature of the quadratic loss $\ell$. This leads to replacing the subsampled Hessian matrix \eqref{eq:subsampledhessian} by the Gauss-Newton matrix
\bequation
\label{eq:gaussnewtonmatrix}
   G_{\Scal_k^H}(w_k;\xi_{k}^H) = \frac{1}{|\Scal_k^H|} \sum_{i \in \Scal_k^H} J_h(w_k;\xi_{k,i})^T J_h(w_k;\xi_{k,i}) \,.
\eequation
Since the Gauss-Newton matrix only differs from the true Hessian by terms that involve the factors $h(x_\xi;w_k)-y_\xi$, these two matrices are the same when the loss is equal to zero, i.e., when $h(x_\xi;w_k)=y_\xi$.

A challenge in the application of a Gauss-Newton scheme is that the Gauss-Newton matrix is often singular or nearly singular.  In practice, this is typically handled by regularizing it by adding to it a positive multiple of the identity matrix. For least-squares loss functions, the inexact Hessian-free Newton methods of \S\ref{sec:hfn} and the stochastic quasi-Newton methods of \S\ref{sec:stochqn} with gradient displacement vectors defined as in \eqref{bowl} can be applied with (regularized) Gauss-Newton approximations.  This has the benefit that the scaling matrices are guaranteed to be positive definite.

The computational cost of the Gauss-Newton method depends on the dimensionality of the prediction function. When the prediction function is scalar-valued, the Jacobian matrix $J_h$ is a single row whose elements are already being computed as an intermediate step in the computation of the stochastic gradient $\nabla f(w;\xi)$.  However, this is no longer true when the dimensionality is larger than one since then computing the stochastic gradient vector $\nabla f(w;\xi)$ does not usually require the explicit computation of all rows of the Jacobian matrix.  This happens, for instance, in deep neural networks when one uses back propagation \cite{RumeHintWill86a,RumeHintWill86b}.

\paragraph{\textbf{Generalized Gauss-Newton}}

Gauss-Newton ideas can also be generalized for other standard loss functions \cite{Schr01}.  To illustrate, let us consider a slightly more general situation in which loss is measured by a composition of an arbitrary convex loss function $\ell(h,y)$ and a prediction function $h(x;w)$. Combining the affine approximation of the prediction function $h(x_\xi;w)$ with a second order Taylor expansion of the loss function $\ell$ leads to the generalized Gauss-Newton matrix
\bequation
\label{eq:ggnmatrix}
   G_{\Scal_k^H}(w_k;\xi_{k}^H) = \frac{1}{|\Scal_k^H|} \sum_{i \in \Scal_k^H} J_h(w_k;\xi_{k,i})^T H_\ell(w_k;\xi_{k,i}) \: J_h(w_k;\xi_{k,i})
\eequation
(recall~\eqref{eq.gn_prelim}), where $H_\ell(w_k;\xi)=\frac{\partial^2\ell}{\partial h^2}(h(x_\xi;w_k),y_\xi)$ captures the curvature of the loss function $\ell$.  This can be seen as a generalization of \eqref{eq:gaussnewtonmatrix} in which $H_\ell = I$.

When training a deep neural network, one may exploit this generalized strategy by redefining~$\ell$ and~$h$ so that as much as possible of the network's computation is formally performed by~$\ell$ rather than by~$h$.  If this can be done in such a way that convexity of $\ell$ is maintained, then one can faithfully capture second-order terms for $\ell$ using the generalized Gauss-Newton scheme.  Interestingly, in many useful situations, this strategy gives simpler and more elegant expressions for $H_\ell$.  For instance, probability estimation problems often reduce to using logarithmic losses of the form $f(w;\xi) = -\log(h(x_\xi;w))$. The generalized Gauss-Newton matrix then reduces to 
\begin{align}
G_{\Scal_k^H}(w_k;\xi_{k}^H)
&= \frac{1}{|\Scal_k^H|} \sum_{i \in \Scal_k^H} J_h(w_k;\xi_{k,i})^T \frac{1}{h(w;\xi_{k,i})^2} \: J_h(w_k;\xi_{k,i})
\nonumber\\
&= \frac{1}{|\Scal_k^H|} \sum_{i \in \Scal_k^H} \nabla f(w;\xi_{k,i}) \nabla f(w;\xi_{k,i})^T, 
\label{eq:ggnforlogloss} 
\end{align}
which does not require explicit computation of the Jacobian $J_h$.

\subsection{Natural Gradient Method}\label{sec:natural}

We have seen that Newton's method is invariant to \emph{linear} transformations of the parameter vector~$w$.  By contrast, the natural gradient method \cite{Amar98,AmarNaga97} aims to be invariant with respect to all differentiable and invertible transformations. The essential idea consists of formulating the gradient descent algorithm in the space of prediction functions rather than specific parameters.  Of course, the actual computation takes place with respect to the parameters, but accounts for the anisotropic relation between the parameters and the decision function.  That is, in parameter space, the natural gradient algorithm will move the parameters more quickly along directions that have a small impact on the decision function, and more cautiously along directions that have a large impact on the decision function.

We remark at the outset that many authors \cite{ParkAmar00,CaroOlli16} propose \emph{quasi-natural-gradient} methods that are strikingly similar to the \emph{quasi-Newton} methods described in \S\ref{sec:stochqn}. The natural gradient approach therefore offers a different justification for these algorithms, one that involves qualitatively different approximations. It should also be noted that research on the design of methods inspired by the natural gradient is ongoing and may lead to markedly different algorithms \cite{Card98,HoffBlei13,martens2014new}.

\paragraph{\textbf{Information Geometry}}

In order to directly formulate the gradient descent in the space of prediction functions, we must elucidate the geometry of this space. Amari's work on information geometry \cite{AmarNaga97} demonstrates this for parametric density estimation. The space $\Hcal$ of prediction functions for such a problem is a family of densities $h_w(x)$ parametrized by $w\in\Wcal$ and satisfying the normalization condition
\bequationn
   \int h_w(x) \, dx = 1\ \ \text{for all}\ \ w \in \Wcal.
\eequationn
Assuming sufficient regularity, the derivatives of such densities satisfy the identity
\bequation\label{eq:dscores}
  \forall t>0 \quad \int \frac{\partial^t h_w(x)}{\partial w^t} \, dx = \frac{\partial^t}{\partial w^t} \int h_w(x) \, dx = \frac{\partial^t 1}{\partial w^t} = 0.
\eequation

To elucidate the geometry of the space $\Hcal$, we seek to quantify how the density~$h_w$ changes when one adds a small quantity $\delta w$ to its parameter. We achieve this in a statistically meaningful way by observing the Kullback-Leibler (KL) divergence 
\bequation\label{eq:kl}
   D_{KL}(h_w\|h_{w+\delta w}) = \E_{h_w}\left[ \log\left(\frac{h_w(x)}{h_{w+\delta w}(x)}\right) \right],
\eequation
where $\E_{h_w}$ denotes the expectation with respect to the distribution $h_w$.  Note that \eqref{eq:kl} only depends on the values of the two density functions $h_w$ and $h_{w+\delta w}$ and therefore is invariant with respect to any invertible transformation of the parameter~$w$.  Approximating the divergence with a second-order Taylor expansion, one obtains
\bequalin
   D_{KL}(h_w\|h_{w+\delta w})
   &=       \E_{h_w}[\log(h_w(x)) - \log(h_{w+\delta w}(x))] \\
   &\approx -\delta w^T \E_{h_w}\left[\frac{\partial \log(h_w(x))}{\partial w}\right] - \thalf \delta w^T \E_{h_w}\left[ \frac{\partial^2 \log(h_w(x))}{\partial w^2} \right] \delta w,
\eequalin
which, after observing that \eqref{eq:dscores} implies that the first-order term is null, yields
\bequation
  \label{eq:riemann}
    D_{KL}(h_w\|h_{w+\delta w}) \approx \thalf \delta w^T G(w) \delta w.
\eequation
This is a quadratic form defined by the \emph{Fisher information} matrix
\bequation
  \label{eq:fisher}
   G(w) := -\E_{h_w}\left[ \frac{\partial^2 \log(h_w(x))}{\partial w^2} \right]
         = -\E_{h_w}\left[ \left(\frac{\partial \log(h_w(x))}{\partial w}\right)
                         \left(\frac{\partial \log(h_w(x))}{\partial w}\right)^T \right],
\eequation
where the latter equality follows again from~\eqref{eq:dscores}. 
The second form of $G(w)$ is often preferred because it makes clear that the Fisher information matrix $G(w)$ is symmetric and always positive semidefinite, though not necessarily positive definite.

The relation \eqref{eq:riemann} means that the KL divergence behaves locally like a norm associated with $G(w)$.  Therefore, every small region of $\Hcal$ looks like a small region of a Euclidean space.  However, as we traverse larger regions of $\Hcal$, we cannot ignore that the matrix $G(w)$ changes. Such a construction defines a \emph{Riemannian geometry}.\footnote{The objective of information geometry \cite{AmarNaga97} is to exploit the Riemannian structure of parametric families of density functions to gain geometrical insights on the fundamental statistical phenomena. The natural gradient algorithm is only a particular aspect of this broader goal  \cite{Amar98}.}

Suppose, for instance, that we move along a smooth path connecting two densities, call them~$h_{w_0}$ and~$h_{w_1}$.  A parametric representation of the path can be given by a differentiable function, for which we define
\bequationn
  \phi : t \in [0,1] \mapsto \phi(t) \in \Wcal\ \ \text{with}\ \ \phi(0)=w_0\ \ \text{and}\ \ \phi(1)=w_1.
\eequationn
We can compute the length of the path by viewing it as a sequence of infinitesimal segments $[\phi(t),\phi(t+dt)]$ whose length is given by \eqref{eq:riemann}, i.e., the total length is
\bequationn
   \def\dg{\frac{d\phi}{dt}(t)}
   D_\phi = \int_0^1{\sqrt{\left(\dg\right)^T G(\phi(t)) \left(\dg\right)} \, dt}.
\eequationn

An important tool for the study of Riemannian geometries is the characterization of its \emph{geodesics}, i.e., the shortest paths connecting two points.  In a Euclidean space, the shortest path between two points is always the straight line segment connecting them. In a Riemannian space, on the other hand, the shortest path between two points can be curved and does not need to be unique.  Such considerations are relevant to optimization since every iterative optimization algorithm can be viewed as attempting to follow a particular path connecting the initial point~$w_0$ to the optimum~$w_*$.  In particular, following the shortest path is attractive because it means that the algorithm reaches the optimum after making the fewest number of changes to the optimization variables, hopefully requiring the least amount of computation.


\paragraph{\textbf{Natural Gradient}}

Let us now assume that the space $\Hcal$ of prediction functions $\{h_w \::\: w\in\Wcal\}$ has a Riemannian geometry locally described by an identity of the form \eqref{eq:riemann}.  We seek an algorithm that minimizes a functional $F: h_w\in\Hcal\mapsto{F(h_w)=F(w)\in\R{}}$ and is invariant with respect to differentiable invertible transformations of the parameters represented by the vector $w$. 

Each iteration of a typical iterative optimization algorithm computes a new iterate $h_{w_{k+1}}$ on the basis of information pertaining to the current iterate $h_{w_k}$. Since we can only expect this information to be valid in a small region surrounding $h_{w_k}$, we restrict our attention to algorithms that make a step from $h_{w_k}$ to $h_{w_{k+1}}$ of some small length $\eta_k>0$.  The number of iterations needed to reach the optimum then depends directly on the length of the path followed by the algorithm, which is desired to be as short as possible. Unfortunately, it is rarely possible to exactly follow a geodesic using only local information.  We can, however, formulate the greedy strategy that
\bequation
  \label{eq:ngh}
   h_{w_{k+1}} = \mathop{\argmin}_{h\in\Hcal}\ F(h)\ \ \st\ \ D(h_{w_k}\|h) \leq \eta_k^2,
\eequation
and use \eqref{eq:riemann} to reformulate this problem in terms of the parameters:
\bequation
   \label{eq:ngw}
   w_{k+1} = \mathop{\argmin}_{w\in\Wcal}\ F(w)\ \ \st\ \ \thalf (w-w_k)^T G(w_k)\,(w-w_k)\leq\eta_k^2 \,.
\eequation
The customary derivation of the natural gradient algorithm handles the constraint in \eqref{eq:ngw} using a Lagrangian formulation with Lagrange multiplier $1/\alpha_k$.  In addition, since $\eta_k$ is assumed small, it replaces $F(w)$ in \eqref{eq:ngw} by the first-order approximation $F(w_k)+\nabla F(w_k)^T(w-w_k)$.  These two choices lead to the expression
\bequationn
   w_{k+1} = \mathop{\argmin}_{w\in\Wcal}\ \nabla F(w_k)^T(w-w_k) + \frac{1}{2\alpha_k} (w-w_k)^T G(w_k)(w-w_k),
\eequationn
the optimization of the right-hand side of which leads to the natural gradient iteration
\bequation
\label{eq:ngbatch}
   w_{k+1} = w_k - \alpha_k G^{-1}(w_k) \nabla F(w_k).
\eequation
We can also replace $F(w)$ in \eqref{eq:ngw} by a noisy first-order approximation, leading to a stochastic natural gradient iteration where $\nabla F(w_k)$ in \eqref{eq:ngbatch} is replaced by a stochastic gradient estimate.

Both batch and stochastic versions of \eqref{eq:ngbatch} resemble the quasi-Newton update rules discussed in~\S\ref{sec:stochqn}.  Instead of multiplying the gradient by the inverse of an approximation of the Hessian (which is not necessarily positive definite), it employs the positive semidefinite matrix $G(w_k)$ that expresses the local geometry of the space of prediction functions.  In principle, this matrix does not even take into account the objective function $F$.  However, as we shall now describe, one finds that these choices are all closely related in practice.

\paragraph{\textbf{Practical Natural Gradient}}

Because the discovery of the natural gradient algorithm is closely associated with information geometry, nearly all its applications involve density estimation \cite{Amar98,Card98} or conditional probability estimation \cite{ParkAmar00,HoffBlei13,CaroOlli16} using objective functions that are closely connected to the KL divergence. Natural gradient in this context is closely related to Fisher's scoring algorithm~\cite{Osborne92}. For instance, in the case of density estimation, the objective is usually  the negative log likelihood
\bequationn
   F(w) = \frac{1}{n} \sum_{i=1}^{n} -\log(h_w(x_i)) \approx \text{constant} + D_{KL}(P\|h_w),
\eequationn
where $\{x_1,\dots,x_n\}$ represent independent training samples from an unknown distribution $P$. Recalling the expression of the Fisher information matrix \eqref{eq:fisher} then clarifies its connection with the Hessian as one finds that
\bequationn
  G(w) = -\E_{h_w} \left[ \frac{\partial^2\log(h_w(x))}{\partial w^2} \right]\ \ \text{and}\ \ \nabla^2 F(w) = -\E_{P} \left[ \frac{\partial^2\log(h_w(x))}{\partial w^2} \right].
\eequationn
These two expressions do not coincide in general because the expectations involve different distributions.  However, when the natural gradient algorithm approaches the optimum, the parametric density $h_{w_k}$ ideally approaches the true distribution $P$, in which case the Fisher information matrix $G(w_k)$ approaches the Hessian matrix $\nabla^2 F(w_k)$.  This means that the natural gradient algorithm and Newton's method perform very similarly as optimality is approached.

Although it is occasionally possible to determine a convenient analytic expression~\cite{Card98,HoffBlei13}, the numerical computation of the Fisher information matrix $G(w_k)$ in large learning systems is generally very challenging.  Moreover, estimating the expectation \eqref{eq:fisher} with, say, a Monte-Carlo approach is usually prohibitive  due to the cost of sampling the current density estimate $h_{w_k}$.

Several authors \cite{ParkAmar00,CaroOlli16} suggest to use instead a subset of training examples and compute a quantity of the form
\bequationn
     \widetilde{G}(w_k) = \frac{1}{|S_k|} \sum_{i\in S_k} \left(\left.\frac{\partial \log(h_w(x_i))}{\partial w}\right|_{w_k}\right)
                         \left(\left.\frac{\partial \log(h_w(x_i))}{\partial w}\right|_{w_k}\right)^T.
\eequationn
Although such algorithms are essentially equivalent to the generalized Gauss-Newton schemes described in \S\ref{sec:gauss}, the natural gradient perspective comes with an interesting insight into the relation between the generalized Gauss-Newton matrix \eqref{eq:ggnforlogloss} and the Hessian matrix \eqref{eq:subsampledhessian}. Similar to the equality \eqref{eq:fisher}, these two matrices would be equal if the expectation was taken with respect to the model distribution $h_w$ instead of the empirical sample distribution.

\subsection{Methods that Employ Diagonal Scalings}   \label{sec:adagrad}

The methods that we have discussed so far in this section are forced to overcome the fact that when employing an iteration involving an $\R{d} \times \R{d}$ scaling matrix, one needs to ensure that the improved per-iteration progress outweighs the added per-iteration cost. We have seen that these added costs can be as little as $4md$ operations and therefore amount to a moderate multiplicative factor on the cost of each iteration.

A strategy to further reduce this multiplicative factor, while still incorporating second-order-type information, is to restrict attention to \emph{diagonal} or \emph{block-diagonal} scaling matrices.  Rather than perform a more general linear transformation through a symmetric positive definite matrix (i.e., corresponding to a scaling and rotation of the direction), the incorporation of a diagonal scaling matrix only has the effect of scaling the individual search direction components.  This can be efficiently achieved by multiplying each coefficients of the gradient vector by the corresponding diagonal term of the scaling matrix, or, when the prediction function is linear, by adaptively renormalizing the input pattern coefficients \cite{RossMineLang13}.  

\paragraph{\textbf{Computing Diagonal Curvature}}

A first family of algorithms that we consider directly computes the diagonal terms of the Hessian or Gauss-Newton matrix, then divides each coefficient of the stochastic gradient vector $g(w_k,\xi_k)$ by the corresponding diagonal term.  Since the computation overhead of this operation is very small, it becomes important to make sure that the estimation of the diagonal terms of the curvature matrix is very efficient.

For instance, in the context of deep neural networks, \cite{BeckLeCu89} describes a back-propagation algorithm to efficiently compute the diagonal terms of the squared Jacobian matrix \mbox{$J_h(w_k;\xi_k)^T J_h(w_k;\xi_k)$} that appears in the expression of the Gauss-Newton matrix \eqref{eq:gaussnewtonmatrix}.  Each iteration of the proposed algorithm picks a training example, computes the stochastic gradient $g(w_k,\xi_k)$, updates a running estimate of the diagonal coefficients of the Gauss-Newton matrix by
\bequationn
   \def\b#1#2{\big[{#1}\big]_{#2}}
   \b{G_{k}}{i} = (1-\lambda) \b{G_{k-1}}{i} + \lambda \b{J_h(w_k;\xi_k)^T J_h(w_k;\xi_k)}{ii}\ \ \text{for some}\ \ 0<\lambda<1,
\eequationn
then performs the scaled stochastic weight update
\bequationn
   \def\b#1#2{\big[{#1}\big]_{#2}}
   \b{w_{k+1}}{i} = \b{w_k}{i} - \(\frac{\alpha}{\b{G_{k}}{i} + \mu}\)\b{g(w_k,\xi_k)}{i}.
\eequationn
The small regularization constant $\mu>0$ is introduced to handle situations where the Gauss-Newton matrix is singular or nearly singular. Since the computation of the diagonal of the squared Jacobian has a cost that is comparable to the cost of the computation of the stochastic gradient, each iteration of this algorithm is roughly twice as expensive as a first-order stochastic gradient iteration.  The experience described in \cite{BeckLeCu89} shows that improvement in per-iteration progress can be sufficient to modestly outperform a well-tuned SG algorithm.

After describing this algorithm in later work \cite[\S9.1]{LeCuBottOrrMuel98}, the authors make two comments that illustrate well how this algorithm was used in practice. They first observe that the curvature only changes very slowly in the particular type of neural network that was considered.  Due to this observation, a natural idea is to further reduce the computational overhead of the method by estimating the ratios $\alpha/([G_{k+1}]_{i} + \mu)$ only once every few epochs, for instance using a small subset of examples as in \eqref{eq:gaussnewtonmatrix}. The authors also mention that, as a rule of thumb, this diagonal scheme typically improves the convergence speed by only a factor of three relative to SG.  Therefore, it might be more enlightening to view such an algorithm as a scheme to periodically retune a first-order SG approach rather than as a complete second-order method.

\paragraph{\textbf{Estimating Diagonal Curvature}}

Instead of explicitly computing the diagonal terms of the curvature matrix, one can follow the template of \S\ref{sec:stochqn} and directly estimate the diagonal $[H_k]_i$ of the inverse Hessian using displacement pairs $\{(s_k,v_k)\}$ as defined in \eqref{cambert}.  
For instance, \cite{BordBottGall09} proposes to compute the scaling terms $[H_k]_i$ with the running average
\bequationn
   \def\b#1#2{\big[{#1}\big]_{#2}}
   \b{H_{k+1}}{i} = (1-\lambda)\b{H_{k}}{i} + \lambda\mathrm{Proj}\left( \frac{\b{s_k}{i}}{\b{v_k}{i}} \right),
\eequationn
where $\mathrm{Proj(\cdot)}$ represents a projection onto a predefined positive interval. It was later found that a direct application of \eqref{cambert} after a parameter update introduces a correlated noise that ruins the curvature estimate \cite[\S3]{BordBottGall10}.  Moreover, correcting this problem made the algorithm perform substantially worse because the chaotic behavior of the rescaling factors $[H_k]_i$ makes the choice of the stepsize $\alpha$ very difficult.

These problems can be addressed with a combination of two ideas \cite[\S5]{BordBottGall10}. The first idea consists of returning to estimating the diagonal of the Hessian instead of the diagonal of this inverse, which amounts to working with the ratio $[v_k]_i/[s_k]_i$ instead of $[s_k]_i/[v_k]_i$.  The second idea ensures that the effective stepsizes are monotonically decreasing by replacing the running average by the sum
\bequationn
   \def\b#1#2{\big[{#1}\big]_{#2}}
   \b{G_{k+1}}{i} = \b{G_{k}}{i} + \mathrm{Proj}\left( \frac{\b{v_k}{i}}{\b{s_k}{i}} \right).
\eequationn
This effectively constructs a separate diminishing stepsize sequence $\alpha/[G_k]_i$ for each coefficient of the parameter vector. Keeping the curvature estimates in a fixed positive interval ensures that the effective stepsizes decrease at the rate $\Ocal(1/k)$ as prescribed by Theorem~\ref{th.sg_sc}, while taking the local curvature into account. This combination was shown to perform very well when the input pattern coefficients have very different variances \cite{BordBottGall10}, something that often happens, e.g., in text classification problems.

\paragraph{\textbf{Diagonal Rescaling without Curvature}}

The algorithms described above often require some form of regularization to handle situations where the Hessian matrix is (nearly) singular.  To illustrate why this is needed, consider, e.g., optimization of the convex objective function
\bequationn
  F(w_1,w_2) = \thalf w_1^2 + \log(e^{w_2}+e^{-w_2}),
\eequationn
for which ones finds
\bequationn
  \nabla F(w_1,w_2) = \left[\begin{array}{c} w_1 \\ \mathrm{tanh}(w_2) \end{array}\right]\ \ \text{and}\ \ \nabla^2 F(w_1,w_2) =  \left[\begin{array}{cc} 1 & 0 \\ 0 & 1/\mathrm{cosh}^2(w_2) \end{array}\right].
\eequationn
Performing a first-order gradient method update from a starting point of $(3,3)$ yields the negative gradient step $-\nabla F\approx[-3,-1]$, which unfortunately does not point towards the optimum, namely the origin.  Moreover, rescaling the step with the inverse Hessian actually gives a worse update direction $-(\nabla^2F)^{-1}\nabla F\approx[-3,-101]$ whose large second component requires a small stepsize to keep the step well contained.  
Batch second-order optimization algorithms can avoid having to guess a good stepsize by using, e.g., line search techniques.  Stochastic second-order algorithms, on the other hand, cannot rely on such procedures as easily.

This problem is of great concern in situations where the objective function is nonconvex.  For instance, optimization algorithms for deep neural networks must navigate around saddle points and handle near-singular curvature matrices.  It is therefore tempting to consider diagonal rescaling techniques that simply ensure equal progress along each axis, rather than attempt to approximate curvature very accurately.

For instance, \textsc{RMSprop}~\cite{TielHint12} estimates the average magnitude of each element of the stochastic gradient vector $g(w_k,\xi_k)$ by maintaining the running averages
\bequationn
   \def\b#1#2{\big[{#1}\big]_{#2}}
   \b{R_k}{i} = (1-\lambda)\b{R_{k-1}}{i} + \lambda \b{g(w_k,\xi_k)}{i}^2.
\eequationn
The rescaling operation then consists in dividing each component of $g(w_k,\xi_k)$ by the square root of the corresponding running average, ensuring that the expected second moment of each coefficient of the rescaled gradient is close to the unity:
\bequationn
   \def\b#1#2{\big[{#1}\big]_{#2}}
   \b{w_{k+1}}{i} = \b{w_k}{i} - \frac{\alpha}{\sqrt{\b{R_k}{i} + \mu}} \,\b{g(w_k,\xi_k)}{i}.
\eequationn
This surprisingly simple approach has shown to be very effective for the optimization of deep neural networks. Various improvement have been proposed \cite{Zeiler12,KingBa14} on an empirical basis. The theorerical explanation of this performance on nonconvex problems is still the object of active research \cite{DaupVrie15}.

The popular \textsc{Adagrad} algorithm \cite{DuchHazaSing11} can be viewed as a member of this family that replaces the running average by a sum:
\bequationn
   \def\b#1#2{\big[{#1}\big]_{#2}}
   \b{R_k}{i} = \b{R_{k-1}}{i} + \b{g(w_k,\xi_k)}{i}^2.
\eequationn
In this manner, the approach constructs a sequence of diminishing effective stepsizes $\alpha/\sqrt{[R_k]_i+\mu}$ for each coefficient of the parameter vector. This algorithm was initially proposed and analyzed for the optimization of (not necessarily strongly) convex functions for which SG theory suggests diminishing stepsizes that scale with $\Ocal(1/\sqrt{k})$.  \textsc{Adagrad} is also known to perform well on deep learning networks, but one often finds that its stepsizes decrease too aggressively early in the optimization~\cite{Zeiler12}.

\paragraph{\textbf{Structural Methods}}

The performance of deep neural network training can of course be improved by employing better optimization algorithms.  However, it can also be improved by changing the structure of the network in a manner that facilitates the optimization~\cite{IoffSzeg15,HeResidualNets16}. 
We now describe one of these techniques, batch normalization \cite{IoffSzeg15}, and discuss its relation to diagonal second-order methods.

Consider a particular fully connected layer in a deep neural network of the form discussed in \S\ref{sec.deep_neural_nets}.  Using the notation of equation \eqref{eqn.sec2.flayer}, the vector $x_i^{(j)}$ represents the input values of layer $j$ when the network is processing the $i$-th training example. Omitting the layer index for simplicity, let $\xhat_i=(x_i^{(j)},1)$ denote the input vector augmented with an additional unit coefficient and let $\what_r=(W_{r1},\dots,W_{rd_{j-1}},b_r)$ be the $r$th row of the matrix $W_j$ augmented with the $r$th coefficient of the bias vector~$b_j$. The layer outputs are then obtained by applying the activation function to the quantities $s_r=\what_r^T \xhat_i$ for $r\in\{1,\dots, d_j\}$.
Assuming for simplicity that all other parameters of the network are kept fixed, we can write
\bequationn
   F(\what_1,\dots,\what_{d_j}) = \frac{1}{n} \sum_{i=1}^{n} \ell( h( \what_1^T \xhat_i, \what_2^T \xhat_i, \dots, \what_{d_j}^T \xhat_i), y_i),
\eequationn
where $h(s_1,\dots,s_{d_j})$ encapsulates all subsequent layers in the network. The diagonal block of the Gauss-Newton matrix \eqref{eq:ggnmatrix} corresponding to the parameters $\what_r$ then has the form
\bequation
\label{eq.sm.g}
     G_{[r]}  = \frac{1}{|\Scal|} \sum_{i\in \Scal} \left[ 
          \left(\frac{d h}{d s_r}\right)^{\!T} 
          \left(\frac{\partial^2\ell}{\partial h^2}\right)
          \left(\frac{d h}{d s_r}\right)  \right]  \xhat_i \xhat_i^T,
\eequation
which can be viewed as a weighted second moment matrix of the augmented input vectors $\{\xhat_i\}_{i\in\Scal}$. In particular, this matrix is perfectly conditioned if the weighted distribution of the layer inputs is white, i.e., they have zero mean and a unit covariance matrix.  This could be achieved by first preprocessing the inputs by an affine transform that whitens their weighted distribution.

Two simplifications can drastically reduce the computational cost of this operation. First, we can ignore the bracketed coefficient in \eqref{eq.sm.g} and assume that we can use the same whitening transformation for all outputs $r\in\{1,\dots, d_j\}$. Second, we can ignore the input cross-correlations and simply ensure that each input variable has zero mean and unit variance by replacing the input vector coefficients $\xhat_i[t]$ for each $t\in\{1,\dots,d_{j-1}\}$ by the linearly transformed values $\alpha_t \xhat_i[t] + \beta_t$. Despite these simplifications, this normalization operation is very likely to improve the second order properties of the objective function. An important detail here is the computation of the normalization constants $\alpha_t$ and $\beta_t$. Estimating the mean and the standard deviation of each input with a simple running average works well if one expects these quantities to change very slowly. This is unfortunately not true in recent neural networks.\footnote{This used to be true in the 1990s because neural networks were using bounded activation functions such as the sigmoid $s(x)=1/(1+e^{-x})$. However, many recent results were achieved using the ReLU activation function $s(x)=\max\{0,x\}$ which is unbounded and homogeneous. The statistics of the intermediate variables in such network can change extremely quickly during the first phases of the optimization process \cite{LafoVasiBott16}.}

Batch normalization \cite{IoffSzeg15} defines a special kind of neural network layer that performs this normalization using statistics collected with the current mini-batch of examples. The back-propagation algorithm that computes the gradients must of course be adjusted to account for the on-the-fly computation of the normalization coefficients. Assuming that one uses sufficiently large mini-batches, computing the statistics in this manner ensures that the normalization constants are very fresh. This comes at the price of making the output of the neural network on a particular training pattern dependent on the other patterns in the mini-batch. Since these other examples are \emph{a priori} random, this amounts to generating additional noise in the stochastic gradient optimization. Although the variance of this noise is poorly controlled, inserting batch normalization layers in various points of a deep neural network is extremely effective and is now standard practice. Whether one can achieve the same improvement with more controlled techniques remains to be seen.

\section{Other Popular Methods}\label{sec.other}
\setcounter{equation}{0}
\setcounter{theorem}{0}
\setcounter{algorithm}{0}
\setcounter{figure}{0}
\setcounter{table}{0}

Some optimization methods for machine learning are not well characterized as being within the two-dimensional schematic introduced in \S\ref{sec.beyond_sg} (see Figure~\ref{view1} on page~\pageref{view1}), yet represent fundamentally unique approaches that offer theoretical and/or practical advantages.  The purpose of this section is to discuss a few such ideas, namely, gradient methods with momentum, accelerated gradient methods, and coordinate descent methods.  For ease of exposition, we introduce these techniques under the assumption that one is minimizing a continuously differentiable (not necessarily convex) function $F : \R{n} \to \R{}$ and that full gradients can be computed in each iteration.  Then, after each technique is introduced, we discuss how they may be applied in stochastic settings.

\subsection{Gradient Methods with Momentum}\label{sec.momentum}

Gradient methods with momentum are procedures in which each step is chosen as a combination of the steepest descent direction and the most recent iterate displacement.  Specifically, with an initial point $w_1$, scalar sequences $\{\alpha_k\}$ and $\{\beta_k\}$ that are either predetermined or set dynamically, and $w_0 := w_1$, these methods are characterized by the iteration
\bequation\label{moment}
  w_{k+1} \gets w_k - \alpha_k \nabla F(w_k) + \beta_k (w_k - w_{k-1}).
\eequation
Here, the latter is referred to as the momentum term, which, recursively, maintains the algorithm's movement along previous search directions.

The iteration \eqref{moment} can be motivated in various ways; e.g., it is named after the fact that it represents a discretization of a certain second-order ordinary differential equation with friction.  Of course, when $\beta_k = 0$ for all $k \in \N{}$, it reduces to the steepest descent method.  When $\alpha_k = \alpha$ and $\beta_k = \beta$ for some constants $\alpha>0$ and $\beta>0$ for all $k \in \N{}$, it is referred to as the heavy ball method \cite{Poly64}, which is known to yield a superior rate of convergence as compared to steepest descent with a fixed stepsize for certain functions of interest.  For example, when $F$ is a strictly convex quadratic with minimum and maximum eigenvalues given by $c > 0$ and $L \geq c$, respectively, steepest descent and the heavy ball method each yield a linear rate of convergence (in terms of the distance to the solution converging to zero) with contraction constants respectively given by
\bequation\label{attainable}
  \frac{\kappa - 1}{\kappa + 1}\ \ \text{and}\ \ \frac{\sqrt\kappa - 1}{\sqrt\kappa + 1}\ \ \text{where}\ \ \kappa := \frac{L}{c} \geq 1.
\eequation
Choosing $(\alpha,\beta)$ to achieve these rates requires knowledge of $(c,L)$, which might be unavailable.  Still, even without this knowledge, the heavy ball method often outperforms steepest descent.

Additional connections with \eqref{moment} can be made when $F$ is a strictly convex quadratic.  In particular, if $(\alpha_k,\beta_k)$ is chosen optimally for all $k \in \N{}$, in the sense that the pair is chosen to solve
\bequation\label{eq.CG_momentum}
  \min_{(\alpha,\beta)}\ F(w_k-\alpha\nabla F(w_k)+\beta(w_k-w_{k-1})),
\eequation
then \eqref{moment} is exactly the linear conjugate gradient (CG) algorithm.  While the heavy ball method is a stationary iteration (in the sense that the pair $(\alpha,\beta)$ is fixed), the CG algorithm is nonstationary and its convergence behavior is relatively more complex; in particular, the step-by-step behavior of CG depends on the eigenvalue distribution of the Hessian of $F$ \cite{GoluVanL12}.  That said, in contrast to the heavy ball method, CG has a finite convergence guarantee.  This, along with the fact that problems with favorable eigenvalue distributions are quite prevalent, has lead to the great popularity of CG in a variety of situations.  
More generally, nonlinear CG methods, which also follow the procedure in \eqref{moment}, can be viewed as techniques that approximate the optimal values defined by~\eqref{eq.CG_momentum} when $F$ is not quadratic.

An alternative view of the heavy ball method is obtained by expanding \eqref{moment}:
\bequationn
  w_{k+1} \gets w_k - \alpha \sum_{j=1}^k \beta^{k-j} \nabla F(w_k);
\eequationn
thus, each step can be viewed as an exponenetially decaying average of past gradients.  By writing the iteration this way, one can see that the steps tend to accumulate contributions in directions of persistent descent, while directions that oscillate tend to get cancelled, or at least remain small.

This latter interpretation provides some intuitive explanation as to why a stochastic heavy ball method, and stochastic gradient methods with momentum in general, might be successful in various settings.  In particular, their practical performance have made them popular in the community working on training deep neural networks \cite{SutsMartDahlHint13}.  Replacing the true gradient with a stochastic gradient in \eqref{moment}, one obtains an iteration that, over the long run, tends to continue moving in directions that the stochastic gradients suggest are ones of improvement, whereas movement is limited along directions along which contributions of many stochastic gradients cancel each other out.  Theoretical guarantees about the inclusion of momentum in stochastic settings are elusive, and although practical gains have been reported \cite{LeenOrr93,SutsMartDahlHint13}, more experimentation is needed.

\subsection{Accelerated Gradient Methods}\label{sec.accelerated}

A method with an iteration similar to \eqref{moment}, but with its own unique properties, is the accelerated gradient method proposed by Nesterov \cite{Nest83}.  Written as a two-step procedure, it involves the updates
\bequation\label{accelerated_two}
  \baligned
    \wtilde_k &\gets w_k + \beta_k(w_k - w_{k-1}) \\ \text{and}\ \ 
    w_{k+1} &\gets \wtilde_k - \alpha_k \nabla F(\wtilde_k),
  \ealigned
\eequation
which leads to the condensed form
\bequation\label{accelerated}
  w_{k+1} \gets w_k - \alpha_k \nabla F(w_k + \beta_k(w_k - w_{k-1})) + \beta_k(w_k - w_{k-1}).
\eequation
In this manner, it is easy to compare the approach with \eqref{moment}.  In particular, one can describe their difference as being a reversal in the order of computation.  In \eqref{moment}, one can imagine taking the steepest descent step and then applying the momentum term, whereas \eqref{accelerated} results when one follows the momentum term first, then applies a steepest descent step (with the gradient evaluated at $\wtilde_k$, not at $w_k$).

While this difference may appear to be minor, it is well known that \eqref{accelerated} with appropriately chosen $\alpha_k = \alpha > 0$ for all $k \in \N{}$ and $\{\beta_k\} \nearrow 1$ leads to an \emph{optimal} iteration complexity when $F$ is convex and continuously differentiable with a Lipschitz continuous gradient.  Specifically, while in such cases a steepest descent method converges with a distance to the optimal value decaying with a rate $\Ocal(\tfrac1k)$, the iteration \eqref{accelerated} converges with a rate $\Ocal(\tfrac1{k^2})$, which is provably the best rate that can be achieved by a gradient method.  Unfortunately, no intuitive explanation as to how Nesterov's method achieves this optimal rate has been widely accepted.  Still, one cannot deny the analysis and the practical gains that the technique has offered.

Acceleration ideas have been applied in a variety of other contexts as well, including for the minimization of nonsmooth convex functions; see \cite{LinMairHarc15}.  For now, we merely mention that when applied in stochastic settings---with stochastic gradients employed in place of full gradients---one can only hope that acceleration might improve the constants in the convergence rate offered in Theorem~\ref{th.sg_sc}; i.e., the rate itself cannot be improved \cite{AgarBartRaviWain12}.

\subsection{Coordinate Descent Methods}

Coordinate descent (CD) methods are among the oldest in the optimization literature.  As their name suggests, they operate by taking steps along coordinate directions: one attempts to minimize the objective with respect to a single variable while all others are kept fixed, then other variables are updated similarly in an iterative manner.  Such a simple idea is easy to implement, so it is not surprising that CD methods have a long history in many settings.  Their limitations have been documented and well understood for many years (more on these below), but one can argue that their advantages were not fully recognized until recent work in machine learning and statistics demonstrated their ability to take advantage of certain structures commonly arising in practice.

The CD method for minimizing $F : \R{d} \to \R{}$ is given by the iteration
\bequation\label{eq.coordinate_descent}
  w_{k+1} \gets w_k - \alpha_k \nabla_{i_k} F(w_k) e_{i_k},\ \ \text{where}\ \ \nabla_{i_k} F(w_k) := \frac{\partial F}{\partial w^{i_k}}(w_k),
\eequation
$w^{i_k}$ represents the $i_k$-th element of the parameter vector, and $e_{i_k}$ represents the $i_k$-th coordinate vector for some $i_k \in \{1,\dots,d\}$.  In other words, the solution estimates $w_{k+1}$ and $w_k$ differ only in their $i_k$-th element as a result of a move in the $i_k$-th coordinate from $w_k$. 

Specific versions of the CD method are defined by the manner in which the sequences $\{\alpha_k\}$ and $\{i_k\}$ are chosen.  In some applications, it is possible to choose $\alpha_k$ as the global minimizer of $F$ from $w_k$ along the $i_k$-th coordinate direction.  An important example of this, which has contributed to the recent revival of CD methods, occurs when the objective function has the form $F(w) = q(w) + \|w\|_1$ when $q$ is a convex quadratic.  Here, the exact minimization along each coordinate is not only possible, but desirable as it promotes the generation of sparse iterates; see also \S\ref{sec.nonsmooth}.  More often, an exact one-dimensional minimization of $F$ is not practical, in which case one is typically satisfied with $\alpha_k$ yielding a sufficient reduction in $F$ from $w_k$.  For example, so-called second-order CD methods compute $\alpha_k$  as the minimizer of a quadratic model of $F$ along the $i_k$-th coordinate direction.

Concerning the choice of $i_k$, one could select it in each iteration in at least three different ways: by cycling through $\{1,\dots,d\}$; by cycling through a random reordering of these indices (with the indices reordered after each set of $d$ steps); or simply by choosing an index randomly with replacement in each iteration.  Randomized CD algorithms (represented by the latter two strategies) have superior theoretical properties than the cyclic method (represented by the first strategy) as they are less likely to choose an unfortunate series of coordinates; more on this below.  However, it remains an open question whether such randomized algorithms are more effective in typical applications.

We mention in passing that it is also natural to consider a \emph{block-coordinate descent} method in which a handful of elements are chosen in each iteration.  This is is particularly effective when the objective function is (partially) block separable, which occurs in matrix factorization problems and least squares and logistic regression when each sample only depends on a few features.  Clearly, in such settings, there are great advantages of a block-coordinate descent approach.  However, since their basic properties are similar to the case of using a single index in each iteration, we focus on the iteration \eqref{eq.coordinate_descent}.

\paragraph{\textbf{Convergence Properties}}

Contrary to what intuition might suggest, a CD method is not guaranteed to converge when applied to minimize any given continuously differentiable function.  Powell \cite{Powe73} gives an example of a \emph{nonconvex} continuously differentiable function of three variables for which a cyclic CD method with $\alpha_k$ chosen by exact one-dimensional minimization cycles without converging to a solution, i.e., at any limit point the gradient of $F$  is nonzero.  Although one can argue that failures of this type are unlikely to occur in practice, particularly for a randomized CD method, they show the weakness of the myopic strategy in a CD method that considers only one variable at a time.  This is in contrast with the full gradient method, which guarantees convergence to stationarity even when the objective is nonconvex.

On the other hand, if the objective $F$ is strongly convex, the CD method will not fail and one can establish a linear rate of convergence. The analysis is very simple when using a constant stepsize and we present one such result to provide some insights into the tradeoffs that arise with a CD approach.  Let us assume that $\nabla F$ is coordinate-wise Lipschitz continuous in the sense that, for all $w \in \R{d}$, $i \in \{1,\dots,d\}$, and $\Delta w^i \in \R{}$, there exists a constant $L_i > 0$ such that
\bequation
  |\nabla_i F(w + \Delta w^i e_i) - \nabla_i F(w)| \leq L_i|\Delta w^i|.
\eequation
We then define the maximum coordinate-wise Lipschitz constant as
\bequationn
  \Lhat := \max_{i \in \{1,\dots,d\}} L_i.
\eequationn
Loosely speaking, $\Lhat$ is a bound on the curvature of the function along all coordinates.

\btheorem\label{th.cd}
  Suppose that the objective function $F : \R{d} \to \R{}$ is continuously differentiable, strongly convex with constant $c > 0$, and has a gradient that is coordinate-wise Lipschitz continuous with constants $\{L_1,\dots,L_d\}$.  In addition, suppose that $\alpha_k = 1/\Lhat$ and $i_k$ is chosen independently and uniformly from $\{1,\dots,d\}$ for all $k \in \N{}$.  Then, for all $k \in \N{}$, the iteration \eqref{eq.coordinate_descent} yields
  \bequation\label{eq.cd_converge}
    \E[F(w_{k+1})]- F_* \leq \(1 - \frac{c}{d\Lhat}\)^k(F(w_1) - F_*).
  \eequation
\etheorem
\bproof
  As Assumption~\ref{ass.Lipschitz} leads to \eqref{eq.lipschitz_bound}, coordinate-wise Lipschitz continuity of $\nabla F$ yields
  \bequationn
    F(w_{k+1}) \leq F(w_k) + \nabla_{i_k} F(w_k)(w_{k+1}^{i_k} - w_k^{i_k}) + \thalf \Lhat(w_{k+1}^{i_k} - w_k^{i_k})^2.
  \eequationn
  Thus, with the stepsize chosen as $\alpha_k = 1/\Lhat$, it follows that
  \bequationn
    F(w_{k+1}) - F(w_k) \leq -\tfrac{1}{\Lhat} \nabla_{i_k} F(w_k)^2 + \tfrac{1}{2\Lhat} \nabla_{i_k} F(w_k)^2 = -\tfrac{1}{2\Lhat} \nabla_{i_k} F(w_k)^2.
  \eequationn
  Taking expectations with respect to the distribution of $i_k$, one obtains
  \bequalin
    \E_{i_k}[F(w_{k+1})] - F(w_k)
      &\leq -\tfrac{1}{2\Lhat} \E_{i_k}[\nabla_{i_k} F(w_k)^2] \\
      &=    -\tfrac{1}{2\Lhat} \(\tfrac{1}{d} \sum_{i=1}^d \nabla_i F(w_k)^2\) \\
      &=    -\tfrac{1}{2\Lhat d} \|\nabla F(w_k)\|_2^2.
  \eequalin
  Subtracting $F_*$, taking total expectations, recalling \eqref{eq.sc2}, and applying the above inequality repeatedly over the first $k \in \N{}$ iterations yields \eqref{eq.cd_converge}, as desired.
\eproof

A brief overview of the convergence properties of other CD methods under the assumptions of Theorem~\ref{th.cd} and in other settings is also worthwhile.  First, it is interesting to compare the result of Theorem~\ref{th.cd} with a result obtained using the \emph{deterministic} Gauss-Southwell rule, in which $i_k$ is chosen in each iteration according to the largest (in absolute value) component of the gradient.  Using this approach, one obtains a similar result in which $c$ is replaced by $\chat$, the strong convexity parameter as measured by the $\ell_1$-norm \cite{NutiSchmLaraFrieKoep15}.  Since $\tfrac{c}{n} \leq \chat \leq c$, the Gauss-Southwell rule can be up to $n$ times faster than a randomized strategy, but in the worst case it is no better (and yet more expensive due to the need to compute the full gradient vector in each iteration).  Alternative methods have also been proposed in which, in each iteration, the index $i_k$ is chosen randomly with probabilities proportional to $L_i$  or according to the largest ratio $|\nabla_i F(w_k)|/\sqrt{L_i}$ \cite{leventhal2010randomized,nesterov2012efficiency}. These strategies also lead to linear convergence rates with constants that are better in some cases.

\paragraph{\textbf{Favorable Problem Structures}}

Theorem~\ref{th.cd} shows that a simple randomized CD method is linearly convergent with constant dependent on the parameter dimension $d$.  At first glance, this appears to imply that such a method is less efficient than a standard full gradient method.  However, in situations in which $d$ coordinate updates can be performed at cost similar to the evaluation of one full gradient, the method is competitive with a full gradient method both theoretically and in practice.  Classes of problems in this category include those in which the objective function is
\bequation\label{eq.cd_F}
  F(w) = \frac{1}{n} \sum_{j=1}^n \Ftilde_j(x_j^Tw) + \sum_{i=1}^d \Fhat_i(w^i),
\eequation
where, for all $j \in \{1,\dots,n\}$, the function $\Ftilde_j$ is continuously differentiable and dependent on the \emph{sparse} data vector $x_j$, and, for all $i \in \{1,\dots,d\}$, the function $\Fhat_i$ is a (nonsmooth) regularization function.  Such a form arises in least squares and logistic regression; see also \S\ref{sec.nonsmooth}.

For example, consider an objective function of the form
\bequationn
  f(w) = \half\|Xw - y\|_2^2 + \sum_{i=1}^d \Fhat_i(w^i)\ \ \text{with}\ \ X = \bbmatrix x_1 & \cdots & x_n \ebmatrix,
\eequationn
which might be the original function of interest or might represent a model of \eqref{eq.cd_F} in which the first term is approximated by a convex quadratic model.  In this setting,
\bequationn
  \nabla_{i_k} f(w_{k+1}) = x_{i_k}^Tr_{k+1} + \Fhat_{i_k}'(w_{k+1}^{i_k})\ \ \text{with}\ \ r_{k+1} := Aw_{k+1} - b,
\eequationn
where, with $w_{k+1} = w_k + \beta_k e_{i_k}$, one may observe that $r_{k+1} = r_k + \beta_k x_{i_k}$.  That is, since the residuals $\{r_k\}$ can be updated with cost proportional to the number of nonzeros in $x_{i_k}$, call it $\text{nnz}(x_{i_k})$, the overall cost of computing the search direction in iteration $k+1$ is also $\Ocal(\text{nnz}(x_{i_k}))$.  On the other hand, an evaluation of the entire gradient requires a cost of $\Ocal(\sum_{i=1}^n \text{nnz}(x_i))$.
 
Overall, there exist various types of objective functions for which minimization by a CD method (with exact gradient computations) can be effective.  These include objectives that are (partially) block separable (which arise in dictionary learning and non-negative matrix factorization problems), have structures that allow for the efficient computation of individual gradient components, or are diagonally dominant in the sense that each step along a coordinate direction yields a reduction in the objective proportional to that which would have been obtained by a step along the steepest descent direction.  Additional settings in which CD methods are effective are online problems where gradient information with respect to a group of variables becomes available in time, in which case it is natural to update these variables as soon as information is received.

\paragraph{\textbf{Stochastic Dual Coordinate Ascent}}

What about \emph{stochastic} CD methods?  As an initial thought, one might consider the replacement of $\nabla_{i_k} F(w_k)$ in \eqref{eq.coordinate_descent} with a stochastic approximation, but this is not typical since one can usually as easily compute a $d$-dimensional stochastic gradient to apply an SG method.  However, an interesting setting for the application of stochastic CD methods arises when one considers approaches to minimize a convex objective function of the form \eqref{eq.cd_F} by maximizing its \emph{dual}.  In particular, defining the convex conjugate of $\Ftilde_j$ as $\Ftilde_j^\star(u) := \max_w (w^Tu - \Ftilde_j(w))$, the Fenchel-Rockafellar dual of \eqref{eq.cd_F} when $\Fhat_i(\cdot) = \tfrac\lambda2 (\cdot)^2$ for all $i \in \{1,\dots,d\}$ is given by
\bequationn
  F_{dual}(v) = \frac{1}{n} \sum_{j=1}^n -\Ftilde_j^\star(-v_j) - \frac{\lambda}{2} \left\| \frac{1}{\lambda n} \sum_{j=1}^n v_jx_j \right\|_2^2.
\eequationn
The stochastic dual coordinate ascent (SDCA) method \cite{ShalZhan13} applied to a function of this form has an iteration similar to \eqref{eq.coordinate_descent}, except that negative gradient steps are replaced by gradient steps due to the fact that one aims to maximize the dual.  At the conclusion of a run of the algorithm, the corresponding \emph{primal} solution can be obtained as $w \gets \frac{1}{\lambda n} \sum_{j=1}^n v_jx_j$.  The per-iteration cost of this approach is on par with that of a stochastic gradient method.

\paragraph{\textbf{Parallel CD Methods}}

We close this section by noting that CD methods are also attractive when one considers the exploitation of parallel computation.  For example, consider a multicore system in which the parameter vector~$w$ is stored in shared memory.  Each core can then execute a CD iteration independently and in an asynchronous manner, where if~$d$ is large compared to the number of cores, then it is unlikely that two cores are attempting to update the same variable at the same time.  Since, during the time it takes a core to perform an update, the parameter vector~$w$ has likely changed (due to updates produced by other cores), each update is being made based on slightly stale information.  However, convergence of the method can be proved, and improves when one can bound the degree of staleness of each update.  For further information and insight into these ideas, we refer the reader to \cite{BertTsit89,LiuWrigReBittSrid15}.

\section{Methods for Regularized Models} \label{sec.nonsmooth}
\setcounter{equation}{0}
\setcounter{theorem}{0}
\setcounter{algorithm}{0}
\setcounter{figure}{0}
\setcounter{table}{0}

Our discussion of structural risk minimization (see \S\ref{sec.SRM}) highlighted the key role played by regularization functions in the formulation of optimization problems for machine learning.  The optimization methods that we presented and analyzed in the subsequent sections (\S\ref{sec.overview}--\S\ref{sec.other}) are all applicable when the objective function involves a smooth regularizer, such as the squared $\ell_2$-norm.  In this section, we expand our investigation by considering optimization methods that handle the regularization as a distinct entity, in particular when that function is \emph{nonsmooth}.  One such regularizer that deserves special attention is the~$\ell_1$-norm, which induces sparsity in the optimal solution vector.  For machine learning, sparsity can be beneficial since it leads to simpler models, and hence can be seen as a form of \emph{feature selection}, i.e., for biasing the optimization toward solutions where only a few elements of the parameter vector are nonzero.

Broadly, this section focuses on the nonsmooth optimization problem
\bequation\label{nsprob}
  \min_{w \in \R{d}}\ \Phi(w) := F(w) + \lambda \Omega(w),
\eequation
where $F : \R{d} \to \R{}$ includes the composition of a loss and prediction function, $\lambda > 0$ is a regularization parameter, and $\Omega : \R{d} \to \R{}$ is a convex, nonsmooth regularization function.  Specifically, we pay special attention to methods for solving the problem
\bequation\label{l1prob}
  \min_{w\in\R{d}}\ \phi(w) := F(w) + \lambda\| w\|_1.
\eequation
As discussed in \S\ref{sec.case_studies}, it is often necessary to solve a series of such problems over a sequence of values for the parameter $\lambda$.  For further details in terms of problem~\eqref{l1prob}, we refer the reader to \cite{sra2011optimization,bach2012optimization} for examples in a variety of applications.  However, in our presentation of optimization methods, we assume that $\lambda$ has been prescribed and is fixed.  We remark in passing that \eqref{l1prob} has as a special case the well-known LASSO problem \cite{tibshirani1996regression} when $F(w) = \|Aw - b\|_2^2$ for some $A \in \R{n \times d}$ and $b \in \R{n}$.

Although nondifferentiable, the regularized $\ell_1$ problem \eqref{l1prob} has a structure that can be exploited in the design of algorithms.    The algorithms that have been proposed can be grouped into classes of first- or second-order methods, and distinguished as those that either minimize the nonsmooth objective directly, as in a \emph{proximal gradient} method, or by approximately minimizing a sequence of more complicated models, such as in a \emph{proximal Newton} method.

There exist other \emph{sparsity-inducing} regularizers besides the $\ell_1$-norm, including group-sparsity-inducing regularizers that combine $\ell_1$- and $\ell_2$-norms taken with respect to groups of variables \cite{sra2011optimization,bach2012optimization}, as well as the nuclear norm for optimization over spaces of matrices \cite{candes2009exact}.  While we do not discuss other such regularizers in detail, our presentation of methods for $\ell_1$-norm regularized problems represents how methods for alternative regularizers can be designed and characterized.

As in the previous sections, we introduce the algorithms in this section under the assumption that $F$ is continuously differentiable and full, batch gradients can be computed for it in each iteration, commenting on stochastic method variants once motivating ideas have been described.

\subsection{First-Order Methods for Generic Convex Regularizers}

The fundamental algorithm in unconstrained smooth optimization is the gradient method.  For solving problem~\eqref{nsprob}, the \emph{proximal gradient} method represents a similar fundamental approach.  Given an iterate $w_k$, a generic proximal gradient iteration, with $\alpha_k > 0$, is given by
\bequation\label{proxgrad}
  w_{k+1} \gets \arg\min_{w\in\R{d}} \(F(w_k) + \nabla F(w_k)^T(w - w_k) + \frac{1}{2\alpha_k}\|w - w_k\|_2^2 + \lambda\Omega(w)\).
\eequation
The term \emph{proximal} refers to the presence of the third term in the minimization problem on the right-hand side, which encourages the new iterate to be close to $w_k$.  Notice that if the regularization (i.e., last) term were not present, then \eqref{proxgrad} exactly recovers the gradient method update $w_{k+1} \gets w_k - \alpha_k\nabla F(w_k)$; hence, as previously, we refer to $\alpha_k$ as the stepsize parameter.  On the other hand, if the regularization term is present, then, similar to the gradient method, each new iterate is found by minimizing a model formed by a first-order Taylor series expansion of the objective function plus a simple scaled quadratic.  Overall, the only thing that distinguishes a proximal gradient method from the gradient method is the regularization term, which is left untouched and included explicitly in each step computation.

To show how an analysis similar to those seen in previous sections can be used to analyze \eqref{proxgrad}, we prove the following theorem.  In it, we show that if $F$ is strongly convex and its gradient function is Lipschitz continuous, then the iteration yields a global linear rate of convergence to the optimal objective value provided that the stepsizes are sufficiently small.  

\btheorem\label{th:ista}
  Suppose that $F : \R{d} \to \R{}$ is continuously differentiable, strongly convex with constant $c > 0$, and has a gradient that is Lipschitz continuous with constant $L > 0$.  In addition, suppose that $\alpha_k = \alpha \in (0,1/L]$ for all $k \in \N{}$.  Then, for all $k \in \N{}$, the iteration~\eqref{proxgrad} yields
  \bequationn
    \Phi(w_{k+1}) - \Phi(w_*) \leq (1 - \alpha c)^k(\Phi(w_1) - \Phi(w_*)),
  \eequationn
  where $w_* \in \R{d}$ is the unique global minimizer of $\Phi$ in \eqref{nsprob}.
\etheorem
\bproof
  Since $\alpha_k = \alpha \in (0,1/L]$, it follows from \eqref{eq.lipschitz_bound} that
  \bequalin
    \Phi(w_{k+1})
      &=    F(w_{k+1}) + \lambda \Omega(w_{k+1}) \\
      &\leq F(w_k) + \nabla F(w_k)^T(w_{k+1} - w_k) + \thalf L \|w_{k+1} - w_k\|_2^2 + \lambda \Omega(w_{k+1}) \\
      &\leq F(w_k) + \nabla F(w_k)^T(w_{k+1} - w_k) + \tfrac{1}{2\alpha} \|w_{k+1} - w_k\|_2^2 + \lambda \Omega(w_{k+1}) \\
      &\leq F(w_k) + \nabla F(w_k)^T(w - w_k) + \tfrac{1}{2\alpha} \|w - w_k\|_2^2 + \lambda \Omega(w)\ \ \text{for all}\ \ w \in \R{d},
  \eequalin
  where the last inequality follows since $w_{k+1}$ is defined by \eqref{proxgrad}.  Representing $w = w_k + d$, we obtain
  \bequalin
    \Phi(w_{k+1})
      &\leq F(w_k) + \nabla F(w_k)^Td + \tfrac{1}{2\alpha} \|d\|_2^2 + \lambda \Omega(w_k + d) \\
      &\leq F(w_k) + \nabla F(w_k)^Td + \thalf c \|d\|_2^2 - \thalf c \|d\|_2^2 + \tfrac{1}{2\alpha} \|d\|_2^2 + \lambda \Omega(w_k + d) \\
      &\leq F(w_k + d) + \lambda \Omega(w_k + d) - \thalf c \|d\|_2^2 + \tfrac{1}{2\alpha} \|d\|_2^2 \\
      &=    \Phi(w_k + d) + \thalf(\tfrac{1}{\alpha} - c)\|d\|_2^2,
  \eequalin
  which for $d = -\alpha c(w_k - w_*)$ means that
  \begin{align}
    \Phi(w_{k+1})
      &\leq \Phi(w_k - \alpha c(w_k - w_*)) + \thalf(\tfrac{1}{\alpha} - c)\|\alpha c(w_k - w_*)\|_2^2 \nonumber \\
      &=    \Phi(w_k - \alpha c(w_k - w_*)) + \thalf\alpha c^2(1 - \alpha c)\|w_k - w_*\|_2^2. \label{eq.ista_proof_1}
  \end{align}
  On the other hand, since the $c$-strongly convex function $\Phi$ satisfies (e.g., see \cite[pp.~63--64]{Nest04})
  \bequation\label{eq.cstrongbound}
    \baligned
      \Phi(\tau w + (1-\tau) \wbar) \leq&\ \tau \Phi(w) + (1-\tau)\Phi(\wbar) - \tfrac{1}{2} c \tau (1-\tau) \|w-\wbar\|_2^2 \\
      \text{for all} &\ (w,\wbar,\tau) \in \R{d} \times \R{d} \times [0,1],
    \ealigned
  \eequation
  we have (considering $\wbar = w_k$, $w = w_*$, and $\tau = \alpha c \in (0,1]$ in \eqref{eq.cstrongbound}) that
  \begin{align}
    \Phi(w_k - \alpha c(w_k - w_*))
      &\leq \alpha c \Phi(w_*) + (1 - \alpha c)\Phi(w_k) - \thalf c(\alpha c)(1 - \alpha c)\|w_k - w_*\|_2^2 \nonumber \\
      &=    \alpha c \Phi(w_*) + (1 - \alpha c)\Phi(w_k) - \thalf \alpha c^2(1 - \alpha c)\|w_k - w_*\|_2^2. \label{eq.ista_proof_2}
  \end{align}
  Combining \eqref{eq.ista_proof_2} with \eqref{eq.ista_proof_1} and subtracting $\Phi(w_*)$, it follows that
  \bequationn
    \Phi(w_{k+1}) - \Phi(w_*) \leq (1 - \alpha c)(\Phi(w_k) - \Phi(w_*)).
  \eequationn
  The result follows by applying this inequality repeated over the first $k \in \N{}$ iterations.
\eproof

One finds in Theorem~\ref{th:ista} an identical result as for a gradient method for minimizing a smooth strongly convex function.  As in such methods, the choice of the stepsize $\alpha$ is critical in practice; the convergence guarantees demand that it be sufficiently small, but a value that is too small might unduly slow the optimization process.  

The proximal gradient iteration \eqref{proxgrad} is practical only when the proximal mapping
\bequationn
  \text{prox}_{\lambda\Omega,\alpha_k}(\wtilde) := \arg\min_{w\in\R{n}} \(\lambda\Omega(w) + \frac{1}{2\alpha_k}\|w - \wtilde\|_2^2\)
\eequationn
can be computed efficiently.  This can be seen in the fact that the iteration \eqref{proxgrad} can equivalently be written as $w_{k+1} \gets \text{prox}_{\lambda\Omega,\alpha_k}(w_k - \alpha_k\nabla F(w_k))$, i.e., the iteration is equivalent to applying a proximal mapping to the result of a gradient descent step.  Situations when the proximal mapping is inexpensive to compute include when $\Omega$ is the indicator function for a simple set, such as a polyhedral set,  when it is the $\ell_1$-norm, or, more generally, when it is separable.

A stochastic version of the proximal gradient method can be obtained, not surprisingly, by replacing $\nabla F(w_k)$ in \eqref{proxgrad} by a stochastic approximation $g(w_k,\xi_k)$.  The iteration remains cheap to perform (since $F(w_k)$ can be ignored as it does not effect the computed step).  The resulting method attains similar behavior as a stochastic gradient method; analyses can be found, e.g., in~\cite{schmidt2011convergence,atchade2014stochastic}.

We now turn our attention to the most popular nonsmooth regularizer, namely the one defined by the $\ell_1$ norm.

\subsubsection{Iterative Soft-Thresholding Algorithm (ISTA)}   \label{istasection}
In the context of solving the $\ell_1$-norm regularized problem~\eqref{l1prob}, the proximal gradient method is
\bequation\label{proxgrad1}
  w_{k+1} \gets \arg\min_{w\in\R{d}} \(F(w_k) + \nabla F(w_k)^T(w - w_k) + \frac{1}{2\alpha_k}\|w - w_k\|_2^2 + \lambda\| w \|_1\).
\eequation
The optimization problem on the right-hand side of this expression is separable and can be solved in closed form.  The solution can be written component-wise as
\bequation\label{issta_separable}
  [w_{k+1}]_i \gets \bcases [w_k - \alpha_k\nabla F(w_k)]_i + \alpha_k\lambda & \text{if $[w_k - \alpha_k\nabla F(w_k)]_i < -\alpha_k\lambda$} \\ 0 & \text{if $[w_k - \alpha_k\nabla F(w_k)]_i \in [-\alpha_k\lambda,\alpha_k\lambda]$} \\ [w_k - \alpha_k\nabla F(w_k)]_i - \alpha_k\lambda & \text{if $[w_k - \alpha_k\nabla F(w_k)]_i > \alpha_k\lambda$}. \ecases
\eequation
One also finds that this iteration can be written, with $(\cdot)_+ := \max\{\cdot,0\}$, as
\bequation\label{issta}
  w_{k+1} \gets {\cal T}_{\alpha_k\lambda}(w_k - \alpha_k \nabla F(w_k)),\ \ \text{where}\ \ [{\cal T}_{\alpha_k\lambda}(\wtilde)]_i = (|\wtilde_i| - \alpha_k\lambda)_+ \sgn(\wtilde_i).
\eequation
In this form, ${\cal T}_{\alpha_k\lambda}$ is referred to as the soft-thresholding operator, which leads to the name \emph{iterative soft-thresholding algorithm} (ISTA) being used for \eqref{proxgrad1}--\eqref{issta_separable} \cite{Dono95,DaubDefrDeMo04}.

It is clear from \eqref{issta_separable} that the ISTA iteration induces sparsity in the iterates.  If the steepest descent step with respect to $F$ yields a component with absolute value less than $\alpha_k\lambda$, then that component is set to zero in the subsequent iterate; otherwise, the operator still has the effect of shrinking components of the solution estimates in terms of their magnitudes.  When only a stochastic estimate $g(w_k,\xi_k)$ of the gradient is available, it can be used instead of $\nabla F(w_k)$.

A variant of ISTA with acceleration (recall \S\ref{sec.accelerated}), known as FISTA \cite{BeckTebo09}, is popular in practice.
%
We also mention that effective techniques have been developed for computing the stepsize $\alpha_k$, in ISTA or FISTA, based on an estimate of the Lipschitz constant of $\nabla F$  or on curvature measured in recent iterations \cite{BeckTebo09,WrigNowaFigu09,BeckCandGran11}.

\subsubsection{Bound-Constrained Methods for $\ell_1$-norm Regularized Problems}
 
By observing the structure created by the $\ell_1$-norm, one finds that an equivalent \emph{smooth} reformulation of problem~\eqref{l1prob} is easily derived.  In particular, by writing $w = u - v$ where $u$ plays the \emph{positive part} of $w$ while $v$ plays the \emph{negative part}, problem \eqref{l1prob} can equivalently be written as
\bequation\label{l1probsmooth}
  \min_{(u,v)\in\R{d}\times\R{d}}\ \tilde\phi(u,v)\ \ \st\ \ (u,v) \geq 0,\ \ \text{where}\ \ \tilde\phi(u,v) = F(u-v) + \lambda \sum_{i=1}^d (u_i + v_i).
\eequation
Now with a bound-constrained problem in hand, one has at their disposal a variety of optimization methods that have been developed in the optimization literature.

The fundamental iteration for solving bound-constrained optimization problems is  the \emph{gradient projection} method.  In the context of~\eqref{l1probsmooth}, the iteration reduces to
\bequation\label{eq.gradproj}
  \bbmatrix u_{k+1} \\ v_{k+1} \ebmatrix \gets P_+\(\bbmatrix u_k \\ v_k \ebmatrix - \alpha_k \bbmatrix \nabla_u \tilde\phi(u_k,v_k) \\ \nabla_v \tilde\phi(u_k,v_k) \ebmatrix\) = P_+\(\bbmatrix u_k - \alpha_k \nabla F(u_k-v_k) - \alpha_k\lambda e \\ v_k + \alpha_k \nabla F(u_k-v_k) - \alpha_k\lambda e \ebmatrix\),
\eequation
where $P_+$ projects onto the nonnegative orthant and $e \in \R{d}$ is a vector of ones.  

Interestingly, the gradient projection method can also be derived from the perspective of a proximal gradient method where the regularization term $\Omega$ is chosen to be the indicator function for the feasible set (a box).  In this case, the mapping ${\cal T}_{\alpha_k\lambda}$ is replaced by the projection operator onto the bound constraints, causing the corresponding proximal gradient method to coincide with the gradient projection method. In the light of this observation,  one should expect the iteration~\eqref{eq.gradproj} to inherit the property of being globally linearly convergent when $F$ satisfies the assumptions of Theorem~\ref{th:ista}.  
However, since the variables in \eqref{l1probsmooth} have been split into positive and negative parts, this property is maintained \emph{only if} the iteration maintains complementarity of each iterate pair, i.e., if $[u_k]_i[v_k]_i = 0$ for all $k \in \N{}$ and $i \in \{1,\dots,d\}$.  This behavior is also critical for the practical performance of the method in general since, without it, the algorithm would not generate sparse solutions.  In particular, maintaining this property allows the algorithm to be implemented in such a way that one effectively only needs $d$ optimization variables.

A natural question that arises is whether the iteration \eqref{eq.gradproj} actually differs from an ISTA iteration, especially given that both are built upon proximal gradient ideas.  In fact, the iterations can lead to the same update, but do not always.  Consider, for example, an iterate $w_k = u_k - v_k$ such that for $i \in \{1,\dots,d\}$ one finds $[w_k]_i > 0$ with $[u_k]_i = [w_k]_i$ and $[v_k]_i = 0$.  (A similar look, with various signs reversed, can be taken when $[w_k]_i < 0$.)  If $[w_k - \alpha_k \nabla F(w_k)]_i > \alpha_k\lambda$, then \eqref{issta_separable} and \eqref{eq.gradproj} yield
\bequalin{}
  [w_{k+1}]_i &\gets [w_k - \alpha_k\nabla F(w_k)]_i - \alpha_k\lambda > 0 \\ \text{and}\ \ 
  [u_{k+1}]_i &\gets [u_k - \alpha_k\nabla F(w_k)]_i - \alpha_k\lambda > 0.
\eequalin
However, it is important to note the step taken in the negative part; in particular, if $[\nabla F(w_k)]_i \leq \lambda$, then $[v_{k+1}]_i \gets 0$, but, if $[\nabla F(w_k)]_i > \lambda$, then $[v_{k+1}]_i \gets \alpha_k \nabla F(w_k) - \alpha_k \lambda$, in which case the lack of complementarity between $u_{k+1}$ and $v_{k+1}$ should be rectified.  A more significant difference arises when, e.g., $[w_k - \alpha_k \nabla F(w_k)]_i < -\alpha_k\lambda$, in which case \eqref{issta_separable} and \eqref{eq.gradproj} yield
\bequalin{}
  [w_{k+1}]_i &\gets [w_k - \alpha_k\nabla F(w_k)]_i + \alpha_k\lambda < 0, \\
  [u_{k+1}]_i &\gets 0, \\ \text{and}\ \ 
  [v_{k+1}]_i &\gets [v_k + \alpha_k\nabla F(w_k)]_i - \alpha_k\lambda > 0.
\eequalin
The pair $([u_{k+1}]_i,[v_{k+1}]_i)$ are complementary, but $[w_{k+1}]_i$ and $[-v_{k+1}]_i$ differ by $[w_k]_i > 0$.

Several first-order \cite{FiguNowaWrig07,fan2008liblinear} and second-order \cite{Schm10} gradient projection methods have been proposed to solve \eqref{l1prob}.  Such algorithms should be preferred over similar techniques for general bound-constrained optimization, e.g., those in~\cite{ZhuByrdLuNoce97,lin1999newton}, since such general techniques may be less effective by not exploiting the structure of the reformulation \eqref{l1probsmooth} of \eqref{l1prob}.

A stochastic projected gradient method, with $\nabla F(w_k)$ replaced by $g(w_k,\xi_k)$, has similar convergence properties as a standard SG method; e.g., see \cite{NemiJudiLanShap09}.  These properties apply in the present context, but also apply when a proximal gradient method is used to solve~\eqref{nsprob} when $\Omega$ represents the indicator function of a box constraint.
 
\subsection{Second-Order Methods}\label{l1-second}

We now turn our attention to methods that, like Newton's method for smooth optimization, are designed to solve regularized problems through successive minimization of second-order models constructed along the iterate sequence $\{w_k\}$.  As in a proximal gradient method, the smooth function~$F$ is approximated by a Taylor series and the regularization term is kept unchanged.  We focus on two classes of methods for solving~\eqref{l1prob}: \emph{proximal Newton} and \emph{orthant-based} methods.

Both classes of methods fall under the category of \emph{active-set} methods. One could also consider the application of an \emph{interior-point} method to solve the bound-constrained problem \eqref{l1probsmooth}~\cite{NoceWrig06}.  This, by its nature, constitutes a second-order method that would employ Hessians of $F$ or corresponding quasi-Newton approximations.  However, a disadvantage of the interior-point approach is that, by staying away from the boundary of the feasible region, it does not promote the fast generation of sparse solutions, which is in stark contrast with the methods described below.  

\subsubsection{Proximal Newton Methods}\label{l1-prox}

We use the term \emph{proximal Newton} to refer to techniques that directly minimize the nonsmooth function arising as the sum of a quadratic model of~$F$ and the regularizer.  In particular, for solving problem~\eqref{l1prob}, a proximal Newton method is one that constructs, at each $k \in \N{}$, a model
\bequation\label{sqa}
  q_k(w) = F(w_k) + \nabla F(w_k)^T(w - w_k) + \frac{1}{2}(w - w_k)^TH_k(w - w_k) + \lambda \|w\|_1 \approx \phi(w).
\eequation
where $H_k$ represents $\nabla^2 F(w_k)$ or a quasi-Newton approximation of it.  This model has a similar form as the one in~\eqref{proxgrad1}, except that the simple quadratic is replaced by the quadratic form defined by~$H_k$.  A proximal Newton method would involve (approximately) minimizing this model to compute a trial iterate $\wtilde_k$, then a stepsize $\alpha_k > 0$ would be taken from a predetermined sequence or chosen by a line search to ensure that the new iterate $w_{k+1} \gets w_k + \alpha_k(\wtilde_k - w_k)$ yields $\Phi(w_{k+1}) < \Phi(w_k)$.

Proximal Newton methods are more challenging to design, analyze, and implement than proximal gradient methods.  That being said, they can perform better in practice once a few key challenges are addressed.  The three ingredients below have proved to be essential in ensuring the practical success and scalability of a proximal Newton method.  
For simplicity, we assume throughout that $H_k$ has been chosen to be positive definite.

\paragraph{\textbf{Choice of Subproblem Solver}}

The model $q_k$ inherits the nonsmooth structure of $\phi$, which has the benefit of allowing a proximal Newton method to cross manifolds of nondifferentiability while simultaneously promoting sparsity of the iterates.  However, the method needs to overcome the fact that the model $q_k$ is nonsmooth, which makes the subproblem for minimizing $q_k$ challenging.  Fortunately, with the particular structure created by a quadratic plus an $\ell_1$-norm term, various methods are available for minimizing such a nonsmooth function.  For example, coordinate descent is particularly well-suited in this context \cite{tseng2009coordinate,hsieh2011sparse} since the global minimizer of $q_k$ along a coordinate descent direction can be computed analytically.  Such a minimizer often occurs at a point of nondifferentiability (namely, when a component is zero), thus ensuring that the method will generate sparse iterates.  Updating the gradient of the model $q_k$ after each coordinate descent step can also be performed efficiently, even if $H_k$ is given as a limited memory quasi-Newton approximation \cite{scheinberg2013practical}. Numerical experiments have shown that employing a coordinate descent iteration is more efficient in certain applications than employing, say, an ISTA iteration to minimize $q_k$, though the latter is also a viable strategy in some applications.  

\paragraph{\textbf{Inaccurate Subproblem Solves}}

A proximal Newton method needs to overcome the fact that,  in large-scale settings, it is impractical to minimize $q_k$ accurately for all $k \in \N{}$.  Hence, it is natural to consider the computation of an inexact minimizer of $q_k$.  The issue then becomes: what are practical, yet theoretically sufficient termination criteria when computing an approximate minimizer of the nonsmooth function~$q_k$?  A common suggestion in the literature has been to use the norm of the minimum-norm subgradient of $q_k$ at a given approximate minimizer.  However, this measure is not continuous, making it inadequate for these purposes.\footnote{Consider the one-dimensional case of having $q_k(w) = |w|$.  The minimum-norm subgradient of $q_k$ has a magnitude of 1 at all points, except at the minimizer $w^*=0$; hence, this norm does not provide a measure of proximity to $w^*$.}  Interestingly, the norm of an ISTA step  is an appropriate measure.  In particular, letting $\text{ista}_k(w)$ represent the result of an ISTA step applied to $q_k$ from $w$, the value $\|\text{ista}_k(w) - w\|_2$ satisfies the following two key properties: $(i)$ it equals zero if and only if $w$ is a minimizer of $q_k$, and $(ii)$ it varies continuously over $\R{d}$.  

Complete and sufficient termination criteria are then as follows: a trial point $\wtilde_k$ represents a sufficiently accurate minimizer of $q_k$ if, for some $\eta \in [0,1)$, one finds
\bequationn
  \|\text{ista}_k(\wtilde_k) - \wtilde_k\|_2 \leq \eta \|\text{ista}(w_k) - w_k\|_2\ \ \text{and}\ \ q_k(\wtilde_k) < q_k(w_k).
\eequationn
The latter condition, requiring a decrease in $q_k$, must also be imposed since the ISTA criterion alone does not exert sufficient control to ensure convergence.  Employing such criteria, it has been observed to be efficient to perform the minimization of $q_k$ inaccurately at the start, and to increase the accuracy of the model minimizers as one approaches the solution.  A superlinear rate of convergence for the overall proximal Newton iteration can be obtained by replacing $\eta$ by $\eta_k$ where $\{\eta_k\} \searrow 0$, along with imposing a stronger descent condition on the decrease in $q_k$ \cite{byrd2013inexact}.

\paragraph{\textbf{Elimination of Variables}}

Due to the structure created by the $\ell_1$-norm regularizer, it can be effective in some applications to first identify a set of \emph{active} variables---i.e., ones that are predicted to be equal to zero at a minimizer for $q_k$---then compute an approximate minimizer of $q_k$ over the remaining \emph{free} variables.  Specifically, supposing that a set $\Acal_k \subseteq \{1,\dots,d\}$ of active variables has been identified, one may compute an (approximate) minimizer of $q_k$ by (approximately) solving the reduced-space problem
\bequation\label{prob.reduced}
  \min_{w \in \R{d}}\ q_k(w)\ \ \st\ \ [w]_i = 0, \ i \in \Acal_k.
\eequation
Moreover, during the minimization process for this problem, one may have reason to believe that the process may be improved by adding or removing elements from the active set estimate $\Acal_k$.  In any case, performing the elimination of variables imposed in \eqref{prob.reduced} has the effect of reducing the size of the subproblem, and can often lead to fewer iterations being required in the overall proximal Newton method.  How should the active set $\Acal_k$ be defined? A technique that has become popular recently is to use sensitivity information 
as discussed in more detail in the next subsection.

\subsubsection{Orthant-Based Methods}

Our second class of second-order methods is based on the observation that the $\ell_1$-norm regularized objective $\phi$ in problem~\eqref{l1prob} is smooth in any orthant in $\R{d}$.  Based on this observation, orthant-based methods construct, in every iteration, a smooth quadratic model of the objective, then produce a search direction by minimizing this model.  After performing a line search designed to reduce the objective function, a new orthant is selected and the process is repeated. This approach can be motivated by the success of second-order gradient projection methods for bound constrained optimization, which at every iteration employ a gradient projection search to identify an active set and  perform a minimization of the objective function over a space of free variables to compute a search direction.

The selection of an orthant is typically done using sensitivity information.  Specifically, with the minimum norm subgradient of $\phi$ at $w \in \R{d}$, which is given component-wise for all $i \in \{1,\dots,d\}$ by
\begin{align}\label{eqn:MNSG}
 \ghat_i(w) = \begin{cases}
   [\nabla F(w)]_i + \lambda & \text{if}\ w_i>0\ \text{or}\ \{w_i=0\ \text{and}\ [\nabla F(w)]_i + \lambda <0 \} \\
   [\nabla F(w)]_i - \lambda & \text{if}\ w_i<0\ \text{or}\ \{w_i=0\ \text{and}\ [\nabla F(w)]_i - \lambda >0 \} \\
   0 & \text{otherwise}, \end{cases} 
\end{align}
the active orthant for an iterate $w_k$ is characterized by the sign vector
\begin{equation}\label{zeta}
  \zeta_{k,i} = \begin{cases} \sgn([w_k]_i) & \text{if}\ [w_k]_i\neq 0 \\ \sgn(-[\hat g(w_k)]_i) & \text{if}\ [w_k]_i= 0. \end{cases}  
\end{equation}
%
%
%
%
Along these lines, let us also define the subsets of $\{1,\dots,d\}$ given by
\begin{align}
  \mathcal{A}_k &= \left\{i : [w_k]_i = 0\ \text{and}\ \left|[\nabla F(w_k)]_i\right| \leq \lambda \right\} \label{aset} \\ \text{and}\ 
  \mathcal{F}_k &= \{i : [w_k]_i \neq 0 \} \cup \left\{i : [w_k]_i =0\ \text{and}\ \left|[\nabla F(w_k)]_i\right| > \lambda \right\}, \label{fsure} 
\end{align}
where $\Acal_k$ represents the indices of variables that are active and kept at zero while $\Fcal_k$ represents those that are free to move.

Given these quantities, an orthant-based method proceeds as follows.  First, one computes the (approximate) solution $d_k$ of the (smooth) quadratic problem
\bequalin
  \min_{d \in \R{n}} &\ \ghat(w_k)^Td + \thalf d^TH_kd \\
  \st & \ d_i=0,\ \ i \in \mathcal{A}_k,
\eequalin
where $H_k$ represents $\nabla^2 F(x^k)$ or an approximation of it.  Then, the algorithm performs a line search---over a path contained in the current orthant---to compute the next iterate.  For example, one option is a projected backtracking line search along $d_k$, which involves computing the largest $\alpha_k$ in a decreasing geometric sequence so
\bequationn
  F(P_k(w_k + \alpha_k d_k)) < F(w_k).
\eequationn 
Here, $P_k(w)$ projects $w \in \mathbb{R}^d$ onto the orthant defined by $\zeta^k$, i.e.,
\begin{equation}
      [P_k(w)]_i = \begin{cases} 
      w_i & \text{if}\ \sgn(w_i) = \zeta_{k,i} \\
      0 & \mbox{otherwise.}
      \end{cases}  
\end{equation}
In this way, the initial and final points of an iteration lie in the same orthant.  Orthant-based methods have proved to be quite useful in practice; e.g., see  \cite{andrew2007scalable,byrd2012family}.

\paragraph{\textbf{Commentary}}

Proximal Newton and orthant-based methods represent two efficient classes of second-order active-set methods for solving the $\ell_1$-norm regularized problem \eqref{l1prob}. The proximal Newton method is reminiscent of the sequential quadratic programming method (SQP) for constrained optimization; they both solve a complex subproblem that yields a useful estimate of the optimal active set. Although solving the piecewise quadratic model \eqref{sqa} is very expensive in general, the coordinate descent method has proven to be well suited for this task, and allows the proximal Newton method to be applied to very large problems \cite{hsieh2013big}. Orthant-based methods have shown to be equally useful, but in a more heuristic way, since some popular implementations lack convergence guarantees \cite{fan2008liblinear,byrd2012family}.  Stochastic variants of both proximal Newton and orthant-based schemes can be devised in natural ways, and generally inherit the properties of stochastic proximal gradient methods as long as the Hessian approximations are forced to possess eigenvalues within a positive interval.

\section{Summary and Perspectives}\label{sec.topics_not_covered}
\setcounter{equation}{0}
\setcounter{theorem}{0}
\setcounter{algorithm}{0}
\setcounter{figure}{0}
\setcounter{table}{0}

Mathematical optimization is one of the foundations of machine learning, touching almost every aspect of  the discipline.  In particular, numerical optimization algorithms, the main subject of this paper, have played an integral role in the transformational progress that machine learning has experienced over the past two decades.  In our study, we highlight the dominant role played by the stochastic gradient method (SG) of Robbins and Monro \cite{RobbMonr51}, whose success derives from its superior work complexity guarantees.  A concise, yet broadly applicable convergence and complexity theory for SG is presented here, providing insight into how these guarantees have translated into practical gains.

Although the title of this paper suggests that our treatment of optimization methods for machine learning is comprehensive, much more could be said about this rapidly evolving field.  Perhaps most importantly, we have neither discussed nor analyzed at length the opportunities offered by parallel and distributed computing, which may alter our perspectives in the years to come.  In fact, it has already been widely acknowledged that SG, despite its other benefits, may not be the best suited method for emerging computer architectures. 

This leads to our discussion of a spectrum of methods that  have the potential to surpass SG in the next generation of optimization methods for machine learning.  These methods, such as those built on noise reduction and second-order techniques, offer the ability to attain improved convergence rates, overcome the adverse effects of high nonlinearity and ill-conditioning, and exploit parallelism and distributed architectures in new  ways.  There are important methods that are not included in our presentation---such as the alternating direction method of multipliers (ADMM) \cite{EcksBert92,GabaMerc76,GlowMarr75} and the expectation-maximization (EM) method and its variants \cite{DempLairRubi77,Wu83}---but our study covers many of the core algorithmic frameworks in optimization for machine learning, with emphasis on methods and theoretical guarantees that have the largest impact on practical performance.

With the great strides that have been made and the various avenues for continued contributions, numerical optimization promises to continue to have a profound impact on the rapidly growing field of machine learning.

\appendix

\section{Convexity and Analyses of SG}\label{app.convex}
\setcounter{equation}{0}

Our analyses of SG in \S\ref{sec.sg} can be characterized as relying primarily on smoothness in the sense of Assumption~\ref{ass.Lipschitz}.  This has advantages and disadvantages.  On the positive side, it allows us to prove convergence results that apply equally for the minimization of convex and nonconvex functions, the latter of which has been rising in importance in machine learning; recall~\S\ref{sec.deep_neural_nets}.  It also allows us to present results that apply equally to situations in which the stochastic vectors are unbiased estimators of the gradient of the objective, or when such estimators are scaled by a symmetric positive definite matrix; recall~\eqref{brackii}.  A downside, however, is that it requires us to handle the minimization of nonsmooth models separately, which we do in~\S\ref{sec.nonsmooth}.

As an alternative, a common tactic employed by many authors is to leverage convexity instead of smoothness, allowing for the establishment of guarantees that can be applied in smooth and nonsmooth settings.  For example, a typical approach for analyzing stochastic gradient-based methods is to commence with the following fundamental equations related to squared distances to the optimum:
\begin{align}
  \|w_{k+1}-w_*\|_2^2 - \|w_{k}-w_*\|_2^2
    &= 2 (w_{k+1}-w_k)^T(w_k-w_*) + \|w_{k+1}-w_k\|_2^2 \nonumber\\
    &= -2\alpha_k g(w_k,\xi_k)^T(w_k-w_*) + \alpha_k^2 \|g(w_k,\xi_k)\|_2^2. \label{eq.wbound}
\end{align}
Assuming that $\E_{\xi_k}[g(w_k,\xi_k)] = \ghat(w_k) \in \partial F(w_k)$, one then obtains
\begin{align}
   &\ \E_{\xi_k}[\|w_{k+1}-w_*\|_2^2] - \|w_k - w_*\|_2^2 \nonumber \\
  =&\ -2\alpha_k\ghat(w_k)^T(w_k-w_*) + \alpha_k^2 \E_{\xi_k}[\|g(w_k,\xi_k)\|_2^2], \label{eq.convex_uh_oh}
\end{align}
which has certain similarities with~\eqref{eq.expected_decrease_pre}.  One can now introduce an assumption of convexity to bound the first term on the right-hand side in this expression; in particular, convexity offers the subgradient inequality
\bequationn
  \ghat(w_k)^T(w_k-w_*) \geq F(w_k)-F(w_*) \geq 0
\eequationn
while strong convexity offers the stronger condition \eqref{eq.sc}.  Combined with a suitable assumption on the second moment of the stochastic subgradients to bound the second term in the expression, the entire right-hand side can be adequately controlled through judicious stepsize choices.  The resulting analysis then has many similarities with that presented in~\S\ref{sec.sg}, especially if one introduces an assumption about Lipscithz continuity of the gradients of $F$ in order to translate results on decreases in the distance to the solution in terms of decreases in $F$ itself.  The interested reader will find a clear exposition of such results in \cite{NemiJudiLanShap09}.

Note, however, that one can see in~\eqref{eq.convex_uh_oh} that analyses based on distances to the solution do not carry over easily to nonconvex settings or when (quasi-)Newton type steps are employed.  In such situations, without explicit knowledge of $w_*$, one cannot easily ensure that the first term on the right-hand side can be bounded appropriately.

\section{Proofs}\label{sec.proofs}
\setcounter{equation}{0}

\bproof[Inequality~\eqref{eq.lipschitz_bound}]
Under Assumption~\ref{ass.Lipschitz}, one obtains
\bequalin
F(w) &= F(\wbar) + \int_{0}^{1} \frac{\partial F(\wbar+t(w-\wbar))}{\partial t}\,dt \\
     &= F(\wbar) + \int_{0}^{1} \nabla F(\wbar+t(w-\wbar))^T (w-\wbar) \, dt \\
     &= F(\wbar) + \nabla F(\wbar)^T(w-\wbar)
                 + \int_{0}^{1} \left[ \nabla F(\wbar+t(w-\wbar)) - \nabla F(\wbar) \right]^T (w-\wbar) \, dt \\
     &\leq F(\wbar) + \nabla F(\wbar)^T(w-\wbar)
                 + \int_{0}^{1} L \| t (w-\wbar)\|_2 \|w-\wbar\|_2 \, dt,
\eequalin
from which the desired result follows.
\eproof

\bproof[Inequality~\eqref{eq.sc2}]
  Given $w \in \R{d}$, the quadratic model
  \bequationn
    q(\wbar) := F(w) + \nabla F(w)^T(\wbar-w) + \thalf c \|\wbar-w\|_2^2
  \eequationn
  has the unique minimizer $\wbar_* := w - \tfrac{1}{c}\nabla F(w)$ with $q(\wbar_*) = F(w) - \tfrac{1}{2c}\|\nabla F(w)\|_2^2$.  Hence, the inequality \eqref{eq.sc} with $\wbar = w_*$ and any $w \in \R{d}$ yields
  \bequationn
    F_*
      \geq F(w) + \nabla F(w)^T(w_*-w) + \thalf c \|w_*-w\|_2^2 \\
      \geq F(w) - \tfrac{1}{2c}\|\nabla F(w)\|_2^2,
  \eequationn
  from which the desired result follows.
\eproof

\bproof[Corollary~\ref{th.sg_dd}]
Define $G(w) := \|\nabla F(w)\|_2^2$ and let $L_G$ be the Lipschitz constant of $\nabla G(w) = 2 \nabla^2 F(w) \nabla F(w)$.  Then,
  \bequalin
    G(w_{k+1}) - G(w_k)
      &\leq \nabla G(w_k)^T (w_{k+1}-w_k) + \thalf L_G \|w_{k+1}-w_k\|^2_2 \\
      &\leq -\alpha_k \nabla G(w_k)^T g(w_k,\xi_k) + \thalf \alpha_k^2 L_G \|g(w_k,\xi_k)\|^2_2.
  \eequalin
  Taking the expectation with respect to the distribution of $\xi_k$, one obtains from Assumptions~\ref{ass.Lipschitz} and \ref{ass.sg} and the inequality \eqref{eq.second_moment} that
  \bequalin
        &\ \E_{\xi_k}[G(w_{k+1}] - G(w_k) \\
    \leq&\ -2 \alpha_k \nabla F(w_k)^T \nabla^2 F(w_k)^T E_{\xi_k}[g(w_k,\xi_k)] + \thalf \alpha_k^2 L_G E_{\xi_k}[\|g(w_k,\xi_k)\|^2_2] \\
    \leq&\ 2 \alpha_k \|\nabla F(w_k)\|_2 \| \nabla^2 F(w_k)\|_2 \|E_{\xi_k}[g(w_k,\xi_k)]\|_2 + \thalf \alpha_k^2 L_G E_{\xi_k}[\|g(w_k,\xi_k)\|^2_2] \\
    \leq&\ 2 \alpha_k L \mu_G \|\nabla F(w_k)\|^2_2 + \thalf\alpha_k^2 L_G (M + M_V \|\nabla F(w_k)\|^2_2).
\eequalin
  Taking the total expectation simply yields
  \bequation\label{eq.gincreases}
    \baligned
          &\ \E[G(w_{k+1})] - \E[G(w_k)] \\
      \leq&\ 2 \alpha_k L \mu_G \E[\|\nabla F(w_k)\|^2_2] + \thalf\alpha_k^2 L_G (M + M_V \E[\|\nabla F(w_k)\|^2_2]).
    \ealigned
  \eequation
  Recall that Theorem~\ref{th.sg_nc} establishes that the first component of this bound is the term of a convergent sum.  The second component of this bound is also the term of a convergent sum since $\sum_{k=1}^\infty \alpha_k^2$ converges and since $\alpha_k^2\leq\alpha_k$ for sufficiently large $k \in \N{}$, meaning that again the result of Theorem~\ref{th.sg_nc} can be applied.  Therefore, the right-hand side of \eqref{eq.gincreases} is the term of a convergent sum. Let us now define
  \bequationn
    \baligned
    S_K^+ &= \sum_{k=1}^{K}\max(0,\E[G(w_{k+1})] - \E[G(w_k)]) \\ \text{and}\ \ 
    S_K^- &= \sum_{k=1}^{K}\max(0,\E[G(w_k)] - \E[G(w_{k+1})]).
    \ealigned
  \eequationn
Since the bound \eqref{eq.gincreases} is positive and forms a convergent sum, the nondecreasing sequence $S_K^+$ is upper bounded and therefore converges. Since, for any $K \in \N{}$, one has $G(w_K) = G(w_0) + S_K^+ - S_K^- \geq0$, the nondecreasing sequence $S_K^-$ is upper bounded and therefore also converges. Therefore $G(w_K)$ converges and Theorem~\ref{th.sg_liminf} means that this limit must be zero. 
\eproof

\bibliographystyle{plain}
\bibliography{optml_bib}

\end{document}